\documentclass[conference]{IEEEtran}
\IEEEoverridecommandlockouts

\usepackage{amssymb}
\usepackage{amsthm}
\usepackage{booktabs}
\usepackage{caption}
\usepackage{cite}
\usepackage{color}
\usepackage{enumerate}
\let\labelindent\relax 
\usepackage{enumitem}
\usepackage{float}
\usepackage{graphicx}
\usepackage{hyperref}
\hypersetup{
	colorlinks   = true,
	citecolor    = blue
}
\usepackage{makecell}
\usepackage{multirow}
\usepackage{multicol}
\usepackage{pgfplots}
\usepackage{pgfplotstable}
\usepackage{ragged2e}
\usepackage{stmaryrd}
\usepackage{tabu}
\usepackage{tabularx}
\usepackage{tikz}
\usepackage{todonotes}
\usepackage{url}
\usepackage{verbatim}
\usepackage[english]{babel}
\usepackage [autostyle, english = american]{csquotes}
\MakeOuterQuote{"}
\usepackage[framemethod=tikz]{mdframed} 
\usepackage{algpseudocode}
\usepackage{nccmath}
\definecolor{darkred}{rgb}{.6,0,0}

\newtheorem{theorem}{Theorem}[section]

\newtheorem{lemma}[theorem]{Lemma}

\theoremstyle{definition}

\newcommand{\xor}{\oplus}

\newcommand{\band}{\odot}

\newcommand{\BigO}[1]{\ensuremath{\operatorname{O}\left(#1\right)}}

\newcommand{\bitset}{\{0,1\}}

\newcommand{\Z}[1]{\ensuremath{\mathbb{Z}}_{2^{#1}}}

\newcommand{\Mb}[1]{\ensuremath{\mathbf{#1}\times}}

\setlist[description]{style=unboxed,leftmargin=0cm}

\newenvironment{myitemize}{
	\begin{list}{{$\bullet$}}{
			\setlength\partopsep{0pt}
			\setlength\parskip{0pt}
			\setlength\parsep{0pt}
			\setlength\topsep{2pt}
			\setlength\itemsep{4pt}
			\setlength{\itemindent}{0pt}
			\setlength{\leftmargin}{8pt}
		}
	}{
	\vspace{1mm}
	\end{list}
}

\newenvironment{mylist}{
	\begin{list}{{$\bullet$}}{
			\setlength\partopsep{0pt}
			\setlength\parskip{0pt}
			\setlength\parsep{0pt}
			\setlength\topsep{2pt}
			\setlength\itemsep{1pt}
			\setlength{\itemindent}{0pt}
			\setlength{\leftmargin}{9pt}
		}
	}{
	\end{list}
}

\newcounter{itemcount}

\newcommand{\tabref}[1]{Table~\protect\ref{tab:#1}}
\newcommand{\secref}[1]{Section~\protect\ref{sec:#1}}

\newcommand{\figlab}[1]{\label{fig:#1}}

\newenvironment{boxfig*}[2]{
	\begin{figure*}[h!]		
		\fontsize{5}{5}\selectfont
		\newcommand{\FigCaption}{#1}
		\newcommand{\FigLabel}{#2}
		\vspace{-.05cm}
		\begin{center}
			\begin{small}			 
				\begin{adjustbox}{max width=\textwidth}
					\begin{tabular}{@{}|@{~~}l@{~~}|@{}}
						\hline
						\rule[-1ex]{0pt}{1ex}\begin{minipage}[b]{.95\linewidth}
							\vspace{1ex}	
						}{%
						\end{minipage}\\
						\hline
					\end{tabular}	
				\end{adjustbox}		
			\end{small}
			\vspace{-0.1cm}
			\caption{\FigCaption}
			\figlab{\FigLabel}
		\end{center}
		\vspace{-.38cm}
	\end{figure*}
}

\newenvironment{myboxfig*}[2]{
	\begin{figure*}[!htb]		
		\fontsize{5}{5}\selectfont
		\newcommand{\FigCaption}{#1}
		\newcommand{\FigLabel}{#2}
		\vspace{-.10cm}
		\begin{center}
			\caption{\FigCaption}
			\begin{small}			 
				\begin{adjustbox}{max width=\textwidth}
					\begin{tabular}{@{}|@{~~}l@{~~}|@{}}
						\hline
						\rule[-1ex]{0pt}{1ex}\begin{minipage}[b]{.95\linewidth}
							\vspace{1ex}	
						}{%
						\end{minipage}\\
						\hline
					\end{tabular}	
				\end{adjustbox}		
			\end{small}
			\vspace{-0.25cm}
			\figlab{\FigLabel}
		\end{center}
		\vspace{-.38cm}
	\end{figure*}
}

\newcommand{\boxref}[1]{Fig.~\ref{#1}}

\RequirePackage{mdframed}

\newenvironment{titlebox}[5]
{\mdfsetup{
		style=#2,
		innertopmargin=1.1\baselineskip,
		skipabove={\dimexpr0.2\baselineskip+\topskip\relax},
		skipbelow={1em},needspace=3\baselineskip,
		singleextra={\node[#3,right=10pt,overlay] at (P-|O){~{\sffamily\bfseries #1 }};},%
		firstextra={\node[#3,right=10pt,overlay] at (P-|O) {~{\sffamily\bfseries #1 }};},
		frametitleaboveskip=9em,
		innerrightmargin=5pt
	}
	\newcommand{\TitleCaption}{#4}
	\newcommand{\TitleLabel}{#5}
	\begin{mdframed}[font=\small]
		\setlist[itemize]{leftmargin=13pt}\setlist[enumerate]{leftmargin=13pt}\raggedright%
	}
	{\end{mdframed}
	\vspace{-1.5em}
	{\captionof{figure}{\small \TitleCaption}\label{\TitleLabel}}
	\medskip
}

\tikzstyle{normal} = [thick, fill=white, text=black, draw, rounded corners, rectangle, minimum height=.7cm, inner sep=3pt]
\tikzstyle{gray} = [thick, fill=gray!90, text=white, rounded corners, rectangle, minimum height=.7cm, inner sep=3pt]

\mdfdefinestyle{commonbox}{%
	align=center, middlelinewidth=1.1pt,userdefinedwidth=\linewidth
	innerrightmargin=0pt,innerleftmargin=2pt,innertopmargin=5pt,
	splittopskip=7pt,splitbottomskip=3pt
}

\mdfdefinestyle{roundbox}{style=commonbox,roundcorner=5pt,userdefinedwidth=\linewidth}

\newenvironment{systembox}[3]
{\vspace{\baselineskip}\begin{titlebox}{Functionality \normalfont #1}{roundbox}{normal}{#2}{#3}}
	{\end{titlebox}}

\newenvironment{protocolbox}[3]
{\begin{titlebox}{Protocol \normalfont #1}{commonbox}{normal}{#2}{#3}}
	{\end{titlebox}}

\newenvironment{simulatorbox}[3]
{\begin{titlebox}{Simulator \normalfont #1}{commonbox}{normal}{#2}{#3}}
	{\end{titlebox}}

\newenvironment{splittitlebox}[5]
{\mdfsetup{
		style=#2,
		innertopmargin=1.1\baselineskip,
		skipabove={\dimexpr0.2\baselineskip+\topskip\relax},
		skipbelow={1em},needspace=3\baselineskip,
		singleextra={\node[#3,right=10pt,overlay] at (P-|O){~{\sffamily\bfseries #1 }};},%
		firstextra={\node[#3,right=10pt,overlay] at (P-|O) {~{\sffamily\bfseries #1 }};},
		frametitleaboveskip=9em,
		innerrightmargin=5pt
	}
	\newcommand{\TitleCaption}{#4}
	\newcommand{\TitleLabel}{#5}
	\begin{mdframed}[font=\small]
		\setlist[itemize]{leftmargin=13pt}\setlist[enumerate]{leftmargin=13pt}\raggedright%
	}
	{\end{mdframed}
	\vspace{-1.6em}
	{\captionof{figure}{\small \TitleCaption}\label{\TitleLabel}}
	\medskip
}

\mdfdefinestyle{commonsplitbox}{%
	align=center, middlelinewidth=1.1pt,userdefinedwidth=\linewidth
	innerrightmargin=0pt,innerleftmargin=2pt,innertopmargin=5pt,
	splittopskip=7pt,splitbottomskip=3pt
}

\newenvironment{protocolsplitbox}[3]
{\begin{splittitlebox}{Protocol \normalfont #1}{commonsplitbox}{normal}{#2}{#3}}
	{\end{splittitlebox}}

\newenvironment{systembox*}[3]
{\begin{strip}
\vspace{\baselineskip}\begin{titlebox}{Functionality \normalfont #1}{roundbox}{normal}{#2}{#3}}
	{\end{titlebox}
\end{strip}}

\newenvironment{gsystembox*}[3]
{\begin{strip}
\vspace{\baselineskip}\begin{titlebox}{Global Functionality \normalfont #1}{roundbox}{normal}{#2}{#3}}
	{\end{titlebox}
\end{strip}}

\newenvironment{protocolbox*}[3]
{\begin{strip}
\begin{titlebox}{Protocol \normalfont #1}{commonbox}{normal}{#2}{#3}}
	{\end{titlebox}
\end{strip}}

\newenvironment{algobox*}[3]
{\begin{strip}
\begin{titlebox}{Algorithm \normalfont #1}{commonbox}{normal}{#2}{#3}}
	{\end{titlebox}
\end{strip}}

\newenvironment{reductionbox*}[3]
{\begin{strip}
\begin{titlebox}{Reduction \normalfont #1}{commonbox}{normal}{#2}{#3}}
	{\end{titlebox}
\end{strip}}

\newenvironment{gamebox*}[3]
{\begin{strip}
\begin{titlebox}{Game \normalfont #1}{commonbox}{gray}{#2}{#3}}
	{\end{titlebox}
\end{strip}}

\newenvironment{simulatorbox*}[3]
{\begin{strip}
\begin{titlebox}{Simulator \normalfont #1}{commonbox}{normal}{#2}{#3}}
	{\end{titlebox}
\end{strip}}

\mdfdefinestyle{mycommonbox}{%
	align=center, middlelinewidth=1.1pt,userdefinedwidth=\linewidth
	innerrightmargin=0pt,innerleftmargin=2pt,innertopmargin=3pt,
	splittopskip=7pt,splitbottomskip=3pt
}
\mdfdefinestyle{myroundbox}{style=mycommonbox,roundcorner=5pt,userdefinedwidth=\linewidth}

\newenvironment{titlebox*}[5]
{\mdfsetup{
		style=#2,
		innertopmargin=0.3\baselineskip,
		skipabove={0.4em},
		skipbelow={1em},needspace=3\baselineskip,
		frametitleaboveskip=5em,
		innerrightmargin=5pt
	}
	\newcommand{\TitleCaption}{#4}
	\newcommand{\TitleLabel}{#5}
	\begin{mdframed}[font=\small]
		\setlist[itemize]{leftmargin=13pt}\setlist[enumerate]{leftmargin=13pt}\raggedright%
	}
	{\end{mdframed}
	\vspace{-2em}
	{\captionof{figure}{\normalfont \TitleCaption}\label{\TitleLabel}}
	\medskip
}

\newenvironment{myprotocolbox}[3]
{\begin{titlebox*}{Protocol \normalfont #1}{mycommonbox}{normal}{#2}{#3}}
	{\end{titlebox*}}

\newenvironment{mysystembox*}[3]
{\begin{strip}
		\vspace{\baselineskip}\begin{titlebox*}{Functionality \normalfont #1}{myroundbox}{normal}{#2}{#3}}
		{\end{titlebox*}
\end{strip}}

\newenvironment{mygsystembox*}[3]
{\begin{strip}
		\vspace{\baselineskip}\begin{titlebox*}{Global Functionality \normalfont #1}{myroundbox}{normal}{#2}{#3}}
		{\end{titlebox*}
\end{strip}}

\newenvironment{myprotocolbox*}[3]
{\begin{strip}
		\begin{titlebox*}{Protocol \normalfont #1}{mycommonbox}{normal}{#2}{#3}}
		{\end{titlebox*}
\end{strip}}

\newenvironment{myalgobox*}[3]
{\begin{strip}
		\begin{titlebox*}{Algorithm \normalfont #1}{mycommonbox}{normal}{#2}{#3}}
		{\end{titlebox*}
\end{strip}}

\newenvironment{myreductionbox*}[3]
{\begin{strip}
		\begin{titlebox*}{Reduction \normalfont #1}{mycommonbox}{normal}{#2}{#3}}
		{\end{titlebox*}
\end{strip}}

\newenvironment{mygamebox*}[3]
{\begin{strip}
		\begin{titlebox*}{Game \normalfont #1}{mycommonbox}{gray}{#2}{#3}}
		{\end{titlebox*}
\end{strip}}

\newenvironment{mysimulatorbox*}[3]
{\begin{strip}
		\begin{titlebox*}{Simulator \normalfont #1}{mycommonbox}{normal}{#2}{#3}}
		{\end{titlebox*}
\end{strip}}

\newcommand{\algoHead}[1]{\vspace{0.2em} \underline{\textbf{#1}} \vspace{0.3em}}

\makeatletter
\algnewcommand{\ExtendedState}[1]{\State
	\parbox[t]{\dimexpr\linewidth-\ALG@thistlm}{\hangindent=\algorithmicindent\strut\hangafter=3#1\strut}}
\makeatother

\algnewcommand\algorithmicinput{\textbf{Input:}}
\algnewcommand\Input{\item[\algorithmicinput]}

\algrenewcommand{\algorithmiccomment}[1]{{\color{gray}// #1}}

\let\emptyset\varnothing

\newcommand{\ckt}{\ensuremath{\mathsf{ckt}}}
\newcommand{\gate}{\mathsf{g}}
\newcommand{\wire}{\mathsf{w}}
\newcommand{\PS}{\ensuremath{\mathsf{A}}} 
\newcommand{\IS}{\ensuremath{\mathsf{I}}}
\newcommand{\OS}{\ensuremath{\mathsf{O}}}  
\newcommand{\MS}{\ensuremath{\mathsf{M}}} 
\newcommand{\DF}{\ensuremath{\mathsf{D}}}

\newcommand{\wv}{\mathsf{v}}
\newcommand{\wx}{\mathsf{x}}
\newcommand{\wy}{\mathsf{y}}
\newcommand{\wz}{\mathsf{z}}

\newcommand{\Wv}{\wire_{\wv}}
\newcommand{\Wx}{\wire_{\wx}}
\newcommand{\Wy}{\wire_{\wy}}
\newcommand{\Wz}{\wire_{\wz}}
\newcommand{\Wxyz}{\ensuremath{\Wx, \Wy, \Wz}}

\newcommand{\addop}{\ensuremath{\texttt{+}}}
\newcommand{\multop}{\ensuremath{\times}}
\newcommand{\addgate}{\ensuremath{\gate = (\Wxyz, \addop)}}
\newcommand{\multgate}{\ensuremath{\gate = (\Wxyz, \multop)}}

\newcommand{\Adder}{\ensuremath{\mathsf{Add}}}
\newcommand{\Sub}{\ensuremath{\mathsf{Sub}}}

\newcommand{\negl}{\ensuremath{\mathsf{negl}}}
\newcommand{\csec}{\kappa}
\newcommand{\sparam}{\ensuremath{s}}
\newcommand{\abort}{\ensuremath{\mathtt{abort}}}

\newcommand{\continue}{\ensuremath{\mathtt{continue}}}
\newcommand{\flag}{\ensuremath{\mathsf{flag}}}

\newcommand{\Partyset}{\ensuremath{\mathcal{P}}}

\newcommand{\Adv}{\ensuremath{\mathcal{A}}}
\newcommand{\Sim}{\ensuremath{\mathcal{S}}}

\newcommand{\Hash}{\ensuremath{\mathsf{H}}}

\newcommand{\rtt}{\ensuremath{\mathsf{rtt}}}

\newcommand{\SELECT}{\ensuremath{\mathsf{select}}}
\newcommand{\INPUT}{\ensuremath{\mathsf{Input}}}
\newcommand{\OUTPUT}{\ensuremath{\mathsf{Output}}}

\newcommand{\ESet}{\ensuremath{P_1, P_2, P_3}}		
\newcommand{\EInSet}{\ensuremath{ \{1, 2, 3\}}}		
\newcommand{\PSet}{\ensuremath{P_0, P_1, P_2, P_3}}	
\newcommand{\PInSet}{\ensuremath{\{0, 1, 2, 3\}}}	

\newcommand{\val}{\ensuremath{\mathsf{v}}} 
\newcommand{\bitb}{\ensuremath{\mathsf{b}}} 
\newcommand{\pad}{\ensuremath{\mathsf{\lambda}}} 
\newcommand{\mask}{\ensuremath{\mathsf{m}}}

\newcommand{\sqd}{\ensuremath{\left[\cdot\right]}}
\newcommand{\sqr}[1]{\ensuremath{\left[#1\right]}}
\newcommand{\sqrA}[1]{\ensuremath{\left[#1\right]_{1}}}
\newcommand{\sqrB}[1]{\ensuremath{\left[#1\right]_{2}}}
\newcommand{\sqrC}[1]{\ensuremath{\left[#1\right]_{3}}}

\newcommand{\sgrd}{\ensuremath{\langle \cdot \rangle}}
\newcommand{\sgr}[1]{\ensuremath{\langle #1 \rangle}}

\newcommand{\shrd}{\ensuremath{\llbracket \cdot \rrbracket}}
\newcommand{\shr}[1]{\ensuremath{\llbracket #1 \rrbracket}}
\newcommand{\Mask}[1]{\ensuremath{\mask_{#1}}}
\newcommand{\Pad}[1]{\ensuremath{\pad_{#1}}}
\newcommand{\PadA}[1]{\ensuremath{\pad_{#1,1}}}
\newcommand{\PadB}[1]{\ensuremath{\pad_{#1,2}}}
\newcommand{\PadC}[1]{\ensuremath{\pad_{#1,3}}}
\newcommand{\PadV}[2]{\ensuremath{\pad_{#1,#2}}}

\newcommand{\shareA}[1]{\ensuremath{\llbracket #1 \rrbracket}^{\bf A}}
\newcommand{\shareB}[1]{\ensuremath{\llbracket #1 \rrbracket}^{\bf B}}

\newcommand{\Gammaxy}{\ensuremath{\gamma_{\wx \wy}}}
\newcommand{\GammaxyA}{\ensuremath{\gamma_{\wx \wy,1}}}
\newcommand{\GammaxyB}{\ensuremath{\gamma_{\wx \wy,2}}}
\newcommand{\GammaxyC}{\ensuremath{\gamma_{\wx \wy,3}}}

\newcommand{\GammaxyjA}{\ensuremath{\gamma_{{\wx}_j {\wy}_j,1}}}
\newcommand{\GammaxyjB}{\ensuremath{\gamma_{{\wx}_j {\wy}_j,2}}}
\newcommand{\GammaxyjC}{\ensuremath{\gamma_{{\wx}_j {\wy}_j,3}}}

\newcommand{\GammaxyV}[1]{\ensuremath{\gamma_{\wx \wy,#1}}}

\newcommand{\commit}{\ensuremath{\mathsf{Com}}}
\newcommand{\Commit}[1]{\ensuremath{\commit(#1)}}

\newcommand{\Key}[1]{\ensuremath{\mathsf{k}_{#1}}}

\newcommand{\Sh}{\ensuremath{\mathsf{Sh}}}
\newcommand{\aSh}{\ensuremath{\mathsf{aSh}}}
\newcommand{\Rec}{\ensuremath{\mathsf{Rec}}}
\newcommand{\fRec}{\ensuremath{\mathsf{fRec}}}
\newcommand{\Add}{\ensuremath{\mathsf{Add}}}
\newcommand{\Mult}{\ensuremath{\mathsf{Mult}}}
\newcommand{\Zero}{\ensuremath{\mathsf{Zero}}}
\newcommand{\FourPC}{\ensuremath{\mathsf{4PC}}}

\newcommand{\piSh}{\ensuremath{\Pi_{\Sh}}}
\newcommand{\piaSh}{\ensuremath{\Pi_{\aSh}}}
\newcommand{\piRec}{\ensuremath{\Pi_{\Rec}}}
\newcommand{\pifRec}{\ensuremath{\Pi_{\fRec}}}
\newcommand{\piAdd}{\ensuremath{\Pi_{\Add}}}
\newcommand{\piMult}{\ensuremath{\Pi_{\Mult}}}
\newcommand{\piZero}{\ensuremath{\Pi_{\Zero}}}
\newcommand{\piFourPC}{\ensuremath{\Pi_{\FourPC}}}

\newcommand{\shrdG}{\ensuremath{\llbracket \cdot \rrbracket}^{\bf G}}
\newcommand{\shareG}[1]{\ensuremath{\llbracket #1 \rrbracket}^{\bf G}}
\newcommand{\shareGP}[2]{\ensuremath{\llbracket #1 \rrbracket}^{\bf G}_{P_{#2}}}

\newcommand{\RG}{\ensuremath{\mathsf{R}}}

\newcommand{\KeyGA}[1]{\ensuremath{\mathsf{K}_{#1}^{0}}}
\newcommand{\KeyGB}[1]{\ensuremath{\mathsf{K}_{#1}^{1}}}
\newcommand{\KeyGAct}[1]{\ensuremath{\mathsf{K}_{#1}^{#1}}}

\newcommand{\piShG}{\ensuremath{\Pi_{\Sh}^{\bf G}}}

\newcommand{\GarA}{\ensuremath{\mathsf{G1}}}
\newcommand{\GarB}{\ensuremath{\mathsf{G2}}}
\newcommand{\GarC}{\ensuremath{\mathsf{G3}}}
\newcommand{\GarD}{\ensuremath{\mathsf{G4}}}

\newcommand{\Decode}{\ensuremath{\mathsf{Decode}}}
\newcommand{\Garb}[1]{\ensuremath{\mathsf{Gar(#1)}}}

\newcommand{\vSh}{\ensuremath{\mathsf{vSh}}}
\newcommand{\DotP}{\ensuremath{\mathsf{DotP}}}

\newcommand{\pivSh}{\ensuremath{\Pi_{\vSh}}}
\newcommand{\pivShA}{\ensuremath{\Pi_{\vSh}^{\bf A}}}
\newcommand{\pivShB}{\ensuremath{\Pi_{\vSh}^{\bf B}}}
\newcommand{\pivShG}{\ensuremath{\Pi_{\vSh}^{\bf G}}}
\newcommand{\piDotP}{\ensuremath{\Pi_{\DotP}}}

\newcommand{\vc}{\ensuremath{\mathsf{c}}}
\newcommand{\vp}{\ensuremath{\mathsf{p}}}
\newcommand{\vq}{\ensuremath{\mathsf{q}}}
\newcommand{\vr}{\ensuremath{\mathsf{r}}}
\newcommand{\vu}{\ensuremath{\mathsf{u}}}
\newcommand{\vv}{\ensuremath{\mathsf{v}}}
\newcommand{\vx}{\ensuremath{\mathsf{x}}}
\newcommand{\vy}{\ensuremath{\mathsf{y}}}
\newcommand{\vz}{\ensuremath{\mathsf{z}}}

\newcommand{\GB}{\textsf{G2B}}
\newcommand{\BG}{\textsf{B2G}}
\newcommand{\GA}{\textsf{G2A}}
\newcommand{\AG}{\textsf{A2G}}
\newcommand{\AB}{\textsf{A2B}}
\newcommand{\BA}{\textsf{B2A}}
\newcommand{\BitA}{\textsf{Bit2A}}
\newcommand{\BitInj}{\textsf{BitInj}}

\newcommand{\PiGB}{\ensuremath{\Pi_{\GB}}}
\newcommand{\PiBG}{\ensuremath{\Pi_{\BG}}}
\newcommand{\PiGA}{\ensuremath{\Pi_{\GA}}}
\newcommand{\PiAG}{\ensuremath{\Pi_{\AG}}}
\newcommand{\PiAB}{\ensuremath{\Pi_{\AB}}}
\newcommand{\PiBA}{\ensuremath{\Pi_{\BA}}}
\newcommand{\PiBitA}{\ensuremath{\Pi_{\BitA}}}
\newcommand{\PiBitInj}{\ensuremath{\Pi_{\BitInj}}}

\newcommand{\vecP}{\ensuremath{\vec{\mathbf{p}}}}
\newcommand{\vecQ}{\ensuremath{\vec{\mathbf{q}}}}
\newcommand{\vecW}{\ensuremath{\vec{\mathbf{w}}}}
\newcommand{\vecX}{\ensuremath{\vec{\mathbf{x}}}}
\newcommand{\vecY}{\ensuremath{\vec{\mathbf{y}}}}

\newcommand{\Mat}[1]{\ensuremath{\mathbf{#1}}}

\newcommand{\vd}{\ensuremath{\mathsf{d}}}
\newcommand{\vm}{\ensuremath{\mathsf{m}}}
\newcommand{\vt}{\ensuremath{\mathsf{t}}}
\newcommand{\vrt}{\ensuremath{\mathsf{r}^\mathsf{t}}}
\newcommand{\maxv}{\ensuremath{\mathsf{max}}}
\newcommand{\MSB}{\ensuremath{\mathsf{msb}}}

\newcommand{\MultTr}{\ensuremath{\mathsf{MultTr}}}
\newcommand{\BitExt}{\ensuremath{\mathsf{BitExt}}}
\newcommand{\ReLU}{\ensuremath{\mathsf{relu}}}
\newcommand{\dReLU}{\ensuremath{\mathsf{drelu}}}
\newcommand{\Sig}{\ensuremath{\mathsf{sig}}}

\newcommand{\piMultTr}{\ensuremath{\Pi_{\MultTr}}}
\newcommand{\piBitExt}{\ensuremath{\Pi_{\BitExt}}}
\newcommand{\piReLU}{\ensuremath{\Pi_{\ReLU}}}

\newcommand{\piSig}{\ensuremath{\Pi_{\Sig}}}

\newcommand{\FSETUP}{\ensuremath{\mathcal{F}_{\mathsf{setup}}}} 
\newcommand{\FZERO}{\ensuremath{\mathcal{F}_{\mathsf{Zero}}}} 
\newcommand{\FFOURPC}{\ensuremath{\mathcal{F}_{\mathsf{\FourPC}}}}
\newcommand{\FSh}{\ensuremath{\mathcal{F}_{\Sh}}} 
 
\newcommand{\FvSh}{\ensuremath{\mathcal{F}_{\vSh}}} 
\newcommand{\FRec}{\ensuremath{\mathcal{F}_{\Rec}}} 
\newcommand{\FMult}{\ensuremath{\mathcal{F}_{\Mult}}} 

\newcommand{\FDotP}{\ensuremath{\mathcal{F}_{\DotP}}}
\newcommand{\FBitA}{\ensuremath{\mathcal{F}_{\BitA}}}
\newcommand{\FBitInj}{\ensuremath{\mathcal{F}_{\BitInj}}}
\newcommand{\FMultTr}{\ensuremath{\mathcal{F}_{\MultTr}}} 
\newcommand{\FBitExt}{\ensuremath{\mathcal{F}_{\BitExt}}} 
\newcommand{\SimFourPC}{\ensuremath{{\mathcal S}_{\FourPC}}}
\pgfplotstableread[col sep=comma]{
	Name,    Ours,     ABY3
	Boston,  106667,  4077 
	Weather, 106667,  1743
	CalCOFI,   106667,  735
}\linregtable

\pgfplotstableread[col sep=comma]{
	Name,      Ours,     ABY3
	Candy,     12549,  2172 
	Epileptic, 12549,  290
	Recipe,    12549,  79
}\logregtable

\begin{document}
\date{}
\title{Trident: Efficient 4PC Framework for Privacy Preserving Machine Learning}
\author{
	\IEEEauthorblockN{Harsh Chaudhari\IEEEauthorrefmark{1},
		Rahul Rachuri\IEEEauthorrefmark{2}
		\thanks{This work was done when Rahul Rachuri was at IIIT-B, India.}, 
		Ajith Suresh\IEEEauthorrefmark{1}\textsuperscript{\textsection}}
	\IEEEauthorblockA{\IEEEauthorrefmark{1}Indian Institute of Science, Bangalore, Email: \{chaudharim, ajith\}@iisc.ac.in}
	\IEEEauthorblockA{\IEEEauthorrefmark{2}Aarhus University, Denmark, Email: rachuri@cs.au.dk}
}
\maketitle
\begingroup\renewcommand\thefootnote{\textsection}
\footnotetext{Corresponding Author.}
\endgroup
\begin{abstract}
Machine learning has started to be deployed in fields such as healthcare and finance, which involves dealing with a lot of sensitive data. This propelled the need for and growth of privacy-preserving machine learning (PPML). We propose an actively secure four-party protocol (4PC), and a framework for PPML, showcasing its applications on four of the most widely-known machine learning algorithms -- Linear Regression, Logistic Regression, Neural Networks, and Convolutional Neural Networks.

Our 4PC protocol tolerating at most one malicious corruption is practically more efficient than Gordon et al. (ASIACRYPT 2018) as the 4th party in our protocol is not active in the online phase, except input sharing and output reconstruction stages. Concretely, we reduce the online communication as compared to them by 1 ring element. We use the protocol to build an efficient mixed-world framework (Trident) to switch between the Arithmetic, Boolean, and Garbled worlds. Our framework operates in the offline-online paradigm over rings and is instantiated in an outsourced setting for machine learning, where the data is secretly shared among the servers. Also, we propose conversions especially relevant to privacy-preserving machine learning. With the privilege of having an extra honest party, we outperform the current state-of-the-art  ABY3 (for three parties), in terms of both rounds as well as communication complexity.

The highlights of our framework include using minimal number of expensive circuits overall as compared to ABY3. This can be seen in our technique for truncation, which does not affect the online cost of multiplication and removes the need for any circuits in the offline phase. Our B2A conversion has an improvement of $\mathbf{7} \times$ in rounds and $\mathbf{18} \times$ in the communication complexity. 

The practicality of our framework is argued through improvements in the benchmarking of the aforementioned algorithms when compared with ABY3. All the protocols are implemented over a 64-bit ring in both LAN and WAN settings. Our improvements go up to $\mathbf{187} \times$ for the training phase and $\mathbf{158} \times$ for the prediction phase when observed over LAN and WAN.
\end{abstract}

\noindent \textcolor{darkred}{Update: An improved version of this framework appears at~\cite{EPRINT:KPRS21} \url{https://arxiv.org/abs/2106.02850}}

\section{Introduction}
\label{sec:intro}
Machine learning is one of the fastest-growing research domains today. Applications for machine learning range from smarter keyboard predictions to better object detection in self-driving cars to avoid collisions. This is in part due to more data being made available with the rise of internet companies such as Google and Amazon, as well as due to the machine learning algorithms themselves getting more robust and accurate. In fact, machine learning algorithms have now started to beat humans at some complicated tasks such as classifying echocardiograms~\cite{MadaniAMA17}, and they are only getting better. Techniques such as deep learning and reinforcement learning are at the forefront making such breakthroughs possible.

The level of accuracy and robustness required is very high to operate in mission-critical fields such as healthcare, where the functioning of the model is vital to the working of the system. Accuracy and robustness are governed by two factors, one of them is the high amount of computing power demanded to train deep learning models. The other factor influencing the accuracy of the model is the variance in the dataset. Variance in datasets comes from collecting data from multiple diverse sources, which is typically infeasible for a single company to achieve.

Towards this, companies such as Microsoft (Azure), Amazon (AWS), Google (Google Cloud), etc. have entered into space by offering "Machine Learning as a Service (MLaaS)". MLaaS works in two different ways, depending on the end-user. The first scenario is companies offering their trained machine learning models that a customer can query to obtain the prediction result. The second scenario is when multiple customers/companies want to come together and train a common model using their datasets, but none of them wish to share the data in the clear. 
While promising, both models require the end-user to make compromises. In the case of an individual customer, privacy of his/her query is not maintained and in the case of companies, policies like the European Union General Data Protection Regulation (GDPR) or the EFF's call for information fiduciary rules for businesses have made it hard and often illegal for companies to share datasets with each other without prior consent of the customers, security, and other criteria met. Even with all these criteria met, data is proprietary information of a company which they would not want to share due to concerns such as competitive advantage.

Due to the huge interest in using machine learning, the field of privacy-preserving machine learning (PPML) has become a fast-growing area of research that addresses the aforementioned problems through techniques for privacy-preserving training and prediction. These techniques when deployed ensure that no information about the query or the datasets is leaked beyond what is permissible by the algorithm, which in some cases might be only the prediction output. Recently there have been a slew of works that have used the techniques of Secure Multiparty Computation (MPC) to perform efficient PPML, works such as \cite{MohasselZ17, MakriRSV17,RiaziWTS0K18, MR18, WaghGC18} making huge contributions.

Secure multiparty computation is an area of extensive research that allows for $n$ mutually distrusting parties to perform computations together on their private inputs, such that no coalition of $t$ parties, controlled by an adversary, can learn any information beyond what is already known and permissible by the algorithm. While MPC has been shown to be practical~\cite{PinkasSSW09, Orlandi11,  ArcherBLKNPSW18}, MPC for a small number of parties in the {\em honest majority} setting ~\cite{AFLNO16,FLNW17,ABFLLNOWW17,LN17,CGHIKLN18,NV18,MRZ15,IshaiKKP15,PatraR18,ByaliJPR18,NV18} has become popular over the last few years due to applications such as financial data analysis~\cite{BogdanovTW12}, email spam filtering~\cite{LaunchburyADM14}, distributed credential encryption\cite{MRZ15}, privacy-preserving statistical studies~\cite{BogdanovKKRST15} that involve only a few parties. This is also evident from popular MPC frameworks such as Sharemind~\cite{BogdanovLW08} and  VIFF~\cite{Gei07}.

{\em Our Setting:}
In this work we deal with the specific case of MPC with 4 parties (4PC), tolerating at most 1 malicious corruption. The state-of-the-art three-party (3PC) PPML frameworks in the honest majority setting such as ABY3~\cite{MR18}, SecureNN~\cite{WaghGC18}, and ASTRA~\cite{CCPS19} (prediction only) have fast and efficient protocols for the semi-honest case but are significantly slower when it comes to the malicious setting. This is primarily due to the underlying operations such as Dot Product, Secure Comparison, and Truncation being more expensive in the malicious setting. For instance, the Dot Product protocol of ABY3 incurs communication cost that is linearly dependent on the size of the underlying vector. Since these operations are performed many times, especially during the training phase, more efficient protocols for these operations are crucial in building a better PPML framework. 

The motivation behind our 4PC setting is to investigate the performance improvement, both theoretical and practical, over the existing solutions in the 3PC setting, when given the privilege of an additional honest party. We show later in this work that having an extra honest party helps us achieve simpler and much more efficient protocols as compared to 3PC. For instance, operating in 4PC eliminates the need for expensive multiplication triples and allows us to perform a dot product at a cost that is independent of the size of the two vectors. 

Our ML constructions are built on a new 4PC scheme, instead of the one proposed by Gordon et al.~\cite{GordonR018}, primarily due to the following reasons:
\begin{description}
	\item[1)] Our protocol requires only three out of the four parties to be active during most of the online phase. On the contrary, \cite{GordonR018} demands all the four parties to be active during the online phase. Thus, our protocol is more efficient in the setting where the computation is outsourced to a set of servers.
	\smallskip
	\item[2)] Using the new secret sharing scheme, our protocol shifts $25\%$ ($1$ ring element) of the online communication to the offline phase, thus improving the online efficiency.
\end{description} 

While our setting is more communication efficient, we assume the presence of an extra honest party which demands an additional 3 pairwise authentic channels when compared to that of 3PC. However, {\em monetary cost}~\cite{PinkasRTY19} is an important parameter to look at since the servers need to be running for a long time for complex ML models. The time servers run for and the compute power of the servers dictate the cost of operation. As the fourth party in our framework does not have to be online throughout the online phase, we can shut the server down for most of the online phase. Aided by this fact, the total monetary cost, which would be the total cost of hiring 4 servers to run our framework for either the training or the prediction phase of an algorithm, we come out ahead of ABY3, primarily because the total running time of the servers in our framework is much lower. More details about monetary cost are presented in Appendix~\ref{app:Bench}.

{\em Offline-online paradigm:} 
To improve efficiency, a class of MPC protocols operates in the {\em offline-online} paradigm~\cite{Bea91}. Data-independent computations are carried out in the offline phase, doing so paves way for a fast and efficient online phase of the protocol. Moreover, since the computations performed in the offline phase are data-independent, not all the parties need to be active throughout this phase, placing less reliance on each party. This paradigm has proved its ability to improve the efficiency of protocols in both theoretical \cite{Bea91,Bea95,BH06,BH08,BFO12,CP17}  and practical  \cite{DPSZ12,SPDZ2,SPDZ3,KOS16,BaumDTZ16,DamgardOS17,CramerDESX18,RiaziWTS0K18,KellerPR18} domains. It is especially useful in a scenario like MLaaS, where the same functions need to be performed many times and the function descriptions are known beforehand. Furthermore, we operate in the outsourced setting of MPC, which allows for an arbitrary number of parties to come together and perform their joint computation via a set of servers. Each server can be thought of as a representative for a subset of data owners, or as an independent party. The advantage of this setting is that it allows the framework to easily scale for a large number of parties and the security notions reduce to that of a standard 4PC between the servers.

{\em Rings vs Fields:} 
In the pursuit of practical efficiency, protocols in MPC that operate over rings are preferred to ones that work over finite fields. This is because of the way computations are carried out in standard 32/64-bit CPUs. Since these architectures have been around for a while, many algorithms are optimized for them. Moreover, operating over rings means that we do not have to override basic operations such as addition and multiplication, unlike with finite fields.

Although MPC techniques have been making a lot of progress towards being practically efficient, we cannot directly use the current best MPC protocols to perform PPML. This is mainly due to two reasons, which are:
\begin{description}
	\item[1)] MPC techniques operate in three different worlds -- Arithmetic, Boolean, and Garbled. Each of these worlds is naturally better suited to carry out certain types of computations. For example, the Arithmetic domain (over a ring $\Z{\ell}$) is more suited to perform addition whereas the Garbled world is more suited to perform division. Activation functions used in machine learning, such as Rectified Linear Unit (ReLU), have operations that alternate between multiplications and comparisons. Operating in only one of the worlds, as most of the current MPC techniques do, does not give us the maximum possible efficiency. The mixed protocol framework for MPC was first shown to be practical by TASTY~\cite{HeneckaKSSW10}, which combined Garbled circuits and homomorphic encryption. The idea was later applied to the ML domain by SecureML~\cite{MohasselZ17}, ABY3~\cite{MR18} etc., where protocols to switch between the three worlds were proposed. These mixed world frameworks have proven to be orders of magnitude more efficient than operating in a single world.
	\smallskip
	\item[2)] Since most of the computations and intermediate values in machine learning are decimal numbers, we embed them over a ring by allocating the least significant bits to the fractional part. But several multiplications performed may lead to an overflow. A naive solution to avoid this is to use a large ring to accommodate a fixed number of multiplications, but the number of multiplications for machine learning varies based on the algorithm, making this infeasible. SecureML tackled this problem through truncation, which approximates the value by sacrificing the accuracy by an infinitesimal amount, performed after every multiplication. This technique, however, does not extend into the 3PC or 4PC setting, due to the attack described in ABY3, requiring us to come up with new techniques.
\end{description}
Frameworks such as SecureML and ABY3 have tackled both these issues in the honest majority setting by proposing ways to switch between the three worlds efficiently, as well as efficient ways to do truncation. ABY3 is a lot more efficient than SecureML, in large part due to the 3PC primitives it uses. But ABY3 cannot avoid some expensive operations such as evaluation of a Ripple Carry Adder (RCA) in its truncation and activation functions. Truncation and activation functions -- ReLU and Sigmoid, need rounds proportional to the underlying ring size in ABY3. This gives a lot of scope for improvement in the efficiency, which we achieve through our 4PC framework.

\subsection{Our Contribution}
We propose an efficient framework for mixed world computations in the four-party honest majority setting with active security over the ring $\Z{\ell}$. Our protocols are optimized for PPML and follow the offline-online paradigm. Our improvements come from having an additional honest party in the protocol. Our contributions can be summed up as follows:
\begin{description}
	\item[1)] {\em Efficient 4PC Protocol}: We propose an efficient four-party protocol with active security which proceeds through a masked evaluation inspired by Gordon et al.~\cite{GordonR018}. Our protocol requires $3$ ring elements in the online phase per multiplication as opposed to $4$ of ~\cite{GordonR018}, achieving a $25\%$ improvement. This improvement is achieved by not compromising on the total cost ($6$ ring elements).  Another significant advantage of our protocol is that the fourth party is not required for evaluation in the online phase. This is not the case with ~\cite{GordonR018}, where all the parties need to be online throughout the protocol execution. In addition to the stated contributions, our framework also achieves fairness without affecting the complexity of a multiplication gate.
	\begin{table}[htb!]
		\centering
		\resizebox{.38\textwidth}{!}{
			\begin{tabular}{c|l|r|r}
				\toprule
				Conversion & Ref. & Rounds & Communication \\
				\midrule
				\multirow{2}{*}{$\GB$} 
				& ABY3        & $1$ & $\kappa$\\
				& {\bf This}  & $1$ & $3$ \\
				\midrule
				\multirow{2}{*}{$\GA$} 
				& ABY3        & $1$ & $2\ell \kappa$\\
				& {\bf This}  & $1$ & $3 \ell$\\
				\midrule
				\multirow{2}{*}{$\BG$} 
				& ABY3        & $1$ & $2\kappa$\\
				& {\bf This}  & $1$ & $\kappa$ \\
				\midrule
				\multirow{2}{*}{$\AG$} 
				& ABY3        & $1$ & $2\ell \kappa$\\
				& {\bf This}  & $1$ & $\ell \kappa$\\
				\midrule
				\multirow{2}{*}{$\AB$} 
				& ABY3        & $1 + \log\ell$    & $9\ell \log\ell + 9\ell$\\
				& {\bf This}  & $1 + \log \ell$   & $3 \ell \log \ell + \ell$\\
				\midrule
				\multirow{2}{*}{$\BA$} 
				& ABY3        & $1 + \log\ell$  & $9\ell \log\ell + 9\ell$\\
				& {\bf This}  & $1$ & $3 \ell$\\
				\bottomrule
			\end{tabular}
		}
		\caption{\small Online cost of share conversions of ABY3\cite{MR18} and {\bf This} work. $\ell$ denotes the size of underlying ring in bits and $\kappa$ denotes the computational security parameter.}\label{tab:ConvIntro}
	\end{table}
	\smallskip
	\item[2)] {\em Fast Mixed World Computation}: We propose a framework -- Trident, that is geared towards a high throughput online phase as compared to the existing alternatives. This throughput is achieved by making use of an additional honest party. Every one of the conversions we propose to switch between the worlds is more efficient in terms of online communication complexity as compared to ABY3, with our improvements ranging from $2\times$ to $2\kappa/3\times$, where $\kappa$ denotes the computational security parameter. More concretely, if we aim for 128-bit computational security, our framework gives a maximum improvement of $\approx 85\times$. For instance, the technique we propose to perform bit composition ($\BA$) requires only $1$ round, as opposed to $1 + \log\ell$ rounds in ABY3, which translates to a $7 \times$ gain for a 64-bit ring. The table below provides the concrete cost of our online phase in comparison to ABY3. The overall cost comparison can be found in Table~\ref{tab:ConvM}.
	\smallskip
	\item[3)]  {\em Efficient Truncation}: The highlight of the protocol we propose for truncation is that it can be combined with our multiplication protocol with no additional cost in the online phase. In contrast, the online cost for multiplication in ABY3 increases from $9$ to $12$ ring elements, which gives us a $4\times$ improvement in online communication. Moreover, we forgo the need for $(2\ell - 2)$-round Ripple Carry Adders (RCA), as opposed to ABY3, in the offline phase resulting in an improvement of $63 \times$ in rounds for a 64-bit ring.
	\vspace{-2mm}
	\begin{table}[htb!]
		\centering
		\resizebox{.44\textwidth}{!}{
		\begin{tabular}{c|l|r|r}
			\toprule
			Conversion & Ref. & Rounds & Communication \\
			\midrule
			\multirow{2}{*}{\makecell{Multiplication\\with Truncation}} 
			& ABY3        & $1$ & $12\ell$\\
			& {\bf This}  & $1$ & $3 \ell$\\
			\midrule
			\multirow{2}{*}{\makecell{Secure\\Comparison}} 
			& ABY3        & $\log \ell$ & $18 \ell \log \ell$\\
			& {\bf This}  & $3$ & $5 \ell + 2$\\
			\midrule
			\multirow{2}{*}{\makecell{$\BitA$\\$\shareB{b} \rightarrow \shr{b}$}} 
			& ABY3        & $2$ & $18\ell$\\
			& {\bf This}  & $1$ & $3 \ell$\\
			\midrule
			\multirow{2}{*}{\makecell{$\BitInj$\\$\shareB{b}\shr{\val} \rightarrow \shr{b \val}$}} 
			& ABY3        & $3$ & $27\ell$\\
			& {\bf This}  & $1$ & $3 \ell$\\
			\midrule
			\multirow{2}{*}{ReLU} 
			& ABY3        & $3 + \log \ell$ & $45 \ell$\\
			& {\bf This}  & $4$ & $8 \ell + 2$\\
			\midrule
			\multirow{2}{*}{Sigmoid} 
			& ABY3        & $4 + \log \ell$ & $81 \ell + 9$\\
			& {\bf This}  & $5$ & $16 \ell + 7$\\
			\bottomrule
		\end{tabular}
		}
		\caption{\small Online cost of ML conversions of ABY3\cite{MR18} and {\bf This} work. $\ell$ denotes the size of underlying ring in bits.}\label{tab:MLIntro}
	\end{table}
	\vspace{-4mm}
	\item[4)] {\em ML Building Blocks}: The building blocks for ML highlighted in the \tabref{MLIntro} (more details in \tabref{ConvMLM}), have improvements ranging from $2\times$ - $3\times$ in the round complexity and $5\times$ - $9\times$ in the communication complexity. For the activation functions ReLU and Sigmoid, our solution brings down the round complexity from $\BigO{\log\ell}$ (of ABY3) to a constant. 
	\smallskip
	\item[5)] {\em Implementation}: We implement all the stated protocols and test them over LAN and WAN. We benchmark the training and prediction phases of the algorithms -- Linear Regression, Logistic Regression, Neural Networks (NN), and Convolutional Neural networks (CNN). In order to compare with ABY3, we implement their protocols as well and obtain the benchmarks in our environment. For the training phase of Linear Regression, we have improvements across different configurations in the range of $2 \times$ to $251.84 \times$. Similarly, for Logistic Regression, our improvements lie in the range of $2.71 \times$ to $67.88 \times$. The respective range of improvements for NN and CNN are $2.94 \times$-$68.04 \times$ and $3.19 \times$-$45.64 \times$.
	\tabref{GainIntro} gives the concrete gain of the aforementioned algorithms over the most widely used MNIST dataset~\cite{MNIST10}, which has 784 features, implemented with a batch size of 128. Moreover, our framework is able to process $23$ online iterations of NN in a second for a batch size of 128, over LAN. This is a huge improvement over ABY3, which can process only $2.5$ iterations, that too in the semi-honest setting. Similarly, for CNN, we can process $10.46$ iterations as opposed to $2$ of ABY3.
	\vspace{-2mm}
	\begin{table}[htb!]
		\centering
		\resizebox{.4\textwidth}{!}{
		\begin{tabular}{c|r|r|r|r}
			\toprule
			Network & \makecell{Linear\\Regression} & \makecell{Logistic\\Regression} & \makecell{NN} & \makecell{CNN}\\
			\midrule
		    LAN         & $\Mb{81.08}$ & $\Mb{27.07}$ & $\Mb{68.08}$ & $\Mb{45.64}$\\
			WAN         & $\Mb{2.17}$  & $\Mb{2.76}$  & $\Mb{2.97}$  & $\Mb{3.19}$\\
			\bottomrule
		\end{tabular}
		}
		\caption{\small Gain in online throughput for ML Training over ABY3\cite{MR18} for $d=784$ features and batch size of 128.}\label{tab:GainIntro}
	\end{table}
\end{description}

\vspace{-6mm}
We also provide results for the prediction phase and give throughput (no. of predictions per second) comparison details for the aforementioned algorithms, using real-world datasets. The gain in online throughput for prediction ranges from $3\times$ to $145.18 \times$ for Linear Regression and $3 \times$ to $158.40 \times$ for Logistic Regression over LAN and WAN combined. Similarly, the online throughput gain ranges from $335.44\times$ to $421.72 \times$ for NN and $598.44 \times$ to $759.65 \times$ for CNN.
\section{Preliminaries and Definitions}
\label{sec:prelim}

We consider a set of four parties $\Partyset = \{ P_0, P_1, P_2, P_3 \}$ that are connected by pair-wise private and authentic channels in a synchronous network. The function $f$ to be evaluated is expressed as a circuit $\ckt$, whose topology is publicly known and is evaluated over either an arithmetic ring $\Z{\ell}$ or a Boolean ring $\Z{1}$, consisting of $2$-input addition and multiplication gates. 
The term $\DF$ denotes the multiplicative depth of the circuit, while $\IS, \OS, \PS, \MS$ denote the number of input wires, output wires, addition gates and multiplication gates respectively in $\ckt$. 

We use the notation $\Wv$ to denote a wire $\wire$ with value $\wv$ flowing through it. We use $\gate = (\Wxyz, \mathsf{op})$ to denote a gate in the $\ckt$ with left input wire $\Wx$, right input wire $\Wy$, output wire $\Wz$ and operation $\mathsf{op}$, which is either addition ($\addop$) or multiplication ($\multop$). 

For a vector $\vecX$, $\vx_i$ denotes the $i^{th}$ element in the vector. For two vectors $\vecX$ and $\vecY$ of length $\vd$, the dot product is given by, $\vecX \band \vecY = \sum_{i = 1}^{\vd} \vx_i \vy_i$. 
Given two matrices $\Mat{X}, \Mat{Y}$, the operation $\Mat{X} \circ \Mat{Y}$ denotes the matrix multiplication.

\paragraph{Shared Key Setup}
In order to facilitate non-interactive communication, parties use functionality $\FSETUP$ that establishes pre-shared random keys for a pseudo-random function (PRF) among them. Similar setup for the three-party case can be found in \cite{FLNW17, ABFLLNOWW17, RiaziWTS0K18, MR18, CCPS19}. 

In our protocols, we make use of a {\em collision-resistant} hash function, denoted by $\Hash()$, to save communication. We defer the formal details of key setup and hash function to Appendix~\ref{app:4PC}.   

\section{Our 4PC Protocol}
\label{sec:FourPC}
In this section, we provide details for our 4PC protocol. We begin with the sharing semantics in \secref{sematics} followed by explaining the relevant building blocks in \secref{build_blocks}. We elaborate on the stages of our protocol in \secref{stages_4pc}. Lastly, in \secref{fair4pc}, we show how to improve the security to achieve fairness.

\subsection{Sharing Semantics}
\label{sec:sematics}
In this section, we explain three variants of secret sharing that are used in this work. The sharings work over both arithmetic ($\Z{\ell}$) and boolean ($\Z{1}$) rings.

\paragraph{$\sqd$-sharing}
A value $\val$ is said to be $\sqd$-shared among parties $\ESet$, if the parties $P_1, P_2$ and $P_3$ respectively hold the values $\val_1, \val_2$ and $\val_3$ such that $\val = \val_1 + \val_2 + \val_3$. We use $\sqd_{P_i}$ to denote the $\sqd$-share of  party $P_i$ for $i \in \EInSet$.

\paragraph{$\sgrd$-sharing}
A value $\val$ is said to be $\sgrd$-shared among parties $\ESet$, if the parties $P_1, P_2$ and $P_3$ respectively holds values $(\val_2, \val_3), (\val_3, \val_1)$ and $(\val_1, \val_2)$ such that $\val = \val_1 + \val_2 + \val_3$. We denote $\sgrd$-shares of the parties as follows:
\vspace{-2mm}
\begin{center}
	$\sgr{\val}_{P_1} = (\val_2, \val_3),~\sgr{\val}_{P_2} = (\val_3, \val_1),~\sgr{\val}_{P_3} = (\val_1, \val_2)$
\end{center}

\paragraph{$\shrd$-sharing}
A value $\val$ is said to be $\shrd$-shared among parties $\PSet$, if 
\begin{myitemize}
	\item[--] there exist values $\Pad{\val}, \Mask{\val} \in \Z{\ell}$ such that $\Mask{\val} = \val + \Pad{\val}$.
	\item[--] parties $\ESet$ know the value $\Mask{\val}$ in clear, while the value $\Pad{\val}$ is $\sgrd$-shared among them. 
	\item[--] party $P_0$ knows $\PadA{\val}, \PadB{\val}$ and $\PadC{\val}$ in clear.
\end{myitemize}
We denote the $\shrd$-shares of the parties as follows:
{\small
\begin{center}
		\begin{tabu} to 1\textwidth { l  l }
			$\shr{\val}_{P_0} = (\PadA{\val}, \PadB{\val}, \PadC{\val})$ & 
			$\shr{\val}_{P_1} = (\Mask{\val}, \PadB{\val}, \PadC{\val})$\\
			$\shr{\val}_{P_2} = (\Mask{\val}, \PadC{\val}, \PadA{\val})$ &
			$\shr{\val}_{P_3} = (\Mask{\val}, \PadA{\val}, \PadB{\val})$\\
		\end{tabu}
\end{center}
}
\noindent We use $\shr{\val} = (\Mask{\val}, \sgr{\Pad{\val}})$ to denote the $\shrd$-share of $\val$.

\paragraph{Linearity of the secret sharing schemes}
Given the $\sqd$-sharing of $\wx, \wy$ and public constants $c_1, c_2$, parties can locally compute $\sqr{c_1 \wx + c_2 \wy} = c_1 \sqr{\wx} + c_2 \sqr{\wy}$.
\begin{align*}
\sqr{c_1 \wx + c_2 \wy} &= (c_1 \wx_1 + c_2 \wy_1, c_1 \wx_2 + c_2 \wy_2, c_1 \wx_3 + c_2 \wy_3)\\
&= c_1 \sqr{\wx} + c_2 \sqr{\wy}
\end{align*}

It is easy to see that the linearity trivially extends to $\sgrd$-sharing as well. That is $\sgr{c_1 \wx + c_2 \wy} = c_1 \sgr{\wx} + c_2 \sgr{\wy}$. Similarly, given the $\shrd$-sharing of $\wx, \wy$ and public constants $c_1, c_2$, parties can locally compute $\shr{c_1 \wx + c_2 \wy}$.
Note that the linearity property enables parties to {\em non-interactively} evaluate an addition gate as well as perform the multiplication of their shares with a public constant.

\subsection{Building Blocks}
\label{sec:build_blocks}

\paragraph{Sharing Protocol}
Protocol $\piSh$ (\boxref{fig:piSh}) enables party $P_i$ to generate $\shrd$-share of value $\val$. The offline phase is done using the pre-shared keys in such a way that $P_i$ will get the entire mask $\pad$. During the online phase, $P_i$ computes $\Mask{\val}$ and sends to $\ESet$ who exchange the hash values to check for consistency.
\begin{protocolsplitbox}{$\piSh(P_i, \val)$}{$\shrd$-sharing of a value $\val$ by party $P_i$.}{fig:piSh}
	\justify
	\algoHead{Offline:} 
	\begin{myitemize}
		\item[--] If $P_i = P_0$, parties in $\Partyset\setminus\{P_j\}$ together sample $\PadV{\val}{j}$ for $j \in \EInSet$.
		\item[--] If $P_i = P_k$ for $k \in \EInSet$, parties in $\Partyset$ together sample $\PadV{\val}{k}$. In addition, parties in $\Partyset\setminus\{P_j\}$ together sample $\PadV{\val}{j}$ for $j \in \EInSet \setminus \{k\}$.
	\end{myitemize}
	\justify
	\algoHead{Online:}
	\begin{myitemize}
		\item[--] $P_i$ computes $\Mask{\val} = \val + \Pad{\val}$ and sends to $\ESet$.
		\item[--] $\ESet$ mutually exchange $\Hash(\Mask{\val})$ and $\abort$ if the received values are inconsistent. 
	\end{myitemize}
\end{protocolsplitbox}
\vspace{-3mm}
Looking ahead, we also encounter scenarios where party $P_0$ has to generate $\sgrd$-sharing of a value $\val$ in the offline phase. We call the resultant protocol as $\piaSh$ and the formal details appear in \boxref{fig:piaSh}.
\begin{protocolbox}{$\piaSh(P_0, \val)$}{$\sgrd$-sharing of a value $\val$ by party $P_0$.}{fig:piaSh}
	\justify
	\algoHead{Offline:} 
	\begin{myitemize}
		\item[--] Parties in $\Partyset \setminus \{P_1\}$ sample random $\val_1 \in \Z{\ell}$, while parties in $\Partyset \setminus \{P_2\}$ sample random $\val_2$.
		\item[--] $P_0$ computes $\val_3 = - (\val + \val_1 + \val_2)$ and sends it to both $P_1$ and $P_2$, who exchange $\Hash(\val_3)$ and $\abort$ if there is a mismatch.
	\end{myitemize}
\end{protocolbox}
\vspace{-3mm}
\paragraph{Reconstruction Protocol}
Protocol $\piRec(\Partyset, \val)$ (\boxref{fig:piRec}) enables parties in $\Partyset$ to compute $\val$, given its $\shrd$-share. Towards this, each party receives the missing share from one other party and hash of the missing share from one of the other two parties. If the received shares are consistent, he/she will proceed with the reconstruction. Reconstruction towards a single party can be viewed as a special case of this protocol.
\begin{protocolbox}{$\piRec(\Partyset, \shr{\val})$}{Reconstruction of value $\val$ among parties in $\Partyset$.}{fig:piRec}
	\justify
	\algoHead{Online:}
	\begin{mylist}
		\item[--] $P_1$ receives $\PadA{\val}$ and $\Hash(\PadA{\val})$ from $P_2$ and $P_0$ respectively. 
		\item[--] $P_2$ receives $\PadB{\val}$ and $\Hash(\PadB{\val})$  from $P_3$ and $P_0$ respectively. 
		\item[--] $P_3$ receives $\PadC{\val}$ and $\Hash(\PadC{\val})$ from $P_1$ and $P_0$ respectively. 
		\item[--] $P_0$ receives $\Mask{\val}$ and $\Hash(\Mask{\val})$ from $P_1$ and $P_2$ respectively.
	\end{mylist}
	 $P_i$ for $i \in \PInSet$ $\abort$ if the received values are inconsistent. Else computes $\val = \Mask{\val} - \PadA{\val} - \PadB{\val} - \PadC{\val}$.       
\end{protocolbox}
\vspace{-3mm}

\subsection{Stages of our 4PC protocol}
\label{sec:stages_4pc}
Our protocol $\piFourPC$ consists of three stages, namely -- Input Sharing, Evaluation and Output Reconstruction. We elaborate on each of these stages below.

\paragraph{Input Sharing}
For each wire $\Wv$ holding the value $\wv$, of which $P_{i} \in \Partyset$ is the owner, he/she generates $\shrd$-share of $\wv$ by executing the $\piSh(P_{i}, \wv)$ protocol.

\paragraph{Evaluation}
In this stage, parties evaluate the circuit in a topological order, where the following invariant is maintained for every gate $\gate$: given the inputs of $\gate$ in $\shrd$-shared fashion, the output is generated in the $\shrd$-shared fashion. For the case of an addition gate $\addgate$, the linearity of our sharing scheme maintains this invariant.

For a multiplication gate $\multgate$, the protocol proceeds as follows: during the offline phase, parties $\ESet$ locally compute $\sqd$-shares of $\Gammaxy = \Pad{\wx} \Pad{\wy}$, followed by exchanging them to form a $\sgrd$-sharing of $\Gammaxy$. Before exchanging the shares of $\Gammaxy$, the parties randomize the shares by adding a share of $0$ to the share of $\Gammaxy$ to prevent leakage. In addition, $P_0$ helps the parties in verifying the correctness of shares received in the aforementioned step. During the online phase, the goal is to compute $\Mask{\wz}$. Note that,
\begin{align*}
	\Mask{\wz} &= \wz + \Pad{\wz} = \wx \wy + \Pad{\wz} = (\Mask{\wx} - \Pad{\wx}) (\Mask{\wy} - \Pad{\wy}) + \Pad{\wz}\\
	           &= \Mask{\wx} \Mask{\wy} - \Pad{\wx} \Mask{\wy} - \Pad{\wy} \Mask{\wx} + \Pad{\wx} \Pad{\wy} + \Pad{\wz}
\end{align*} 
Parties $\ESet$ locally compute $\sqd$-share of $\Mask{\wz} - \Mask{\wx} \Mask{\wy}$ followed by an exchange to reconstruct $\Mask{\wz} - \Mask{\wx} \Mask{\wy}$. By the nature of our secret-sharing scheme, every missing share can be computed by two parties. This facilitates the parties to verifiably reconstruct $\Mask{\wz} - \Mask{\wx} \Mask{\wy}$ by having one party send the missing share and the other send a hash of the same.

Each of $\ESet$ locally add $\Mask{\wx} \Mask{\wy}$ to the result to obtain $\Mask{\wz}$. We call the resultant protocol $\piMult$ (\boxref{fig:piMult}).

\begin{protocolbox}{$\piMult(\Wxyz)$}{Multiplication Protocol.}{fig:piMult}
	\justify
	\algoHead{Offline:} 
	\begin{myitemize}
		\item[--] Parties in $\Partyset\setminus\{P_j\}$ together sample $\PadV{\wz}{j}$ for $j \in \EInSet$.
		\item[--] Parties invoke protocol $\piZero$ (\boxref{fig:piZero}) to generate $A, B, \Gamma$ such that $A + B + \Gamma = 0$. Parties locally compute the following:
		\begin{mylist}
			\item[--] $P_0, P_1$ compute $\GammaxyB = \PadB{\wx} \PadB{\wy} + \PadB{\wx} \PadC{\wy} + \PadC{\wx} \PadB{\wy} + A$.
			\item[--] $P_0, P_2$ compute $\GammaxyC = \PadC{\wx} \PadC{\wy} + \PadC{\wx} \PadA{\wy} + \PadA{\wx} \PadC{\wy} + B$.
			\item[--] $P_0, P_3$ compute $\GammaxyA = \PadA{\wx} \PadA{\wy} + \PadA{\wx} \PadB{\wy} + \PadB{\wx} \PadA{\wy} + \Gamma$.
		\end{mylist}
		\item[--] Parties exchange the following:
		\begin{mylist}
			\item[--] $P_1$ receives $\GammaxyC$ and $\Hash(\GammaxyC)$ from $P_2$ and $P_0$ respectively. 
			\item[--] $P_2$ receives $\GammaxyA$ and $\Hash(\GammaxyA)$ from $P_3$ and $P_0$ respectively. 
			\item[--] $P_3$ receives $\GammaxyB$ and $\Hash(\GammaxyB)$ from $P_1$ and $P_0$ respectively. 
		\end{mylist}
		\item[--] $P_i$ for $i \in \EInSet$ $\abort$ if received values are inconsistent.
	\end{myitemize}
	\vspace{-2mm}
	\justify
	\algoHead{Online:} Let $\Mask{\wz}' = \Mask{\wz} - \Mask{\wx} \Mask{\wy}$.
	\begin{myitemize}
		\item[--] Parties locally compute the following:
		\begin{mylist}
			\item[--] $P_1, P_3$ compute $\Mask{\wz,2}' = - \PadB{\wx} \Mask{\wy} - \PadB{\wy} \Mask{\wx} + \GammaxyB + \PadB{\wz}$.
			\item[--] $P_2, P_1$ compute $\Mask{\wz,3}' = - \PadC{\wx} \Mask{\wy} - \PadC{\wy} \Mask{\wx} + \GammaxyC + \PadC{\wz}$.
			\item[--] $P_3, P_2$ compute $\Mask{\wz,1}' = - \PadA{\wx} \Mask{\wy} - \PadA{\wy} \Mask{\wx} + \GammaxyA + \PadA{\wz}$.
		\end{mylist}
		\item[--] Parties exchange the following:
		\begin{mylist}
			\item[--] $P_1$ receives $\Mask{\wz,1}'$ and $\Hash(\Mask{\wz,1}')$ from $P_2$ and $P_3$ respectively. 
			\item[--] $P_2$ receives $\Mask{\wz,2}'$ and $\Hash(\Mask{\wz,2}')$ from $P_3$ and $P_1$ respectively. 
			\item[--] $P_3$ receives $\Mask{\wz,3}'$ and $\Hash(\Mask{\wz,3}')$ from $P_1$ and $P_2$ respectively. 
		\end{mylist}
		\item[--] $P_i$ for $i \in \EInSet$ $\abort$ if the received values are inconsistent. Else, he / she computes $\Mask{\wz} = (\Mask{\wz,1}' + \Mask{\wz,2}' + \Mask{\wz,3}') + \Mask{\wx} \Mask{\wy} = \Mask{\wz}' + \Mask{\wx} \Mask{\wy}$. 
	\end{myitemize}     
\end{protocolbox}

As a very important optimization, note that the exchange of hash values for every multiplication gate can be delayed until the output reconstruction stage. Moreover, all the corresponding values can be appended and hashed, resulting in an overall communication of only $3$ ring elements.

\paragraph{Output Reconstruction}
For each of the output wire $\Wy$ with value $\vy$, parties execute protocol $\piRec(\Partyset, \shr{\vy})$ to reconstruct the output.

\paragraph*{Correctness and Security}
We prove the correctness of $\piFourPC$ below and defer the security details to Appendix~\ref{app:4PCProof}. 
\begin{theorem}[Correctness]
	Protocol $\piFourPC$ is correct.
\end{theorem}
\begin{IEEEproof}
	We claim that for every wire  in $\ckt$, the parties hold a $\shrd$-sharing of the wire value in $\piFourPC$. The correctness for the input and output wires follows from $\piSh$ and $\piRec$ respectively. The claim for addition gates follows from the linearity of $\shrd$-sharing. For a multiplication gate $\multgate$, when evaluated using $\piMult$, the parties receive $\wx \wy + \Pad{\wz}$ in the online phase, resulting in obtaining $\shr{\wz}$. The correctness of $\wx \wy + \Pad{\wz}$ is ensured through the verified reconstruction as shown in $\piMult$.
\end{IEEEproof}

\begin{theorem}[Communication Efficiency]
	\label{thm:Pi4PCCost}
	$\piFourPC$ requires one round with an amortized communication of $3\MS$ ring elements during the offline phase. In the online phase, $\piFourPC$ requires one round with an amortized communication of at most $3\IS$ ring elements in the Input-sharing stage, $\DF$ rounds with an amortized communication of $3\MS$  ring elements for evaluation stage and one round with an amortized communication of $3\OS$  elements for the output-reconstruction stage.
\end{theorem}

\subsection{Achieving Fairness}
\label{sec:fair4pc}
For fairness, we need to ensure that all the parties are {\em alive} in the protocol during the output reconstruction stage. On top of this, we also need to prevent the adversary from mounting a selective abort attack, where he can make some of the honest parties abort the protocol. To achieve this, parties $P_1, P_2, P_3$ set a bit $b$ to $\continue$, if the verification of the multiplication gates was successful, else set it to $\abort$, and send it to $P_0$. $P_0$ then sends $\abort$ back to all the parties if one of the parties sends $\abort$, thus ensuring aliveness. Remaining parties then exchange their reply from $P_0$ and follow the honest-majority in deciding whether to proceed or abort. Since there can be only 1 corruption, all the parties will now be on the same page, preventing a selective abort. If the parties decide to proceed, they exchange the missing shares. Using the fact that there is at most 1 corruption and the structure of our secret-sharing scheme, the most commonly received missing share will be consistent among the honest parties. 

\begin{protocolbox}{$\pifRec(\Partyset, \shr{\val})$}{Fair reconstruction of value $\val$ among parties in $\Partyset$.}{fig:pifRec}
	\justify
	\algoHead{Online:}
	\begin{myitemize}
		\item[--] $\ESet$ set bit $\bitb = \abort$ if the verification for multiplication fails. Else set $\bitb = \continue$.
		\item[--] $\ESet$ send $\bitb$ to $P_0$ who sends back $\abort$, if he/she receives at least one $\abort$ bit. Else sends $\continue$ to $\ESet$.
		\item[--] $\ESet$ mutually exchange the message received from $P_0$. Parties $\abort$ if the majority of the messages received are $\abort$. Else they exchange the missing share as follows:
		\begin{mylist}
			\item[--] $P_0$ receives $\Mask{\val}$ from $P_1, P_2$ and $\Hash(\Mask{\val})$ from $P_3$ respectively.
			\item[--] $P_1$ receives $\PadA{\val}$ from $P_2, P_3$ and $\Hash(\PadA{\val})$ from $P_0$ respectively. 
			\item[--] $P_2$ receives $\PadB{\val}$ from $P_3, P_1$ and $\Hash(\PadB{\val})$ from $P_0$ respectively.
			\item[--] $P_3$ receives $\PadC{\val}$ from $P_1, P_2$ and $\Hash(\PadC{\val})$ from $P_0$ respectively.
		\end{mylist}
		\item[--] $P_i$ for $i \in \PInSet$ chooses the missing share that forms the majority and computes $\val = \Mask{\val} - \PadA{\val} - \PadB{\val} - \PadC{\val}$. 
	\end{myitemize}      
\end{protocolbox}
\section{Mixed Protocol Framework}
\label{sec:MixFrame}
In this section, we present our mixed protocol framework, Trident. Before we go into the details of it, we discuss another world of MPC, called The Garbled World. To evaluate circuits over a ring $\Z{\ell}$, we operate in the arithmetic world and to evaluate boolean circuits ($\Z{\ell}$) we use either the boolean world or the Garbled world, depending on the operation being performed. Superscripts {\{\bf A, B, G\}} are used to indicate the respective worlds. If there is no superscript, the values are assumed to operate in the arithmetic world.

\subsection{The Garbled World}
\label{sec:GWorld}
For the Garbled world, we use the MRZ~\cite{MRZ15} scheme in the 4PC setting. In 4PC, parties $\ESet$ act as the garblers and the party $P_0$ acts as the sole evaluator. As an optimisation, $P_0$ can share his inputs with only $P_1, P_2$ instead of all three parties.  For cross-verification, $P_1$ sends the garbled circuit to $P_0$ while $P_2$ sends a hash of the it to $P_0$. We incorporate the recent optimisations including free XOR~\cite{KS08,KolesnikovMR14}, half-gates~\cite{ZRE15, GLNP15}, fixed-key AES garbling~\cite{BHKR13}. As opposed to the dishonest majority setting, this scheme removes the need for expensive public key primitives (in terms of communication) such as Oblivious Transfers altogether. We present the protocols for a single bit, and each operation can be performed $\ell$ times in parallel to support $\ell$-bit values.

\paragraph{Sharing Semantics}
For a bit $\val$, $\shareG{\val}$ is defined as $\shareG{\val}_{P_i} = \KeyGA{\val} \in \bitset^{\kappa}$ for $i \in \EInSet$ and $\shareG{\val}_{P_0} = \KeyGAct{\val} = \KeyGA{\val} \xor \val \RG$, where $\kappa$ is the computational security parameter.  Here $\RG$ is a global {\it offset} with the least significant bit as one, and is known only to $\ESet$ (generated by shared randomness), and $\RG$ is common across all the $\shrdG$-sharing. It is easy to see that XOR of the least significant bit of $\shareG{\val}_{P_1}$ (resp. $\shareG{\val}_{P_2}, \shareG{\val}_{P_3}$) and $\shareG{\val}_{P_0}$ is $\val$. For a value $\val \in \Z{\ell}$, we abuse the notation $\shareG{\val}$ to denote the set of $\shrdG$-shares of each bit of $\val$. 

\begin{protocolbox}{$\piShG(P_i, \val)$}{$\shrdG$-sharing of $\val$ by $P_i$ for $i \in \EInSet$.}{fig:piShG}
	\justify
	\algoHead{Offline:}
	\begin{myitemize}
		\item[--] $\ESet$ samples a random  $\KeyGA{\val} \in \bitset^{\kappa}$, computes $\KeyGB{\val} = \KeyGA{\val} \xor \RG$ and set  $\shareG{\val}_{P_1} = \shareG{\val}_{P_2} = \shareG{\val}_{P_3} = \KeyGA{\val}$.
		\item[--] $\ESet$ compute commitment of $\KeyGA{\val}, \KeyGB{\val}$. $P_1, P_2$ send the commitments to $P_0$ in a random permuted order, who $\abort$ if the received commitments mismatch.
	\end{myitemize}
	\vspace{-3mm}
	\justify
	\algoHead{Online:}
	\begin{myitemize}
		\item[--] $P_i$ sends $\KeyGA{\val} \oplus \val \RG$ to $P_0$, who sets it as $\shareG{\val}_{P_0}$.
		\item[--] $P_i$ decommits the right key $\KeyGAct{\val}$ to $P_0$, who $\abort$ if the decommitment is incorrect.
	\end{myitemize}
\end{protocolbox}

\paragraph{Input Sharing} 
Protocol $\piShG(P_i, \val)$ enables $P_i$ to generate $\shrdG$-sharing of value $\val$. During the protocol, $P_0$ needs to ensure that it obtains the correct $\KeyGAct{\val}$. To tackle this, we make the garblers commit both keys to $P_0$, who can then verify the correctness by cross-checking the commitments received. The formal details for the case when $P_i$ is one of the garblers appear in \boxref{fig:piShG}.

If $P_i = P_0$, then $\piShG(P_i, \val)$ proceeds as follows: $P_0$ samples random bit $\val_1$, computes $\val_2 = \val \xor \val_1$ and sends $\val_1$ and $\val_2$ to $P_1$ and $P_2$ respectively. Parties execute $\piShG(P_1, \val_1)$ and $\piShG(P_2, \val_2)$ to generate $\shareG{\val_1}$ and $\shareG{\val_2}$ respectively. Parties then locally compute $\shareG{\val} = \shareG{\val_1} \xor \shareG{\val_2}$, using the XOR gate evaluation method via the free-XOR technique. Here, the commitments of the keys need not be permuted, as $P_0$ already knows the actual $\val_1$ and $\val_2$. As an optimization, the computation of $\shareG{\val_1}$ can be offloaded to the offline phase.

\paragraph{Reconstruction}
If $P_i = P_0$, then $P_1, P_2$ send the least significant bit of their shares and $P_i$ verifies if it received the same bit from both $P_1$ and $P_2$. If $P_i$ is one of the garblers, then $P_0$ sends its share to $P_i$. Due to the {\it authenticity} of the underlying garbling scheme \cite{BHR12}, a corrupt $P_0$ cannot send an incorrect share to $P_i$. If there are multiple reconstructions towards $P_i \in \{\ESet\}$, $P_0$ can send the least significant bit of its shares along with a hash of all the corresponding shares.

\paragraph{Operations} 
Let $\vu, \vv \in \bitset$ be $\shrdG$-shared with $P_1, P_2, P_3$ holding the shares $(\KeyGA{\vu}, \KeyGA{\vv})$, and $P_0$ holding the shares $(\KeyGA{\vu} \xor \vu \RG, \KeyGA{\vv} \xor \vv \RG)$. Let $\vc$ denote the output.

\begin{description}
	\item[--] {\em XOR:} The parties locally compute $\shareG{\vc} = \shareG{\vu} \xor \shareG{\vv}$.             
	\item[--] {\em AND:} $\ESet$ sample random $\KeyGA{\vc} \in \bitset^{\kappa}$, compute $\KeyGB{\vu} = \KeyGA{\vu} \xor \RG$ and construct a garbled table for AND using the garbling scheme described in \cite{BHKR13, ZRE15}. $P_1$ sends the garbled table to $P_0$, while $P_2$ sends a hash of the table to $P_0$. $P_0$ evaluates the table\footnote{The garbled table can be send during the offline phase, while $P_0$ needs to evaluate the garbled circuit during the online phase.} to obtain $\shareG{\vc}_{P_0} = \KeyGAct{\vc}$.  $P_i$ for $i \in \EInSet$ sets $\shareG{\vc}_{P_i} = \KeyGA{\vc}$.
\end{description} 

\subsection{Building Blocks}
\label{sec:Build_Blocks}

\paragraph{Verifiable Arithmetic/Boolean Sharing}
Protocol $\pivSh$ (\boxref{fig:pivSh}) allows two parties $P_i, P_j$ to generate $\shrd$-sharing of value $\val$ in a verifiable manner. On a high level, $P_i$ executes $\piSh$ on $\val$, while $P_j$ helps in verification by sending $\Hash(\Mask{\val})$ to parties $\ESet$.
\begin{protocolbox}{$\pivSh(P_i, P_j, \val)$}{Verifiable Arithmetic/Boolean sharing of a value $\val$.}{fig:pivSh}
	\justify
	\algoHead{Offline:} Parties execute offline steps of $\piSh(P_i, \val)$.
	\vspace{-3mm}
	\justify
	\algoHead{Online:}
	\begin{myitemize}
		\item[--] $P_i$ computes $\Mask{\val} = \val + \Pad{\val}$ and sends to $\ESet$.
		\item[--] $P_j$ computes $\Hash(\Mask{\val})$ and sends to $\ESet$, who $\abort$ if the received values are inconsistent. 
	\end{myitemize}        
\end{protocolbox}

We observe that the parties can {\em non-interactively} generate $\shrd$-sharing of a value $\val$ when all of the parties $\ESet$ know $\val$. Parties set $\PadA{\val} = \PadB{\val}= \PadC{\val} = 0$ and $\Mask{\val} = \val$. We abuse the notation and use $\pivSh(P_1, P_2, P_3, \val)$ to denote this protocol.

\paragraph{Verifiable Garbled Sharing}
Protocol $\pivShG$ (\boxref{fig:pivShG}) is adapted from ABY3~\cite{MR18} and allows two parties $P_i, P_j$ to generate $\shrdG$-sharing of value $\val$ in a verifiable manner. When $P_i, P_j$ are both garblers, one of them can send the key while the other can send just the hash to check for inconsistency. If $P_0 = P_j$, the other parties ($P_1, P_2$) send commitments of the keys in order, to $P_0$. In addition, $P_i$ sends the decommitment of the actual key to $P_0$.

\begin{protocolbox}{$\pivShG(P_i, P_j, \val)$}{Verifiable Garbled sharing of a value $\val$.}{fig:pivShG}
	\justify
	\algoHead{Offline:} $\ESet$ locally sample random $\KeyGA{\val} \in \{0,1\}^{\kappa}$, compute $\KeyGB{\val} = \KeyGA{\val} \xor \RG$ and set $\shareGP{\val}{1} = \shareGP{\val}{2} = \shareGP{\val}{3} = \KeyGA{\val}$.
	\justify
	\algoHead{Online:}
	\begin{myitemize}
		\item[--] {\bf If $(P_i, P_j) = (P_1, P_0)$:}
		\begin{mylist}
			\item[--] $P_1, P_2$ compute commitments $\Commit{\KeyGA{\val}}, \Commit{\KeyGB{\val}}$ and send it to $P_0$. In addition, $P_1$ sends decommitment of $\Commit{\KeyGAct{\val}}$ to $P_0$.
			\item[--] $P_0$ $\abort$ if either the received commitments are inconsistent or the decommitment is incorrect. Else he/she sets $\shareGP{\val}{0} = \KeyGAct{\val}$.
		\end{mylist}
	    \item[--] {\bf If $(P_i, P_j) = (P_k, P_0)$ for $k \in \{2,3\}$:} The steps are similar as above.	    
		\item[--] {\bf If $(P_i, P_j) = (P_1, P_2)$:} $P_1$ and $P_2$ sends $\KeyGAct{\val}$ and $\Hash(\KeyGAct{\val})$ respectively to $P_0$, who $\abort$ if the received values are inconsistent. Else he/she sets $\shareGP{\val}{0} = \KeyGAct{\val}$.
		\item[--] {\bf If $(P_i, P_j) = (P_1, P_3)$ or $(P_2, P_3)$:} The steps are similar as above.	  
	\end{myitemize}        
\end{protocolbox}

\paragraph{Dot Product}
Given two vectors $\vecX$ and $\vecY$, each of size $d$, $\piDotP$ (\boxref{fig:piDotP}) computes the dot product $\wz = \vecX \band \vecY$. ABY3~\cite{MR18} and ASTRA~\cite{CCPS19} have proposed efficient dot product protocols for the semi-honest and malicious settings. In the semi-honest case, their dot product cost is the same as a single multiplication, but for the malicious case it scales with the vector size. In comparison, the cost of our dot product is independent of the vector size. On a high level, instead of performing reconstruction for each multiplication $\wx_j \cdot \wy_j$ for $j \in \{1, \ldots, d\}$, parties locally add their shares corresponding to all the multiplications and perform a single exchange.

\begin{protocolsplitbox}{$\piDotP(\vecX, \vecY)$}{Dot Product Protocol.}{fig:piDotP}
	Let $\wz = \vecX \band \vecY$.
	\justify
	\algoHead{Offline:} 
	\begin{myitemize}
		\item[--] Parties in $\Partyset\setminus\{P_j\}$ together sample $\PadV{\wz}{j}$ for $j \in \EInSet$.
		\item[--] Parties invoke protocol $\piZero$ (\boxref{fig:piZero}) to generate $A, B, \Gamma$ such that $A + B + \Gamma = 0$. Parties locally compute the following:
		\begin{mylist}
			\item[--] $P_0, P_1$ compute $\GammaxyB =\sum_{j=1}^{d} \GammaxyjB = \sum_{j=1}^{d} (\PadB{\wx_j} \PadB{\wy_j} + \PadB{\wx_j} \PadC{\wy_j} + \PadC{\wx_j} \PadB{\wy_j}) + A$.
			\item[--] $P_0, P_2$ compute $\GammaxyC = \sum_{j=1}^{d} \GammaxyjC = \sum_{j=1}^{d} (\PadC{\wx_j} \PadC{\wy_j} + \PadC{\wx_j} \PadA{\wy_j} + \PadA{\wx_j} \PadC{\wy_j}) + B$.
			\item[--] $P_0, P_3$ compute $\GammaxyA = \sum_{j=1}^{d} \GammaxyjA = \sum_{j=1}^{d} (\PadA{\wx_j} \PadA{\wy_j} + \PadA{\wx_j} \PadB{\wy_j} + \PadB{\wx_j} \PadA{\wy_j}) + \Gamma$.
		\end{mylist}
		\item[--] Parties exchange the following:
		\begin{mylist}
			\item[--] $P_1$ receives $\GammaxyC$ and $\Hash(\GammaxyC)$ from $P_2$ and $P_0$ respectively. 
			\item[--] $P_2$ receives $\GammaxyA$ and $\Hash(\GammaxyA)$ from $P_3$ and $P_0$ respectively. 
			\item[--] $P_3$ receives $\GammaxyB$ and $\Hash(\GammaxyB)$ from $P_1$ and $P_0$ respectively. 
		\end{mylist}
		\item[--] $P_i$ for $i \in \EInSet$ $\abort$ if the received values are inconsistent.
	\end{myitemize}
	
	\justify
	\algoHead{Online:} Let $\Mask{\wz}' = \Mask{\wz} - \sum_{j=1}^{d}  \Mask{\wx_j} \Mask{\wy_j}$.
	\begin{myitemize}
		\item[--] Parties locally compute the following:
		\begin{mylist}
			\item[--] $P_1, P_3$: $\Mask{\wz,2}' = \sum_{j=1}^{d} (- \PadB{\wx_j} \Mask{\wy_j} - \PadB{\wy_j} \Mask{\wx_j}) + \GammaxyB + \PadB{\wz}$.
			\item[--] $P_2, P_1$: $\Mask{\wz,3}' = \sum_{j=1}^{d} (- \PadC{\wx_j} \Mask{\wy_j} - \PadC{\wy_j} \Mask{\wx_j}) + \GammaxyC + \PadC{\wz}$.
			\item[--] $P_3, P_2$: $\Mask{\wz,1}' = \sum_{j=1}^{d} (- \PadA{\wx_j} \Mask{\wy_j} - \PadA{\wy_j} \Mask{\wx_j}) + \GammaxyA + \PadA{\wz}$.
		\end{mylist}
		\item[--] Parties exchange the following:
		\begin{mylist}
			\item[--] $P_1$ receives $\Mask{\wz,1}'$ and $\Hash(\Mask{\wz,1}')$ from $P_2$ and $P_3$ respectively. 
			\item[--] $P_2$ receives $\Mask{\wz,2}'$ and $\Hash(\Mask{\wz,2}')$ from $P_3$ and $P_1$ respectively. 
			\item[--] $P_3$ receives $\Mask{\wz,3}'$ and $\Hash(\Mask{\wz,3}')$ from $P_1$ and $P_2$ respectively. 
		\end{mylist}
		\item[--] $P_i$ for $i \in \EInSet$ $\abort$ if the received values are inconsistent. Else, he / she computes $\Mask{\wz} = (\Mask{\wz,1}' + \Mask{\wz,2}' + \Mask{\wz,3}') + + \sum_{j=1}^{d} (\Mask{\wx_j} \Mask{\wy_j}) = \Mask{\wz}' + \sum_{j=1}^{d} (\Mask{\wx_j} \Mask{\wy_j})$. 
	\end{myitemize}     
\end{protocolsplitbox}

\subsection{Sharing Conversions}
\label{sec:Share_Conv}
We now discuss the inter-sharing conversions among {\bf A}rithmetic, {\bf B}oolean, and {\bf G}arbled sharing. 

\paragraph{Garbled to Boolean Sharing (\GB)}
To convert a garbled share into the boolean world, $P_1, P_2$ first generate the garbled and boolean shares of a random value ($\vr$) using their shared randomness in the offline phase. In addition, they communicate the garbled circuit which performs the XOR of two bits along with the decoding information (Note that the garbled circuit does not have to be communicated due to the free XOR technique). In the online phase, $P_0$ evaluates and obtains $\val \xor \vr$ and sends it to $P_3$ along with the hash of the corresponding key. Authenticity of the underlying garbling scheme ensures that a corrupt $P_0$ cannot send the wrong bit, as he will not be able to guess the right key for it. 
\begin{protocolbox}{$\PiGB$}{Garbled to Boolean Sharing.}{fig:GB}
	\justify
	\algoHead{Offline:} 
	\begin{myitemize}
		\item[--] $P_1, P_2$ locally sample random $\vr \in \Z{\ell}$. Parties execute $\pivShG(P_1, P_2, \vr)$ and $\pivShB(P_1, P_2, \vr)$ to generate $\shareG{\vr}$ and $\shareB{\vr}$ respectively.
		\item[--] $\ESet$ garble a boolean adder circuit $\Adder(\wx, \wy)$ that computes $\wx \xor \wy$. $P_1$ sends the garbled circuit along with the decoding information to $P_0$, while $P_2$ sends a combined hash to $P_0$. 
	\end{myitemize} 
	\justify
	\algoHead{Online:}
	\begin{myitemize}
		\item[--] $P_0$ evaluates the circuit $\Adder$ on $\val$ and $\vr$ to obtain $\val \xor \vr$. $P_0$ sends $\val \xor \vr$ along with a hash of the actual key corresponding to $\val \xor \vr$ to $P_3$. $P_3$ $\abort$ if the received values are inconsistent. 
		\item[--] Else, parties execute $\pivShB(P_3, P_0, \val \xor \vr)$ to generate $\shareB{\val \xor \vr}$.
		\item[--] Parties locally compute $\shareB{\val} = \shareB{\val \xor \vr} \xor \shareB{\vr}$.
	\end{myitemize}        
\end{protocolbox}

\vspace{-2mm}
\paragraph{Garbled to Arithmetic Sharing (\GA)}
This conversion proceeds in a similar way as $\PiGB$. The major difference is that instead of the garbled circuit for XOR, the parties communicate a circuit for subtraction of two $\ell$-bit values. Note that $P_0$ needs to communicate only a single hash combining all the keys corresponding to the $\ell$-bits of $\val - \vr$.
\begin{protocolbox}{$\PiGA$}{Garbled to Arithmetic Sharing.}{fig:GA}
	\justify
	\algoHead{Offline:} 
	\begin{myitemize}
		\item[--] $P_1, P_2$ locally sample random $\vr \in \Z{\ell}$. Parties execute $\pivShG(P_1, P_2, \vr)$ and $\pivShA(P_1, P_2, \vr)$ to generate $\shareG{\vr}$ and $\shareA{\vr}$ respectively.
		\item[--] $\ESet$ garble a subtractor circuit $\Sub(\wx, \wy)$ that computes $\wx - \wy$. $P_1$ sends the garbled circuit along with the decoding information to $P_0$, while $P_2$ sends a combined hash to $P_0$, who $\abort$ if the received values are inconsistent. 
	\end{myitemize} 
	\justify
	\algoHead{Online:}
	\begin{myitemize}
		\item[--] $P_0$ evaluates the circuit $\Sub$ on $\val$ and $\vr$ to obtain $\val - \vr$. $P_0$ sends $\val - \vr$ along with a combined hash of all the actual keys corresponding to $\val - \vr$ to $P_3$. $P_3$ $\abort$ if the received values are inconsistent. 
		\item[--] Else, parties execute $\pivShA(P_3, P_0, \val - \vr)$ to generate $\shareA{\val - \vr}$.
		\item[--] Parties locally compute $\shareA{\val} = \shareA{\val - \vr} + \shareA{\vr}$.
	\end{myitemize}        
\end{protocolbox}

\vspace{-2mm}
\paragraph{Boolean to Garbled Sharing (\BG)}
Since the bit $\val = (\Mask{\val} \xor \PadA{\val}) \xor (\PadB{\val} \xor \PadC{\val})$ in the boolean world, if we can get the garbled shares of $\vx = (\Mask{\val} \xor \PadA{\val})$ and $\vy = (\PadB{\val} \xor \PadC{\val})$, parties can use the free XOR technique to compute the garbled shares of $\val$ locally. Each of $\vx, \vy$ is possessed by two parties, enabling them to verifiably generate the garbled shares using the protocol $\pivShG$.
\begin{protocolbox}{$\PiBG$}{Boolean to Garbled Sharing.}{fig:BG}
	\justify
	\algoHead{Offline:} $P_0, P_1$ execute $\pivShG(P_1, P_0, \vy)$ to generate $\shareG{\vy}$ where $\vy = \PadB{\val} \xor \PadC{\val}$.
	\justify
	\algoHead{Online:}
	\begin{myitemize}
		\item[--] $P_2, P_3$ execute $\pivShG(P_2, P_3, \vx)$ to generate $\shareG{\vx}$ where $\vx = \Mask{\val} \xor \PadA{\val}$.
		\item[--] Parties locally compute $\shareG{\val} = \shareG{\vx} \xor \shareG{\vy}$. 
	\end{myitemize}        
\end{protocolbox}

\paragraph{Arithmetic to Garbled Sharing (\AG)}
Similar to $\PiBG$, $\val = (\Mask{\val} - \PadA{\val}) - (\PadB{\val} + \PadC{\val})$ and the parties can verifiably generate the garbled shares of $\vx = (\Mask{\val} - \PadA{\val})$ and $\vy = (\PadB{\val} + \PadC{\val})$ using $\pivShG$. In the online phase, the parties evaluate a garbled subtractor circuit to obtain the shares of $\val = \vx - \vy$.
\begin{protocolbox}{$\PiAG$}{Arithmetic to Garbled Sharing.}{fig:AG}
	\justify
	\algoHead{Offline:} 
	\begin{myitemize}
		\item[--]  $P_0, P_1$ execute $\pivShG(P_1, P_0, \vy)$ to generate $\shareG{\vy}$ where $\vy = \PadB{\val} + \PadC{\val}$.
		\item[--] $\ESet$ garble a subtractor circuit $\Sub(\wx, \wy)$ that computes $\wx - \wy$. $P_1$ sends the garbled circuit to $P_0$, while $P_2$ sends a hash of the same to $P_0$, who $\abort$ if the received values are inconsistent. 
	\end{myitemize} 
	\justify
	\algoHead{Online:}
	\begin{myitemize}
		\item[--] $P_2, P_3$ execute $\pivShG(P_2, P_3, \vx)$ to generate $\shareG{\vx}$ where $\vx = \Mask{\val} - \PadA{\val}$.
		\item[--] Parties compute $\shareG{\val} = \shareG{\vx} - \shareG{\vy}$ by evaluating circuit $\Sub$. 
	\end{myitemize}        
\end{protocolbox}

\paragraph{Arithmetic to Boolean Sharing (\AB)}
This conversion proceeds similarly to $\PiAG$, the only difference being parties now generate boolean shares of $\vx = (\Mask{\val} - \PadA{\val})$ and $\vy = (\PadB{\val} + \PadC{\val})$ and evaluate a boolean subtractor circuit instead to compute boolean shares of $\val = \vx - \vy$.
\begin{protocolbox}{$\PiAB$}{Arithmetic to Boolean Sharing.}{fig:AB}
	\justify
	\algoHead{Offline:} $P_0, P_1$ execute $\pivShB(P_1, P_0, \vy)$ to generate $\shareB{\vy}$ where $\vy = \PadB{\val} + \PadC{\val}$.
	 
	\justify
	\algoHead{Online:}
	\begin{myitemize}
		\item[--] $P_2, P_3$ execute $\pivShB(P_2, P_3, \vx)$ to generate $\shareB{\vx}$ where $\vx = \Mask{\val} - \PadA{\val}$.
		\item[--] Parties compute $\shareB{\val} = \shareB{\vx} - \shareB{\vy}$ by evaluating an $\ell$-bit Boolean subtractor circuit $\Sub$. 
	\end{myitemize}        
\end{protocolbox}

\paragraph{Bit to Arithmetic Sharing (\BitA)}

Let $\vu$ and $\vv$ denote the bits $\Pad{\bitb}$ and $\Mask{\bitb}$ respectively over the ring $\Z{\ell}$. Then,
\begin{align*}
	\bitb = \Mask{\bitb} \xor \Pad{\bitb} = \vv + \vu - 2 \vv \vu
\end{align*}
Party $P_0$ generates $\sgrd$-shares of $\vu$ in the offline phase. To ensure the correctness of the shares, parties $\ESet$ check whether the following equation holds -- $(\Pad{\bitb} \xor \vr_b)' = \vu + \vr_b' - 2 \vu \vr_b'$ where the superscript $(')$ denotes the corresponding bits over ring $\Z{\ell}$. After the verification, parties locally convert $\sgr{\vu}$ to $\shr{\vu}$. In the online phase, parties multiply $\shr{\vu}, \shr{\vv}$ to generate $\shr{\vu\vv}$ followed by locally computing $\shr{\bitb} = \shr{\vv} + \shr{\vu} - 2 \shr{\vu \vv}$. Note that since $\Pad{\vv}$ is set to $0$ while executing $\piSh(P_1, P_2, P_3, \vv)$ protocol, $\gamma_{\vu \vv}$-sharing is not needed during multiplication.
\begin{protocolbox}{$\PiBitA$}{Bit to Arithmetic Sharing.}{fig:BitA}
	\justify
	\algoHead{Offline:} Let $\vu$ and $\PadV{\vu}{i}$ denote the bits $\Pad{\bitb}$ and $\PadV{\bitb}{i}$ respectively over the ring $\Z{\ell}$. Here $i \in \EInSet$. 
	\begin{myitemize}
		\item[--] $P_0$ executes $\piaSh(P_0, \vu)$ (\boxref{fig:piaSh}) to generate $\sgr{\vu}$. Let the shares be $\sgr{\vu}_{P_1} = (\vu_2, \vu_3)$, $\sgr{\vu}_{P_2} = (\vu_3, \vu_1)$, and $\sgr{\vu}_{P_3} = (\vu_1, \vu_2)$.
		\item[--] $\ESet$ performs the following check: 
		\begin{mylist}
			\item[--] $P_1, P_2$ sample a random ring element $\vr$ and a random bit $\vr_b$. Let $\vr_b'$ denotes the bit $\vr_b$ over ring $\Z{\ell}$. 
			\item[--] $P_1$ computes $\vx_1 = \PadC{\bitb} \xor \vr_b, \vy_1 = (\vu_2 + \vu_3)(1 - 2 \vr_b') + \vr_b' + \vr$ and sends $(\vx_1, \vy_1)$ to $P_3$. 
			\item[--] $P_2$ computes $\vy_2 = \vu_1 (1 - 2 \vr_b') - \vr$ and sends $\Hash(\vy_2)$ to $P_3$.
			\item[--] $P_3$ computes $\vx = \Pad{\bitb} \xor \vr_b = \vx_1 \xor \PadA{\bitb} \xor \PadB{\bitb}$ and $\abort$ if $\Hash(\vx' - \vy_1) \ne \Hash(\vy_2)$. Here $\vx'$ denotes the bit $\vx$ over ring $\Z{\ell}$.
		\end{mylist}
		\item[--] If the verification succeeds, $\ESet$ converts  $\sgr{\vu}$ to $\shr{\vu}$ locally by setting $\Mask{\vu} = 0$ and $\sgr{\Pad{\vu}} = - \sgr{\vu}$.
	\end{myitemize} 
	\justify
	\algoHead{Online:} Let $\vv$ denotes the bit $\Mask{\bitb}$ over ring $\Z{\ell}$.
	\begin{myitemize}
		\item[--] Parties execute $\pivSh(P_1, P_2, P_3, \vv)$ to generate $\shr{\vv}$.
		\item[--] Parties execute $\piMult$ on $\shr{\vu}$ and $\shr{\vv}$ to generate $\shr{\vu \vv}$.
		\item[--] Parties locally compute $\shr{\bitb} = \shr{\vv} + \shr{\vu} - 2 \shr{\vu \vv}$. 
	\end{myitemize}        
\end{protocolbox}

\paragraph{Boolean to Arithmetic Sharing (\BA)}
We use the fact that a value $\val$ can be expressed as $\sum_{i = 0}^{\ell - 1} 2^i \cdot \val_{i}$, where $\val_i$ denotes the $i$th bit of $\val$ over a ring $\Z{\ell}$. Note that 
\begin{align*}
\val = \sum_{i = 0}^{\ell - 1} 2^i \cdot \val_{i} = \sum_{i = 0}^{\ell - 1} 2^i \cdot (\Mask{\vv_i}' + \Pad{\vu_{i}}' - 2 \Mask{\vv_i}' \cdot \Pad{\vu_{i}}')
\end{align*}
where $\Mask{\vv_i}'$ and $\Pad{\vu_{i}}'$ denote the bits $\Mask{\vv_i}$ and $\Pad{\vu_{i}}$ respectively over the ring $\Z{\ell}$.

The offline phase of $\PiBA$ proceeds similar to $\PiBitA$, where each bit $\Pad{\val_i}$ for $i \in \{0,\ldots,\ell-1 \}$ is converted to $\sgrd$-share. During the online phase, parties locally compute $\sqd$-shares of $\val$ followed by generating $\shrd$-shares of it by executing $\pivSh$ protocol. Parties then locally add their shares to obtain $\shr{\val}$.
\begin{protocolsplitbox}{$\PiBA$}{Boolean to Arithmetic Sharing V2.}{fig:BA}
	\justify
	\algoHead{Offline:} Let $\val_i$ denotes the $i$th bit of value $\val$. Let $\vp_i$ and $\PadV{\vp_i}{j}$ denote the bits $\Pad{\val_i}$ and $\PadV{\val_i}{j}$ respectively over the ring $\Z{\ell}$, where $i \in \{0,\ldots,\ell-1 \}$ and $j \in \EInSet$.
	\begin{myitemize}
		\item[--] Parties execute offline steps of $\PiBitA$ on each $\vp_i$ for $i \in \{0,\ldots,\ell-1 \}$, to generate $\sgr{\vp_i}$. Let the shares be $\sgr{\vp_i}_{P_1} = (\vp_{i,2}, \vp_{i,3})$, $\sgr{\vp_i}_{P_2} = (\vp_{i,3}, \vp_{i,1})$, and $\sgr{\vp_i}_{P_3} = (\vp_{i,1}, \vp_{i,2})$.
	\end{myitemize} 
	\justify
	\algoHead{Online:} Let $\vq_i$ for $i \in \{0,\ldots,\ell-1 \}$ denotes bit $\Mask{{\val}_i}$ over $\Z{\ell}$.
	\begin{myitemize}
		\item[--] Parties compute the following:
		\begin{mylist}
			\item[--] $P_1, P_3$ compute $\vx = \sum_{i=0}^{\ell-1} 2^i (\vq_i + \vp_{i,2} - 2 \vq_i \cdot \vp_{i,2})$.
			\item[--] $P_2, P_1$ compute $\vy = \sum_{i=0}^{\ell-1} 2^i (\vp_{i,3} - 2 \vq_i \cdot \vp_{i,3})$.
			\item[--] $P_3, P_2$ compute $\vz = \sum_{i=0}^{\ell-1} 2^i (\vp_{i,1} - 2 \vq_i \cdot \vp_{i,1})$.
		\end{mylist}
		\item[--] Parties generate $\shr{\vx}, \shr{\vy}$ and $\shr{\vz}$ by executing $\pivSh(P_1, P_3, \vx)$, $\pivSh(P_2, P_1, \vy)$ and $\pivSh(P_3, P_2, \vz)$ respectively.
		\item[--] Parties locally compute $\shr{\val} = \shr{\vx} + \shr{\vy} + \shr{\vz}$.
	\end{myitemize}        
\end{protocolsplitbox}

\paragraph{Bit Injection (\BitInj): $\shareB{b}\shr{\val} \rightarrow \shr{b \val}$}
Let $\vy_1$ and  $\vy_2$ denote the values $\Pad{\bitb}$ and $\Pad{\bitb}\Pad{\val}$ respectively over ring $\Z{\ell}$. Similarly, let $\vx_0, \vx_1, \vx_2$ and $\vx_3$ denote the values $(\Mask{\bitb}\Mask{\val})$, $(\Mask{\bitb})$, $(\Mask{\val} -2 \Mask{\val} \Mask{\bitb})$ and $(2 \Mask{\bitb} -1)$ respectively over ring $\Z{\ell}$. Then,
\begin{align*}
\bitb \cdot \val &= (\Mask{\bitb} \xor \Pad{\bitb})(\Mask{\val} - \Pad{\val}) = \vx_0 - \vx_1 \vy_1 + \vx_2 \vy_2 + \vx_3 \vy_3
\end{align*}

\begin{protocolbox}{$\PiBitInj$}{Bit Injection: $\shareB{\bitb}\shareA{\val} \rightarrow \shareA{\bitb \val}$.}{fig:BitInj}
	\justify
	\algoHead{Offline:} Let $\vy_1$ and  $\vy_2$ denote the values $\Pad{\bitb}$ and $\Pad{\bitb}\Pad{\val}$ respectively over $\Z{\ell}$.
	\begin{myitemize}
		\item[--] $P_0$ executes $\piaSh(P_0, \vy_j)$ to generate $\sgr{\vy_j}$ for $j \in \{1,2\}$. Let the shares be $\sgr{\vy_j}_{P_1} = (\vy_{j,2}, \vy_{j,3})$, $\sgr{\vy_j}_{P_2} = (\vy_{j,3}, \vy_{j,1})$, and $\sgr{\vy_j}_{P_3} = (\vy_{j,1}, \vy_{j,2})$. 
		\item[--] Parties verify the correctness of $\sgr{\vy_1}$ using the steps similar to protocol $\PiBitA$ (\boxref{fig:BitA}). To verify the correctness of $\sgr{\vy_2}$, parties proceed as follows: 
		\begin{mylist}
			\item[--] Parties execute $\piZero$ (\boxref{fig:piZero}) to generate $A, B, \Gamma$ such that $A + B + \Gamma = 0$.
			\item[--] $P_1$ computes $\vu_2 = \PadB{\vy_1} \PadB{\val} + \PadB{\vy_1} \PadC{\val} + \PadC{\vy_1} \PadB{\val} + A$.
			\item[--] $P_2$ computes $\vu_3 = \PadC{\vy_1} \PadC{\val} + \PadC{\vy_1} \PadA{\val} + \PadA{\vy_1} \PadC{\val} + B$.
			\item[--] $P_3$ computes $\vu_1 = \PadA{\vy_1} \PadA{\val} + \PadA{\vy_1} \PadB{\val} + \PadB{\vy_1} \PadA{\val} + \Gamma$.
			\item[--] $P_1$ and $P_2$ send $\vz_2$ and $\Hash(-\vz_3)$ respectively to $P_3$, where $\vz_2 = \vu_2 - \vy_{2,2}$ and $\vz_3 = \vu_3 - \vy_{2,3}$. 
			\item[--] $P_3$ sets $\vz_1 = \vu_1 - \vy_{2,1}$ and $\abort$ if $\Hash(\vz_1 + \vz_2) \ne \Hash(-\vz_3)$.
		\end{mylist}
	\end{myitemize} 
	\vspace{-3mm}
	\justify
	\algoHead{Online:} Let $\vx_0, \vx_1, \vx_2$ and $\vx_3$ denote the values $(\Mask{\bitb}\Mask{\val})$, $(\Mask{\bitb})$, $(\Mask{\val} -2 \Mask{\val} \Mask{\bitb})$ and $(2 \Mask{\bitb} -1)$ respectively over ring $\Z{\ell}$.
	\begin{myitemize}
		\item[--] Parties compute the following:
		\begin{mylist}
			\item[--] $P_1, P_3$ compute $\vc_2 = \vx_0 - \vx_1 \PadB{\val} + \vx_2 \vy_{1,2} + \vx_3 \vy_{2,2}$.
			\item[--] $P_2, P_1$ compute $\vc_3 = - \vx_1 \PadC{\val} + \vx_2 \vy_{1,3} + \vx_3 \vy_{2,3}$.
			\item[--] $P_3, P_2$ compute $\vc_1 = - \vx_1 \PadA{\val} + \vx_2 \vy_{1,1} + \vx_3 \vy_{2,1}$.
		\end{mylist}
		\item[--] Parties execute $\pivSh(P_1, P_3, \vc_2)$, $\pivSh(P_2, P_1, \vc_3)$ and $\pivSh(P_3, P_2, \vc_1)$ to generate $\shr{\vc_2}$, $\shr{\vc_3}$ and $\shr{\vc_1}$ respectively.
		\item[--] Parties locally compute $\shr{\bitb \val} = \shr{\vc_1} + \shr{\vc_2} + \shr{\vc_3}$.
	\end{myitemize}        
\end{protocolbox}

\vspace{-2mm}
In the offline phase, $P_0$ generates $\sgrd$-shares of $\Pad{\bitb}'$ and $\Pad{\bitb} \Pad{\val}$ where $\Pad{\bitb}'$ denotes the bit $\Pad{\bitb}$ over $\Z{\ell}$. The check for $\sgr{\Pad{\bitb}'}$ is the same as the one for $\PiBitA$, to check $\sgr{\Pad{\bitb} \Pad{\val}}$ parties proceed  as mentioned in protocol $\PiBitInj$ above. During the online phase, parties locally compute $\sqd$-shares of $\bitb \cdot \val$ followed by generating $\shrd$-shares of it by executing $\pivSh$ protocol. Parties then locally add their shares to obtain $\shr{\bitb \cdot \val}$.

\section{Privacy Preserving Machine Learning}
\label{sec:privML}
Most of the intermediate values in machine learning algorithms involve operating over decimals. To represent decimal values, we use signed two's compliment over $\Z{\ell}$~\cite{MohasselZ17, MR18, CCPS19}, where the most significant bit ($\MSB$) represents the sign and the last $d$ bits represent the fractional part.

In order to perform privacy-preserving machine learning, we need efficient instantiations of three components -- Share Truncation, Secure Comparison, and Non-linear Activation Functions. This section covers our protocols for performing the aforementioned components.

\subsection{Share Truncation}
We take inspiration for truncation from ABY3~\cite{MR18}, where they perform it on shares after evaluating a multiplication gate, preserving the underlying value with very high probability. Our approach improves upon ABY3 by not using any boolean circuits, thus improving the offline round complexity to constant. 

\begin{protocolbox}{$\piMultTr(\Wxyz)$}{Multiplication with Truncation.}{fig:piMultTr}
	\justify
	\algoHead{Offline:} 
	\begin{myitemize}
		\item[--] Parties execute the offline steps of protocol $\piMult(\Wxyz)$ apart from $\Pad{\wz}$ not being generated.
		\item[--] Parties locally sample the following random values:
		\begin{equation*}
		\Partyset \setminus \{P_2\} : \vr_2,~~~~~\Partyset \setminus \{P_1\} : \vr_1,~~~~~\Partyset \setminus \{P_3\} : \vr_3
		\end{equation*}
		\item[--] $P_0$ locally compute $\vr = \vr_1 + \vr_2 + \vr_3$, locally truncates it to obtain $\vrt$ and executes $\piaSh(P_0, \vrt)$ to generate $\sgr{\vrt}$. Let the shares be $\sgr{\vrt}_{P_1} = (\vrt_2, \vrt_3)$, $\sgr{\vrt}_{P_2} = (\vrt_3, \vrt_1)$, and $\sgr{\vrt}_{P_3} = (\vrt_1, \vrt_2)$.
		\item[--] Let $\vr_{\vd}$ and $\vr_{\vd,i}$ denote the last $\vd$ bits of $\vr$ and $\vr_i$ respectively for $i \in \EInSet$. Parties verify the correctness of $\sgr{\vrt}$ as follows:
		\begin{mylist}
			\item[--] $P_1$ samples a random element $\vc$ and computes $\vm_1 = \vr_2 - 2^{\vd} \vrt_2 - \vr_{\vd,2} + \vc$. $P_1$ sends $(\vm_1, \Hash(\vc))$ to $P_2$.
			\item[--] $P_2$ computes $\vm_2 = (\vr_1 + \vr_3) - 2^{\vd} (\vrt_1 + \vrt_3) - (\vr_{\vd,1} + \vr_{\vd,3}) $ and $\abort$ if $\Hash(\vm_1 + \vm_2) \ne \Hash(\vc)$.
		\end{mylist}
		\item[--] Parties locally convert $\sgr{\vrt}$ to $\shr{\vrt}$ by setting $\Mask{\vrt} = 0$ and $\sgr{\Pad{\vrt}} = \sgr{\vrt}$. 
	\end{myitemize}
	\vspace{-3mm}
	\justify
	\algoHead{Online:} Let $\wz' = (\wz - \vr) - \Mask{\wx} \Mask{\wy}$.
	\begin{myitemize}
		\item[--] Parties locally compute the following:
		\begin{mylist}
			\item[--] $P_1, P_3$ compute $\sqrB{\wz'} = - \PadB{\wx} \Mask{\wy} - \PadB{\wy} \Mask{\wx} + \GammaxyB - \vr_2$.
			\item[--] $P_2, P_1$ compute $\sqrC{\wz'} = - \PadC{\wx} \Mask{\wy} - \PadC{\wy} \Mask{\wx} + \GammaxyC - \vr_3$.
			\item[--] $P_3, P_2$ compute $\sqrA{\wz'} = - \PadA{\wx} \Mask{\wy} - \PadA{\wy} \Mask{\wx} + \GammaxyA - \vr_1$.
		\end{mylist}
		\item[--] Parties exchange the following:
		\begin{mylist}
			\item[--] $P_1$ receives $\sqrA{\wz'}$ and $\Hash(\sqrA{\wz'})$ from $P_2$ and $P_3$ respectively. 
			\item[--] $P_2$ receives $\sqrB{\wz'}$ and $\Hash(\sqrB{\wz'})$ from $P_3$ and $P_1$ respectively. 
			\item[--] $P_3$ receives $\sqrC{\wz'}$ and $\Hash(\sqrC{\wz'})$ from $P_1$ and $P_2$ respectively. 
		\end{mylist}
		\item[--] $P_i$ for $i \in \EInSet$ $\abort$ if the received values are inconsistent. Else, he computes $(\wz - \vr) = \sqrA{\wz'} + \sqrB{\wz'} + \sqrC{\wz'} + \Mask{\wx} \Mask{\wy}$. 
		\item[--] $\ESet$ locally truncates $(\wz - \vr)$ to obtain $(\wz - \vr)^{\vt}$, followed by executing $\pivSh(\ESet, (\wz - \vr)^{\vt})$ to generate $\shr{(\wz - \vr)^{\vt}}$.
		\item[--] Parties locally compute $\shr{\wz^{\vt}} = \shr{(\wz - \vr)^{\vt}} + \shr{\vrt}$.
	\end{myitemize}     
\end{protocolbox}
We start by generating a random $(\vr, \vrt)$ in the offline phase, where $\vrt$ is the truncated value of $\vr$. The truncated value of $\wz$ can be obtained by first opening and truncating $\wz - r$, and then adding it to $\vrt$. Parties non-interactively generate $\sgr{\vr}$, such that $P_0$ obtains $\vr$. This is followed by $P_0$ generating $\shr{\vrt}$, but since we cannot rely on $P_0$, parties $\ESet$ perform a check to ensure the correctness of the share. On a high level, parties check the relation $\vr = 2^{\vd} \vrt + \vr_{\vd}$, where $\vr_{\vd}$ denotes the last $d$ bits of $\vr$ The formal details of the check are provided in the protocol above and the correctness appears in Lemma~\ref{app:piMultTrC}.
\subsection{Secure Comparison}
The secure comparison technique allows parties to check whether $\wx < \wy$, given arithmetic shares of $\wx, \wy$. In fixed point arithmetic, a simple way to achieve this is by computing $\wx - \wy$, and checking its sign, stored in the $\MSB$ position. This protocol, inspired from~\cite{MR18}, is called Bit Extraction ($\piBitExt$), since it extracts a bit from the given arithmetic shares and outputs the boolean shares of the bit.  
Towards this, $P_0, P_3$ generate the boolean sharing of $\vy = \PadA{\val} + \PadB{\val}$ in the offline, while $p_1, P_2$ generate the boolean sharing of $\vx = \Mask{\val} - \PadC{\val}$ in the online. Parties then evaluate the optimized Parallel Prefix Adder of~\cite{MR18} to obtain the MSB bit in boolean shared format.

\subsection{Activation Functions}
\label{sec:actFunc}

\paragraph{ReLU}
The ReLU function is defined as $\ReLU(\val) = \maxv(0, \val)$. This can be viewed as $\ReLU(\val) = (1 \xor \bitb) \val$ where bit $\bitb = 1$ if $\val < 0$ and $0$ otherwise. In order to generate $\shr{\ReLU(\val)}$, parties first execute $\piBitExt$ on $\val$ to obtain $\shareB{\bitb}$ and locally compute $\shareB{1 \xor \bitb}$. This is followed by executing $\PiBitInj$ on $\shareB{1 \xor \bitb}$ and $\shr{\val}$. The derivative of $\ReLU$, denoted by $\dReLU(\val) = (1 \xor \bitb)$.

\paragraph{Sigmoid}
In our protocols, we use the approximation of sigmoid function~\cite{MohasselZ17, MR18, CCPS19}, defined as:
\vspace{-2mm}
\begin{align*}
\Sig(\val) = \left\{
               \begin{array}{lll}
                   0                  & \quad \val < -\frac{1}{2} \\
                   \val + \frac{1}{2} & \quad - \frac{1}{2} \leq \val \leq \frac{1}{2} \\
                   1                  & \quad \val > \frac{1}{2}
               \end{array}
             \right.
\end{align*}
This can be viewed as, $\Sig(\val) = (1 \xor \bitb_1) \bitb_2 (\val + 1/2) + (1 \xor \bitb_2)$, where $\bitb_1 = 1$ if $\val + 1/2 < 0$ and $\bitb_2 = 1$ if $\val - 1/2 < 0$. The protocol is similar to that of $\ReLU$ apart from an additional bit extraction, bit multiplication and a bit injection is required.

\section{Implementation and Benchmarking}
\label{sec:Implementation}
The improvements of our framework over the current state-of-the-art (ABY3) are showcased through our implementation, comparing the two. The training and prediction phases of Linear Regression, Logistic Regression, Neural Networks (NN), and Convolutional Neural Networks (CNN) are used for benchmarking. While we compare our construction with the malicious ABY3 in this section, the comparison of ours with the semi-honest version of ABY3 and with the 4PC protocol of Gordon et al.\cite{GordonR018} are deferred to Appendix~\ref{app:Bench}.

\paragraph{Environment Details} 
We provide results for both LAN ($1$Gbps bandwidth) and WAN ($40$Mbps bandwidth) settings. In the LAN setting, we have machines with 3.6 GHz Intel Core i7-7700 CPU and 32 GB RAM. In the WAN setting, we use Google Cloud Platform\footnote{https://cloud.google.com/} with machines located in West Europe ($P_0$), East Australia ($P_1$), South Asia ($P_2$) and South East Asia ($P_3$). We use n1-standard-8 instances, where machines are equipped with 2.3 GHz Intel Xeon E5 v3 (Haswell) processors supporting hyper-threading, with 8 vCPUs, and 30 GB RAM. We measured the average round-trip time ($\rtt$) for communicating 1 KB of data between every pair of parties. Over the LAN setting, the $\rtt$ turned out to be $0.296 ms$. In the WAN setting, the $\rtt$ values were
	\begin{center} 
		\resizebox{0.48\textwidth}{!}
		{
		\begin{tabular}{c c c c c c}
			\toprule
			$P_0$-$P_1$ & $P_0$-$P_2$ & $P_0$-$P_3$ & $P_1$-$P_2$ & $P_1$-$P_3$ & $P_2$-$P_3$\\
			\midrule
			$274.83 ms$ & $174.13 ms$ & $219.45 ms$ & $152.3 ms$  & $60.19 ms$  & $92.63 ms$\\
			\bottomrule 
		\end{tabular}
		}
	\end{center}

We implement our protocols using the ENCRYPTO library~\cite{ENCRYPTO} in C++17 over a $64$ bit ring. Since the codes for ABY3 and MRZ~\cite{MRZ15}  are not publicly available, we implement their protocols in our environment. The hash function is instantiated using SHA-256. We use multi-threading, wherever possible, to facilitate efficient computation and communication among the parties. To even out the results, each experiment is run 20 times and the average values are reported. 

\paragraph{Datasets} 
To benchmark the training phase of machine learning algorithms, it is common practice to use synthetic datasets so that we have freedom to choose the parameters of the datasets. However, testing the accuracy of a trained model must be carried out with a real dataset, which is the  MNIST~\cite{MNIST10} in our case. It contains $28 \times 28$ pixel images of handwritten numbers and 784 features. For benchmarking of the prediction phase, we use the following real-world datasets:
\vspace{-5mm}
\begin{center} 
	\resizebox{0.48\textwidth}{!}
	{
		\begin{tabular}{l r r}
			\toprule
			\multicolumn{1}{c}{Dataset} & $\#$features & $\#$samples\\
			\midrule
			{\bf Candy (CD)} Power Ranking~\cite{CANDY} 						& 13   & 85\\
			{\bf Boston (BT)} Housing Prices~\cite{BOSTON} 					& 14   & 506\\
		    {\bf Weather (WR)} Conditions in World War Two~\cite{WEATHER} 	& 31   & $\approx$119000\\
			{\bf CalCOFI (CI)} - Oceanographic Data~\cite{CALCOFI} 			& 74   & $\approx$876000\\
			{\bf Epileptic (EP)} Seizures~\cite{EPILEPTIC} 					& 179  & $\approx$11500\\
			Food {\bf Recipes (RE)}~\cite{RECIPE17} 							& 680  & $\approx$20000\\
			{\bf MNIST} ~\cite{MNIST10} 								& 784  & 70000\\
			\bottomrule 
		\end{tabular}
	}
\end{center}
We choose Boston, Weather and CalCOFI for linear regression, since they are best suited for it, while Candy, Epileptic and Recipes were chosen for logistic regression. For NN and CNN, we used the MNIST dataset.
\subsection{Secure Training}
\label{sec:Bench_Training}
The training phase in most of the machine learning algorithms consists of two stages-- i) forward propagation, where the model computes the output, and ii) backward propagation, where the model parameters are adjusted according to the computed output and the actual output. We define one {\em iteration} in the training phase as one forward propagation followed by a backward propagation.

This section covers the improvements in the training phase of our protocol as compared to ABY3. We report the performance in terms of the number of iterations over varying feature ($d$) and batch sizes ($B$), where $d \in \{10, 100, 1000\}$ and $B \in \{128, 256, 512\}$. In LAN, we use {\em iterations per second} ($\#$it/sec) as the metric, but since $\rtt$ is much higher for WAN, we instead use {\em iterations per minute} ($\#$it/min). 

\paragraph{\bf Linear Regression}
For linear regression, one iteration can be viewed as updating the weight vector $\vecW$ using the Gradient Descent algorithm (GD). The update function for $\vecW$ is given by
\begin{align*}
	\vecW = \vecW - \frac{\alpha}{B} \Mat{X}_i^{T}  \circ (\Mat{X}_i \circ \vecW-\Mat{Y}_i)
\end{align*}
where $\alpha$ denotes the learning rate and $\Mat{X}_i$ denotes a subset of batch size $B$, randomly selected from the entire dataset in the $i$th iteration. Here the forward propagation consists of computing $\Mat{X}_i \circ \vecW$, while the weight vector is updated in the backward propagation. The update function consists of a series of matrix multiplications, which in turn can be achieved using dot product protocols. The operations of subtraction as well as multiplication by a public constant can be performed locally. We observe that the aforementioned update function can be computed entirely in the arithmetic domain and can be viewed in form of $\shrd$-shares as
\begin{align*}
	\shr{\vecW} = \shr{\vecW} - \frac{\alpha}{B} \shr{\Mat{X}_j^{T}}  \circ (\shr{\Mat{X}_j} \circ \shr{\vecW}-\shr{\Mat{Y}_j})
\end{align*}

\begin{table}[ht]
	\centering
	\resizebox{.46\textwidth}{!}{
		\begin{tabular}{c|c|c|r|r|r}
			\toprule
			\multirow{2}[2]{*}{Network} & \multirow{2}[2]{*}{\#features} & \multirow{2}[2]{*}{Ref.} & \multicolumn{3}{c}{Batch Size B}\\ \cmidrule{4-6}
			& & & \multicolumn{1}{c|}{128} &  \multicolumn{1}{c|}{256} &  \multicolumn{1}{c}{512} \\
			\midrule
			\multirow{6}[4]{*}{\makecell{LAN\\$\#$it/sec}} 
			& \multirow{2}{*}{10}   & ABY3          & $287.36$ 	& $246.92$ 	& $186.22$  \\
			&                       & {\bf This}    & $1639.35$ & $1204.82$ & $1162.8$ 	\\
			\cmidrule{2-6}
			& \multirow{2}{*}{100}  & ABY3          & $110.38$ 	& $62.08$ 	& $29.27$ 	\\
			&                       & {\bf This}    & $1587.31$ & $1176.48$ & $1136.37$ \\
			\cmidrule{2-6}
			& \multirow{2}{*}{1000} & ABY3          & $13.51$ 	 & $6.88$ 	& $3.43$ 	\\
			&                       & {\bf This}    & $1095.3$ 	 & $883.4$ 	& $861.33$ 	\\
			\midrule
			\multirow{6}[4]{*}{\makecell{WAN\\$\#$it/min}}   
			& \multirow{2}{*}{10}   & ABY3         & $97.57$ 	 & $97.57$ 	     & $97.57$ 	\\
			&                       & {\bf This}   & $195.14$ 	 & $195.14$ 	 & $195.14$ \\
			\cmidrule{2-6}
			& \multirow{2}{*}{100}  & ABY3          & $97.57$ 	 & $97.11$ 	     & $95.08$ 	\\
			&                       & {\bf This}    & $195.14$ 	 & $195.14$ 	 & $195.14$ \\
			\cmidrule{2-6}
			& \multirow{2}{*}{1000} & ABY3          & $89.95$ 	 & $80.10$ 	     & $68.94$ 	\\
			&                       & {\bf This}    & $195.14$ 	 & $195.14$ 	 & $195.14$ \\
			\bottomrule
		\end{tabular}
	}
	\caption{\small Comparison of ABY3 (Malicious) and {\bf This} for Linear Regression (higher = better).}\label{tab:LinRegBench}
\vspace{-5mm}
\end{table}


\tabref{LinRegBench} provides concrete values for Linear Regression. Our improvement over LAN ranges from $4.88 \times$ to $251.84 \times$ and $2 \times$ to $2.83 \times$ over WAN. The gain comes due to two factors: One being the amount we save through our feature-independent communication of the dot product protocol ($3$ ring elements as opposed to $9d$). The other factor is our efficient truncation protocol, which reduces the online communication from $12$ elements to $3$ elements -- by $75\%$. The reason for the discrepancy in gains in LAN and WAN is because in LAN, the $\rtt$ is in the order of microseconds, and scales with the communication size. In contrast, the $\rtt$ in WAN is in the order of milliseconds and does not scale with communication up to a threshold, within which all our protocols operate. 

\paragraph{\bf Logistic Regression}
The iteration for the case of logistic regression is similar to that of linear regression, apart from an activation function being applied on $\Mat{X}_i \circ \vecW$ in the forward propagation. We instantiate the activation function using sigmoid (Section~\ref{sec:actFunc}). The update function for $\vecW$ is given by
\begin{align*}
\vecW = \vecW - \frac{\alpha}{B} \Mat{X}_i^{T}  \circ \left(\Sig(\Mat{X}_i \circ \vecW)-\Mat{Y}_i\right)
\end{align*}
One iteration of logistic regression incurs an additional cost for computing $\Sig(\Mat{X}_j  \circ {\vecW})$ as compared with that for linear regression.

\begin{table}[ht]
	\centering
	\resizebox{.46\textwidth}{!}{
		\begin{tabular}{c|c|c|r|r|r}
			\toprule
			\multirow{2}[2]{*}{Network} & \multirow{2}[2]{*}{\#features} & \multirow{2}[2]{*}{Ref.} & \multicolumn{3}{c}{Batch Size B}\\ \cmidrule{4-6}
			& & & \multicolumn{1}{c|}{128} & \multicolumn{1}{c|}{256} & \multicolumn{1}{c}{512} \\
			\midrule
			\multirow{6}[4]{*}{\makecell{LAN\\$\#$it/sec}} 
			& \multirow{2}{*}{10}   & ABY3         & $56.95$ 	 & $42.02$ 	 & $30.35$ 	\\
			&                       & {\bf This}   & $338.99$ 	 & $257.01$  & $226.61$ \\
			\cmidrule{2-6}
			& \multirow{2}{*}{100}  & ABY3          & $43.34$ 	 & $27.89$ 	 & $16.2$ 	\\
			&                       & {\bf This}    & $336.71$ 	 & $255.69$  & $225.64$  \\
			\cmidrule{2-6}
			& \multirow{2}{*}{1000} & ABY3          & $11.36$ 	 & $6.06$ 	 & $3.13$	\\
			&                       & {\bf This}    & $307.41$ 	 & $238.44$  & $212.23$ \\
			\midrule
			\multirow{6}[4]{*}{\makecell{WAN\\$\#$it/min}}   
			& \multirow{2}{*}{10}   & ABY3          & $20.54$ 	 & $20.54$ 	 & $20.52$ 	\\
			&                       & {\bf This}    & $55.76$ 	 & $55.76$ 	 & $55.76$ 	\\
			\cmidrule{2-6}
			& \multirow{2}{*}{100}  & ABY3          & $20.54$ 	 & $20.52$ 	 & $20.41$ 	\\
			&                       & {\bf This}    & $55.76$ 	 & $55.76$ 	 & $55.76$ 	\\
			\cmidrule{2-6}
			& \multirow{2}{*}{1000} & ABY3          & $20.18$ 	 & $19.64$ 	 & $18.87$ 	\\
			&                       & {\bf This}    & $55.76$ 	 & $55.76$ 	 & $55.76$ 	\\
			\bottomrule
		\end{tabular}
	}
	\caption{\small Comparison of ABY3 (Malicious) and {\bf This} for Logistic Regression (higher = better).}\label{tab:LogRegBench}
\end{table}

\tabref{LogRegBench} provides concrete values for Logistic Regression. Logistic Regression can be thought of as an execution of Linear Regression followed by a Sigmoid function on the output, due to which our improvements for Linear Regression carry over to Logistic. Our improvement ranges from $5.95 \times$ to $67.88 \times$ over LAN and $2.71 \times$ to $2.96 \times$ over WAN. Our efficient Sigmoid protocol takes this result further by improving upon the round and communication complexity. The round complexity is brought down to constant from $4 + \log\ell$ to $5$. Instantiated over ring $\Z{64}$, this amounts to an improvement of $50\%$. The communication is also improved by $\approx 80\%$ (from $81$ elements to roughly $16$). 

\paragraph{\bf Neural Networks}
For the case of NN, we follow steps similar to that of ABY3, where each node across all the layers, except the last layer, uses ReLU ($\ReLU$) as the activation function. At the output layer, we use the MPC friendly variant of the softmax activation function, $\mathsf{smx(u_i) = \frac{\ReLU(u_i)}{\sum_{j = 1}^{n_{f}} \ReLU(u_j)}}$, proposed by SecureML~\cite{MohasselZ17}. In order to perform the division, we switch from arithmetic to garbled world and then use a division garbled circuit. 

The network is trained using the Gradient Descent, where the forward propagation comprises of computing activation matrices for all the layers in the network. Here, the activation matrix for all the layers except the output, is defined as $\Mat{A}_i = \ReLU(\Mat{U}_i)$, where $\Mat{U}_i = \Mat{A}_{i-1}  \circ \Mat{W}_i$ . $\Mat{A}_{0}$ is initialized to $\Mat{X}_{j}$, where $\Mat{X}_j$ is a subset of batch size $B$, randomly selected from the entire dataset for the $j^{th}$ iteration. The activation matrix for the output layer is defined as $\Mat{A}_m  = \mathsf{smx(\Mat{U}_m)}$. 

During the backward propagation, error matrices are computed first. The error matrix for the output layer is defined as $\Mat{E}_m = (\Mat{A}_m - \Mat{T})$, while for the remaining layers it is defined as $\Mat{E}_i = (\Mat{E}_{i+1} \circ \Mat{W}_i^T) \otimes \dReLU(\Mat{U}_i)$. Here the operation $\otimes$ denotes element wise multiplication and $\dReLU$ denotes the derivative of ReLU. This is followed by updating the weights as $\Mat{W}_i = \Mat{W}_i - \frac{\alpha}{B} \Mat{A}_{i-1}^T  \circ \Mat{E}_i$.

\begin{table}[ht]
	\centering
		\resizebox{.48\textwidth}{!}{
		\begin{tabular}{c|c|r|r|r|r|r|r}
			\toprule
			\multirow{2}[2]{*}{Network} & \multirow{2}[2]{*}{Ref.} 
			& \multicolumn{3}{c}{LAN~($\#$it/sec)} & \multicolumn{3}{|c}{WAN~($\#$it/min)}\\ \cmidrule{3-8}
			& & \multicolumn{1}{c|}{B-128} & \multicolumn{1}{c|}{B-256} & \multicolumn{1}{c|}{B-512}
			  & \multicolumn{1}{c|}{B-128} & \multicolumn{1}{c|}{B-256} & \multicolumn{1}{c}{B-512} \\
			\midrule
			\multirow{2}{*}{\makecell{NN}}	
			& ABY3			& $0.37$ 	 & $0.25$ 	 & $0.15$  & $4.69$ 	 & $4.28$ 	 & $3.87$\\
			& {\bf This} 	& $23.00$ 	 & $13.55$ 	 & $7.70$  & $13.94$ 	 & $13.94$ 	 & $13.79$\\
			\midrule
			\multirow{2}{*}{\makecell{CNN}}	
			& ABY3			& $0.23$ 	 & $0.18$ 	 & $0.13$  & $4.34$ 	 & $4.05$ 	 & $3.71$\\
			& {\bf This} 	& $10.46$ 	 & $5.63$ 	 & $2.99$  & $13.86$ 	 & $13.67$ 	 & $13.16$\\
			\bottomrule
		\end{tabular}
		}
	\caption{\small Comparison of ABY3 (Malicious) and {\bf This} for NN and CNN (higher = better).}\label{tab:NNBench}
\end{table}

We consider a NN with two hidden layers, each having 128 nodes followed by an output layer of 10 nodes. After each layer, the activation function ReLU is applied. NN training involves one forward pass followed by one back-propagation. In LAN, the number of iterations is maximum with a batch of 128, at $22.99$ \#it/sec, and comes down to $7.70$ with a batch size of 512. Similarly, over WAN it is maximum at $13.94$ and comes down to $13.79$. As expected, the \#it/sec has not decreased with increase in features due to our dot product protocol being feature-independent in terms of communication. ABY3 on the other hand, has reported 2.5 \#it/sec with a batch size of 128, in the computationally lighter semi-honest setting. \tabref{NNBench} above provides more details.

We also considered a CNN discussed in \cite{RiaziWTS0K18} with 2 hidden layers, consisting of 100 and 10 nodes. Similar to ABY3, we {\em overestimate} the running time by replacing the convolutional kernel with a fully connected layer. In LAN, the number of iterations is maximum with a batch of 128, at $10.46$ \#it/sec, and comes down to $2.99$ with a batch size of 512. Similarly, over WAN it is maximum at $13.86$ and comes down to $13.16$.

\subsection{Secure Prediction}
\label{sec:Bench_Prediction}
For Secure Prediction, we use online latency of the protocol as a metric to compare both works. The units are milliseconds in LAN and seconds in WAN. We use the MNIST dataset, which has $784$ features, with a batch size of 1 and 100 for benchmarking. Our truncation protocol causes a bit-error at the least significant bit position, which is the same as that of ABY3 and SecureML~\cite{MohasselZ17} due to similarity in the techniques. We refer the readers to SecureML for a detailed analysis of the bit-error. The accuracy of the prediction itself however, ranges from $93\%$ for linear regression to $98.3\%$ for CNN.

\begin{table}[ht]
	\centering
	\resizebox{.48\textwidth}{!}{
		\begin{tabular}{c|c|c|r|r|r|r}
			\toprule
			Network & \makecell{Batch\\Size} & Ref. 
			& \makecell{Linear\\Regression} & \makecell{Logistic\\Regression} & \makecell{NN} & \makecell{CNN} \\ \midrule
			\multirow{4}[3]{*}{\makecell{LAN\\($ms$)}} 
			& \multirow{2}{*}{1}    & ABY3        & $2.08$  & $6.25$   & $73.09$   & $371.1$\\
			&                       & {\bf This}  & $0.25$  & $1.75$   & $4.51$    & $5.4$\\
			\cmidrule{2-7} 
			& \multirow{2}{*}{100}  & ABY3        & $37.80$ & $49.68$  & $1284.95$ & $2010.06$\\
			&                       & {\bf This}  & $0.30$  & $2.55$   & $17.17$   & $39.63$\\
			\midrule
			\multirow{4}[3]{*}{\makecell{WAN\\($s$)}} 
			& \multirow{2}{*}{1}    & ABY3        & $0.47$  & $2.77$   & $6.02$    & $6.25$\\
			&                       & {\bf This}  & $0.16$  & $0.93$   & $2.31$    & $2.31$\\
			\cmidrule{2-7} 
			& \multirow{2}{*}{100}  & ABY3        & $0.49$  & $2.79$   & $7.04$    & $7.5$\\
			&                       & {\bf This}  & $0.16$  & $0.93$   & $2.31$    & $2.32$\\
			\bottomrule
		\end{tabular}
	}
	\caption{\small Online Runtime of ABY3 (Malicious) and {\bf This} for Secure Prediction of Linear, Logistic, NN, and CNN models for $d = 784$. (lower = better).}\label{tab:Inf_Bench}
	\vspace{-3mm}
\end{table}

For Linear Regression, our improvement ranges from $3 \times$ to $126 \times$, considering LAN and WAN together. For Logistic Regression, our improvement ranges from $3 \times$ to $19.48 \times$, considering LAN and WAN together. In NN, we achieve an improvement ranging from $3.05 \times$ to $74.85 \times$. Similarly for CNN, the improvement ranges from $2.71 \times$ to $68.82 \times$.

Though not stated explicitly, our offline cost for linear regression is orders of magnitude more efficient as compared to ABY3. This improvement carries over for logistic regression, NN, and CNN networks as well. A large part of this improvement comes from the difference in the approaches to truncation. ABY3's approach entails using Ripple Carry Adder (RCA) circuits, which consume 128 rounds. ABY3 has pointed out that this was the reason SecureML performed better in total time for a single prediction. Our approach on the other hand, does not use any such circuit, resulting in an improvement of $\approx 15 \times$ in communication and $64 \times$ in rounds. 

\paragraph*{Throughput Comparison}
We use a different metric to better illustrate the impact of our efficiency in the case of secure prediction, which is online throughput. Online throughput over LAN is the number of predictions that can be made in a second, and over WAN it is the number of predictions per minute. We have a total of $32$ threads over $4$ CPU cores, wherein each thread can perform $100$ queries simultaneously without reduction in performance. \tabref{Inf_TP} provides the online throughput comparison of ABY3 and ours for secure prediction over real-world datasets in a LAN setting. The gains for linear regression range from $26.16 \times$ to $145.18 \times$ and from $5.69 \times$ to $158.40 \times$ for logistic regression. Similarly, we observed gains of $335.44 \times$ and $598.44 \times$ for NN and CNN respectively.

\begin{table}[ht]
	\centering
	\resizebox{.485\textwidth}{!}{
		\begin{tabular}{c|r|r|r|r|r|r|r|r}
			\toprule
			\multirow{2}[2]{*}{Ref.} 
			& \multicolumn{3}{c}{Linear Regression}  & \multicolumn{3}{|c|}{Logistic Regression} 
			& \multicolumn{1}{c}{NN} & \multicolumn{1}{|c}{CNN} \\ \cmidrule{2-9}
			& {\bf BT} & {\bf WR} & {\bf CI} 	& {\bf CD} & {\bf EP} & {\bf RE} & \multicolumn{2}{c}{\bf MNIST}\\ \midrule
			ABY3        & $4.08$    & $1.74$   & $0.73$    & $2.20$   & $0.29$    & $0.08$  & $0.46$   & $0.06$ \\
			{\bf This}  & $106.67$  & $106.67$ & $106.67$  & $12.55$  & $12.55$   & $12.55$ & $153.39$ & $37.43$\\
			\bottomrule
		\end{tabular}
	}
	\caption{\small Online Throughput Comparison of ABY3 (Malicious) and {\bf This} for Secure Prediction over LAN. (higher = better)}\label{tab:Inf_TP}
	\vspace{-3mm}
\end{table}

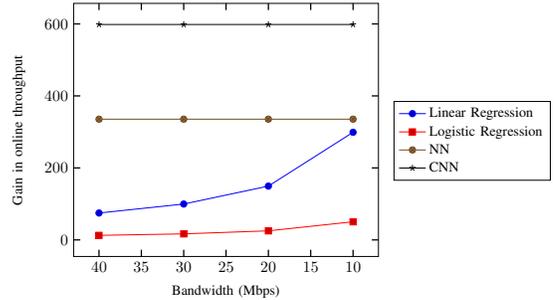
\begin{figure}[htb!]
	\centering
	\resizebox{.4\textwidth}{!}{
	\begin{tikzpicture}
		\begin{axis}[legend style={at={(1.05,0.3)},anchor=south west}, 
					 legend cell align={left}, 
		             xlabel={\small Bandwidth (Mbps)}, 
		             ylabel={\small Gain in online throughput}, 
		             x dir=reverse]
		\addplot plot coordinates { (40, 74.73) (30, 99.64) (20,149.46) (10, 298.93)};
		\addlegendentry{\small Linear Regression}
		\addplot plot coordinates { (40, 12.60) (30, 16.80) (20,25.20) (10, 50.39)};
		\addlegendentry{\small Logistic Regression}
		\addplot plot coordinates { (40, 335.44) (30, 335.44) (20,335.44) (10, 335.44)};
		\addlegendentry{\small NN}
		\addplot plot coordinates { (40, 598.44) (30, 598.44) (20,598.44) (10, 598.44)};
		\addlegendentry{\small CNN}
		
		\end{axis}
		\node[align=center,font=\bfseries, xshift=2.5em, yshift=-2em] (title) at (current bounding box.north) {};
	\end{tikzpicture}
	}
	\caption{\small Throughput Gain in Low-end Networks.}\label{tab:WANGain}
\end{figure}

In WAN, even though our protocols are more communication efficient as compared to ABY3, we could not fully capitalize on it especially for Linear Regression and Logistic Regression. This is due to the limitation of being able to run only 32 CPU threads in parallel, which amounts to a lot of bandwidth not being utilized. This gap can be closed by introducing more CPU threads into our infrastructure. However, since we could not do this, in order to showcase the efficiency better, we limit the bandwidth and compute the gain in online throughput. As evident from the plot in \boxref{tab:WANGain}, the gain increases as we limit the bandwidth. We observe that our protocols for Linear Regression and Logistic Regression achieve maximum bandwidth utilization at around 1.5 Mbps. On the other hand, NN and CNN utilize the entire bandwidth, even at 40 Mbps. So decreasing the bandwidth does have an effect on the throughput for NN and CNN.

\section{Conclusion}
\label{sec:Conclusion}
In this work, we presented an efficient privacy-preserving machine learning framework for the four-party setting, tolerating at most one malicious corruption. The theoretical improvements over the state-of-the-art 3PC framework of ABY3 were backed up by an extensive benchmarking. The improvements show that the availability of an additional honest party can improve the performance of ML protocols while at the same time, decreasing the total monetary cost of hiring the servers.

An interesting open problem is extending this framework to the $n$ party setting. Improving the security to guaranteed output delivery with a minimal trade-off in the concrete performance is another challenging problem. Another direction would be to try to integrate the protocols proposed in this paper into a compiler such as HyCC~\cite{BuscherD0K018}, which already has support for the two-party framework of ABY.
\subsection*{Acknowledgements}
Ajith Suresh would like to acknowledge financial support from Google PhD Fellowship 2019.

\clearpage
\bibliographystyle{IEEEtran}
\bibliography{main_short}

\appendices
\section{Building Blocks}
\label{app:4PC}
\paragraph{Collision Resistant Hash}
Consider a hash function family $\Hash = \mathcal{K}\times \mathcal{L} \rightarrow \mathcal{Y}$. The hash function $\Hash$ is said to be collision resistant if for all probabilistic polynomial-time adversaries $\Adv$, given the description of $\Hash_k$ where {$k \in_R \mathcal{K}$}, there exists a negligible function $\negl()$ such that $\Pr[ (x_1,x_2) \leftarrow \Adv(k):(x_1 \ne x_2) \wedge \Hash_k(x_1)=\Hash_k(x_2)] \leq \negl(\csec)$, where $m = \mathsf{poly}(\csec)$ and $x_1,x_2 \in_R \{0,1\}^m$.

\paragraph{Shared Key Setup}
Let $F : {0, 1}^{\csec} \times {0, 1}^{\csec} \rightarrow X$ be a secure PRF, with co-domain $X$ being $\Z{\ell}$. The set of keys are:
\begin{myitemize}
	\item[--] One key shared between every pair of parties - $\Key{ij}$ for $(P_i, P_j)$ where $i, j \in \PInSet$.
	\item[--] One key shared between every group of three parties - $\Key{ijk}$ for $(P_i, P_j, P_k)$ where $i, j, k \in \PInSet$.
	\item[--] One key shared amongst all - $\Key{\Partyset}$.
\end{myitemize}

We present the ideal world functionality $\FSETUP$ below.

\begin{systembox}{$\FSETUP$}{Functionality for Shared Key Setup}{fig:FSETUP}
	\justify
	$\FSETUP$ interacts with the parties in $\Partyset$ and the adversary $\Sim$. $\FSETUP$ picks random keys $\Key{ij}$ and $\Key{ijk}$ for $i,j, k \in \{0,1,2,3\}$ and $\Key{\Partyset}$. Let $\wy_i$ denote the keys corresponding to party $P_i$. Then
	\begin{myitemize}
		\item[--] $\wy_i = (k_{01}, k_{02}, k_{03}, k_{012}, k_{013}, k_{023}$ and $\Key{\Partyset})$ when $P_i = P_0$.
		\item[--] $\wy_i = (k_{01}, k_{12}, k_{13}, k_{012}, k_{013}, k_{123}$ and $\Key{\Partyset})$ when $P_i = P_1$.
		\item[--] $\wy_i = (k_{02}, k_{12}, k_{23}, k_{012}, k_{023}, k_{123}$ and $\Key{\Partyset})$ when $P_i = P_2$.
		\item[--] $\wy_i = (k_{03}, k_{13}, k_{23}, k_{013}, k_{023}, k_{123}$ and $\Key{\Partyset})$ when $P_i = P_3$.
	\end{myitemize}
	\begin{description}
		\item {\bf Output to adversary: }  If $\Sim$ sends $\abort$, then send $(\OUTPUT, \bot)$ to all the parties. Otherwise,  send $(\OUTPUT, \wy_i)$ to the adversary $\Sim$, where $\wy_i$ denotes the keys corresponding to the corrupt party.
		
		\item {\bf Output to selected honest parties: } Receive $(\SELECT, \{I\})$ from adversary $\Sim$, where $\{I\}$ denotes a subset of the honest parties. If an honest party $P_i$ belongs to $I$, send $(\OUTPUT, \bot)$, else send $(\OUTPUT, \wy_i)$.
	\end{description}
\end{systembox}

\paragraph{Generating Zero Share, $\piZero$}
Protocol $\piZero$ (\boxref{fig:piZero}) enables parties in $\Partyset$ to generate a $\sgrd$-sharing of zero among $\ESet$. In detail, parties $P_1, P_2$ and $P_3$ obtain values $A, B$ and $\Gamma$ respectively such that $A + B + \Gamma = 0$. In addition, $P_0$ obtains all the values $A, B$ and $\Gamma$. The protocol is adapted to the 4PC setting from the 3PC protocol of \cite{AFLNO16}, and we use $\FZERO$ to denote the ideal-world functionality for the same. We omit the proof for $\piZero$ and refer readers to \cite{AFLNO16} since the protocols are almost similar.

\begin{protocolbox}{$\piZero$}{Generating $\sqd$-sharing of zero}{fig:piZero}
	\justify
	\vspace{2.5mm}
	\begin{myitemize}
		\item[--] Parties use the $\FSETUP$ functionality to establish the following set of keys among them:
	    \begin{align*}
	    \Partyset \setminus \{P_3\} : \Key{2},~~~~~\Partyset \setminus \{P_1\} : \Key{3},~~~~~\Partyset \setminus \{P_2\} : \Key{1}
	    \end{align*}
		\item[--] Using the above set of keys and the PRF $F$, parties compute the following:
		\begin{myitemize}
			\item[--] $P_0, P_1: A = F(\Key{2}) - F(\Key{1})$. 
			\item[--] $P_0, P_2: B = F(\Key{3}) - F(\Key{2})$. 
			\item[--] $P_0, P_3: \Gamma = F(\Key{1}) - F(\Key{3})$. 
		\end{myitemize}
	\end{myitemize}
\end{protocolbox}

\section{Analysis of our 4PC protocol}
\label{app:4PCCosts}
\begin{lemma}[Communication]
	\label{app:piSh}
	Protocol $\piSh$ (\boxref{fig:piSh}) is non-interactive in the offline phase and requires $1$ round and an amortized communication of $3 \ell$ bits in the online phase.
\end{lemma}
\begin{IEEEproof}
	During the offline phase, parties sample the $\pad$-shares non-interactively using the shared key setup. During the online phase, $P_i$ sends $\mask$-value to $\ESet$ resulting in $1$ round and a communication of at most $3 \ell$ bits (for the case when $P_i = P_0$). Parties $\ESet$ then mutually exchange the hash of $\mask$-value received from $P_0$. Parties can combine $\mask$-values for several instances into a single hash and hence the cost gets amortized over multiple instances.
\end{IEEEproof}

\begin{lemma}[Communication]
	\label{app:piaSh}
	Protocol $\piaSh$ (\boxref{fig:piaSh}) requires $1$ round and an amortized communication of $2 \ell$ bits in the offline phase.
\end{lemma}
\begin{IEEEproof}
	Protocol $\piaSh$ is performed entirely in the offline phase. During the protocol, $P_0$ computes and sends $\val_3$ to both $P_1$ and $P_2$ resulting in $1$ round and a communication of $2 \ell$ bits. Parties $P_1, P_2$ mutually exchange hash of the value received from $P_0$ to ensure consistency. Similar to $\piSh$, hash for multiple instances can be combined and hence this cost gets amortized over multiple instances.
\end{IEEEproof}

\begin{lemma}[Communication]
	\label{app:piRec}
	Protocol $\piRec$ (\boxref{fig:piRec}) requires $1$ round and an amortized communication of $4 \ell$ bits in the online phase.
\end{lemma}
\begin{IEEEproof}
	During the protocol, each party receives his/her missing share from another party, resulting in $1$ round and a communication of $4 \ell$ bits. In addition, each party receives a hash of the missing share from another party for verification. The hash for multiple instances can be combined to a single hash and thus this cost gets amortized over multiple instances.
\end{IEEEproof}

\begin{lemma}[Communication]
	\label{app:piMult}
	Protocol $\piMult$ (\boxref{fig:piMult}) requires $1$ round and an amortized communication of $3 \ell$ bits in the offline phase and requires $1$ round and an amortized communication of $3 \ell$ bits in the online phase.
\end{lemma}
\begin{IEEEproof}
	During the offline phase, each $\ESet$ receives one share of $\Gammaxy$ from another party resulting in $1$ round and communication of $3 \ell$ bits. Also, each party receives a hash value from $P_0$ in the same round. The values corresponding to all the multiplication gates can be combined into a single hash resulting in an overall communication of just three hash values, which gets amortized.
	During the online phase, each of $\ESet$ receives one $\sqd$-share of $\Mask{\wz} - \Mask{\wx} \Mask{\wy}$ from another party resulting in $1$ round and a communication of $3 \ell$ bits. Also, each party receives a hash value of the same from the third party. The values corresponding to all the multiplication gates can be combined into a single hash resulting in amortization of this cost.
\end{IEEEproof}

\begin{theorem}
	\label{thm:Pi4PCCostApp}
	$\piFourPC$ requires one round with an amortized communication of $3\MS$ ring elements during the offline phase. In the online phase, $\piFourPC$ requires one round with an amortized communication of at most $3\IS$ ring elements in the Input-sharing stage, $\DF$ rounds with an amortized communication of $3\MS$  ring elements for evaluation stage and one round with an amortized communication of $3\OS$  elements for the output-reconstruction stage.
\end{theorem}
\begin{IEEEproof}
	The proof follows from Lemmas \ref{app:piSh}, \ref{app:piMult} and \ref{app:piRec}. 
\end{IEEEproof}

\begin{lemma}[Communication]
	\label{app:pifairRec}
	Protocol $\pifRec$ (\boxref{fig:pifRec}) requires $4$ rounds and an amortized communication of $8 \ell$ bits in the online phase.
\end{lemma}
\begin{IEEEproof}
	In the first round, $P_0$ receives a single bit from parties $\ESet$. In the next round, $P_0$ sends back a single bit to $\ESet$. In the third round, $\ESet$ mutually exchanges the bit received from $P_0$. In the last round, each party receives the missing share from two other parties resulting in the communication of $8 \ell$ bits. In parallel, each party receives a hash of the missing share from the third party. Note that the first three rounds can be performed only once, for all the output wires together, amortizing the corresponding cost. In the last round, parties can compute a single hash corresponding to all the output wires resulting in getting this cost amortized over all the instances.  
\end{IEEEproof}
\section{Analysis of Conversions}
\label{app:ConvCosts}
\subsection{Building Blocks}
\begin{lemma}[Communication]
	\label{app:pivSh}
	Protocol $\pivSh$ (\boxref{fig:pivSh}) is non-interactive in the offline phase and requires $1$ round and an amortized communication of at most $2 \ell$ bits in the online phase.
\end{lemma}
\begin{IEEEproof}
	The cost for the offline phase follows from Lemma~\ref{app:piSh}. In the online phase, parties $P_i, P_j$ sends the masked value to $\ESet$. For the case when $P_0 \in \{P_i, P_j\}$, this requires $1$ round and a communication of $2 \ell$ bits. In other cases, communication of just $\ell$ bits is required. Note that the values corresponding to multiple instances of $\pivSh$ can be combined into a single hash, resulting in amortization of this cost.
\end{IEEEproof}

\begin{lemma}[Communication]
	\label{app:pivShG}
	Protocol $\pivShG$ (\boxref{fig:pivShG}) is non-interactive in the offline phase, while it requires $1$ round and an amortised communication of $\kappa$ bits in the online phase.
\end{lemma}
\begin{IEEEproof}
	For the case when $P_0$ along with one of the garblers $\ESet$ owns $\val$, the protocol requires two commitments and one decommitment. ABY3\cite{MR18} has shown that the number of commitments can be limited to $2 \sparam$, when the number of values to be shared is larger than the statistical security parameter $\sparam$. Consequently, the {\it amortized} cost per instance of $\pivShG$  becomes $\kappa$ bits (where the garbler sends the key $\KeyGAct{\val}$ to $P_0$). For the case when two of the garblers need to share multiple values, one garbler can combine all the actual keys to a single hash and send it to $P_0$, resulting in an amortized cost of $\kappa$ bits.
\end{IEEEproof}

\begin{lemma}[Communication]
	\label{app:piDotP}
	Protocol $\piDotP$ (\boxref{fig:piDotP}) requires $1$ round and an amortized communication of $3 \ell$ bits in the offline phase and requires $1$ round and an amortized communication of $3 \ell$ bits in the online phase.
\end{lemma}
\begin{IEEEproof}
	The proof follows from Lemma~\ref{app:piMult} since the protocol is similar to $\piMult$.
\end{IEEEproof}

\subsection{Sharing Conversions}
We use $|\Add|$ and $|\Sub|$ to denote the size of garbled circuits corresponding to two $\ell$-bit input adder and subtractor circuit respectively. We abuse the notation $|\Decode|$ to denote the size of decoding information for the corresponding garbled circuit. 
The lemmas in this section provide the amortized cost and omit the cost of hash values.
\begin{lemma}[Communication]
	\label{app:GB}
	Protocol $\PiGB$ (\boxref{fig:GB}) requires $1$ round and a communication of $\kappa + 1 + |\Decode|$ bits in the offline phase, while it requires $1$ round and a communication of $3$ bits in the online phase.
\end{lemma}
\begin{IEEEproof}
	During the offline phase, parties execute one instance of $\pivShG$ and $\pivShB$ resulting in $1$ round and a communication of $\kappa + 1$ bits (Lemmas \ref{app:pivSh} and \ref{app:pivShG}). Also, the decoding information for performing an XOR in the garbled world needs to be communicated to $P_0$. During the online phase, $P_0$ communicates a single bit to $P_3$. Also, $P_0$ performs the boolean sharing of the same bit resulting in a total communication of $3$ bits. The verification by $P_3$ can be pushed to the next round as long as the values are not revealed in the next round. Thus parties can proceed with the evaluation after the first round.
\end{IEEEproof}

\begin{lemma}[Communication]
	\label{app:GA}
	Protocol $\PiGA$ (\boxref{fig:GA}) requires $1$ round and a communication of $\ell \kappa + \ell + |\Sub| + |\Decode|$ bits in the offline phase, while it requires $1$ round and a communication of $3 \ell$ bits in the online phase.
\end{lemma}
\begin{IEEEproof}
	The protocol is similar to that of $\PiGB$ (Lemma~\ref{app:GB}) with the main difference being a garbled subtractor circuit $\Sub$ is used for evaluating the final output.
\end{IEEEproof}

\begin{lemma}[Communication]
	\label{app:BG}
	Protocol $\PiBG$ (\boxref{fig:BG}) requires $1$ round and a communication of $\kappa$ bits in the offline phase, while it requires $1$ round and a communication of $\kappa$ bits in the online phase.
\end{lemma}
\begin{IEEEproof}
	The protocol $\PiBG$ involves the execution of an instance of protocol $\pivShG$ in both the offline and online phases. The cost follows directly from Lemma ~\ref{app:pivShG}. 
\end{IEEEproof}

\begin{lemma}[Communication]
	\label{app:AG}
	Protocol $\PiAG$ (\boxref{fig:AG}) requires $-$ rounds and a communication of $\ell \kappa + |\Sub|$ bits in the offline phase, while it requires $1$ round and a communication of $\ell \kappa$ bits in the online phase.
\end{lemma}
\begin{IEEEproof}
	The protocol can be viewed as $\ell$ instances of $\PiBG$ (Lemma~\ref{app:BG}), where each bit of $\val$ is converted to its garbled sharing. Moreover, a garbled subtractor circuit $\Sub$ is used to evaluate the output.
\end{IEEEproof}

\begin{lemma}[Communication]
	\label{app:AB}
	Protocol $\PiAB$ (\boxref{fig:AB}) requires $1$ round and a communication of $3 \ell \log \ell + 2 \ell$ bits in the offline phase, while it requires $1 + \log \ell$ rounds and a communication of $3 \ell \log \ell + \ell$ bits in the online phase.
\end{lemma}
\begin{IEEEproof}
	The protocol proceeds similarly to that of $\PiAG$ apart from the garbled world being replaced with the boolean world. Parties execute a single instance of $\pivShB$ in both the offline and online phases. This results in one round and a communication of $2 \ell$ bits in the offline phase, while it results in one round and communication of $\ell$ bits in the online phases (Lemma~\ref{app:pivSh}). Moreover, parties evaluate a boolean subtractor circuit $\Sub$ (Parallel Prefix Adder version mentioned in ABY3~\cite{MR18}) of $\log \ell$ multiplicative depth and contain $\ell \log \ell$ AND gates. This results in an additional communication of $3 \ell \log \ell$ bits in the offline and online phases (Lemma~\ref{app:piMult}). Moreover, the online rounds increases from $1$ to $1 + \log \ell$.
\end{IEEEproof}

\begin{lemma}[Communication]
	\label{app:BitA}
	Protocol $\PiBitA$ (\boxref{fig:BitA}) requires $2$ rounds and a communication of $3 \ell + 1$ bits in the offline phase, while it requires $1$ round and a communication of $3 \ell$ bits in the online phase.
\end{lemma}
\begin{IEEEproof}
	During the offline phase, $P_0$ executes $\piaSh$ on $\vu$ resulting in one round and a communication of $2 \ell$ bits (Lemma~\ref{app:piaSh}). This is followed by parties $\ESet$ performing the check to ensure the correctness of sharing done by $P_0$. During the check, $P_1$ sends one ring element and a bit to $P_3$, while $P_2$ sends a hash value. This results in an additional round and an amortized communication of $\ell + 1$ bits. 
	
	During the online phase, parties non-interactively generate the $\shrd$-shares of $\val$. This is followed by one arithmetic multiplication, resulting in one round and communication of $3 \ell$ bits in the online phase (Lemma~\ref{app:piMult}). Here, note that the offline phase for the multiplication is not required, since $\Pad{\vv}$ is set to $0$ while executing the protocol $\piSh(P_1, P_2, P_3, \vv)$, resulting in $\gamma_{\vu \vv}$ to be set to $0$.
\end{IEEEproof}

\begin{lemma}[Communication]
	\label{app:BA}
	Protocol $\PiBA$ (\boxref{fig:BA}) requires $2$ rounds and a communication of $3 {\ell}^2 + \ell$ bits in the offline phase, while it requires $1$ round and a communication of $3 \ell$ bits in the online phase.
\end{lemma}
\begin{IEEEproof}
	The offline cost follows from that of protocol $\PiBitA$ since the offline phase can be viewed as $\ell$ instances of that of $\PiBitA$ (Lemma~\ref{app:BitA}). During the online phase, every pair from $\ESet$ executes an instance of $\pivSh$ resulting in one round and communication of $3\ell$ bits (Lemma~\ref{app:pivSh}).
\end{IEEEproof}

\begin{lemma}[Communication]
	\label{app:BitInj}
	Protocol $\PiBitInj$ (\boxref{fig:BitInj}) requires $2$ rounds and a communication of $6 \ell + 1$ bits in the offline phase, while it requires $1$ round and a communication of $3 \ell$ bits in the online phase.
\end{lemma}
\begin{IEEEproof}
	The offline phase of $\PiBitInj$ consists of generating $\sgrd$-shares of $\Pad{\bitb}$ and $\Pad{\bitb} \Pad{\val}$. The case for $\Pad{\bitb}$ is same as that of $\PiBitA$ resulting in two rounds and a communication of $3\ell + 1$ bits. In parallel, $P_0$ execute $\piaSh$ on $\Pad{\bitb} \Pad{\val}$ resulting in an additional communication of $2 \ell$ bits. In order to verify the correctness of this sharing, $\ESet$ performs a check. During the check, $P_1$ sends one ring element to $P_3$, while $P_2$ sends a hash value, resulting in an additional amortised communication of $\ell$ bits. Thus the offline phase requires two rounds and an amortised communication of $6\ell + 1$ bits.
	
	During the online phase, every pair from $\ESet$ executes an instance of $\pivSh$ resulting in one round and communication of $3\ell$ bits (Lemma~\ref{app:pivSh}).
\end{IEEEproof}
\vspace{2mm}
\paragraph*{Cost Comparison}
\tabref{ConvM} provides a comparison of our sharing conversions with ABY3\cite{MR18}. Here, $\GarA$ denotes a two input garbled subtractor circuit, while $\GarB = \Garb{2, Sub, \ell}$ denotes two input garbled subtractor circuit along with its decoding information. Similarly, $\GarC = \Garb{3, Sub, \ell}$ and $\GarD = \Garb{3, Adder, \ell}$ denote three input garbled circuit for subtraction and addition respectively. Here $\ell$ denotes the ring size in bits.

\begin{table}[ht]
	\resizebox{.48\textwidth}{!}{
	\begin{tabular}{l|l|r|r|r|r}
		\toprule
		\multirow{2}{*}{Conv.} & \multirow{2}{*}{Work} 
		& \multicolumn{2}{c}{Offline} 
		& \multicolumn{2}{|c}{Online}\\
		\cmidrule{3-6}
		& & R. & Comm. & R. & Comm.\\
		\midrule
		\multirow{2}{*}{$\GB$} 
		& ABY3        & $1$ & $\kappa$   & $1$ & $\kappa$\\
		& {\bf This}  & $1$ & $\kappa + 1 + |\Decode|$ & $1$ & $3$\\
		\midrule
		\multirow{2}{*}{$\GA$} 
		& ABY3        & $1$ & $|\GarC| + \ell \kappa$   & $1$ & $2 \ell\kappa$\\ 
		& {\bf This}  & $1$ & $\ell \kappa + \ell + |\GarB|$ & $1$ & $3 \ell$\\
		\midrule
		\multirow{2}{*}{$\BG$} 
		& ABY3        & $0$ & $0$        & $1$ & $2 \kappa$ \\
		& {\bf This}  & $1$ & $\kappa$ & $1$ & $\kappa$\\
		\midrule
		\multirow{2}{*}{$\AG$} 
		& ABY3        & $1$ & $|\GarD|$                 & $1$ & $2 \ell \kappa$\\
		& {\bf This}  & $1$ & $\ell \kappa + |\GarA|$ & $1$ & $\ell \kappa$\\
		\midrule
		\multirow{2}{*}{$\AB$} 
		& ABY3        & $3$             & $12\ell\log \ell + 12\ell$  & $1 + \log \ell$ & $9\ell\log \ell + 9\ell$\\ 
		& {\bf This}  & $1$ & $3 \ell \log \ell + 2 \ell$ & $1 + \log \ell$ & $3 \ell \log \ell + \ell$\\
		\midrule
		\multirow{2}{*}{$\BitA$} 
		& ABY3        & $1$ & $24 \ell$ & $2$ & $18 \ell$\\
		& {\bf This}  & $2$ & $3 \ell + 1$ & $1$ & $3 \ell$\\
		\midrule
		\multirow{2}{*}{$\BA$} 
		& ABY3        & $3$             & $12\ell\log \ell + 12\ell$  & $1 + \log \ell$ & $9\ell\log \ell + 9\ell$\\ 
		& {\bf This}  & $2$ & $3 {\ell}^2 + \ell$ & $1$ & $3 \ell$\\
		\midrule
		\multirow{2}{*}{$\BitInj$} 
		& ABY3        & $1$ & $36\ell$  & $3$ & $27\ell$\\
		& {\bf This}  & $2$ & $6 \ell + 1$ & $1$ & $3 \ell$\\
		\bottomrule
	\end{tabular}
	}
	\caption{Sharing conversions of ABY3 and Ours.}\label{tab:ConvM}
\end{table}
\section{Analysis of ML protocols}
\label{app:PrivMLCosts}
\begin{lemma}[Correctness]
	\label{app:piMultTrC}
	In the offline phase of protocol $\piMultTr$ (\boxref{fig:piMultTr}), if a corrupt $P_0$ generates incorrect $\shr{\vrt}$ sharing, then the honest evaluators $\ESet$ $\abort$.
\end{lemma}
\begin{IEEEproof}
To see the correctness, it suffices to show that $\vm_1 + \vm_2 = \vc$ where $\vm_1 = \vr_2 - 2^{\vd} \vrt_2 - \vr_{\vd,2} + \vc$ and $\vm_2 = (\vr_1 + \vr_3) - 2^{\vd} (\vrt_1 + \vrt_3) - (\vr_{\vd,1} + \vr_{\vd,3})$. Note that $\vr = 2^{\vd} \vrt + \vr_{\vd}$ where $\vrt$ denoted the truncated value of $\vr$ and $\vr_{\vd}$ denoted the last $d$ bits of $\vr$. Then,
\begin{align*}
	\vm_1 + \vm_2 &= \left( \vr_2 - 2^{\vd} \vrt_2 - \vr_{\vd,2} + \vc \right)\\
	              &~~~+ \left( (\vr_1 + \vr_3) - 2^{\vd} (\vrt_1 + \vrt_3) - (\vr_{\vd,1} + \vr_{\vd,3}) \right)\\
	              &= (\vr_1 + \vr_2 + \vr_3) - 2^{\vd} (\vrt_1 + \vrt_2 + \vrt_3)\\
	              &~~~- (\vr_{\vd,1} + \vr_{\vd,2} + \vr_{\vd,3}) + \vc\\
	              &= (\vr) - (2^{\vd} \vrt + \vr_{\vd}) + \vc = 0 + \vc = \vc
\end{align*}
\end{IEEEproof}

\begin{lemma}[Communication]
	\label{app:piMultTr}
	Protocol $\piMultTr$ (\boxref{fig:piMultTr}) requires $2$ rounds and a communication of $6 \ell$ bits in the offline phase, while it requires $1$ round and a communication of $3 \ell$ bits in the online phase.
\end{lemma}
\begin{IEEEproof}
	During the offline phase, the offline phase of $\piMult$ is executed, resulting in one round and communication of $3 \ell$ bits. In parallel, $P_0$ executes $\piaSh$ on $\vrt$ resulting in an additional communication of $2 \ell$ bits (Lemma~\ref{app:piaSh}). To verify the correctness of this sharing, parties $\ESet$ performs a check, where $P_1$ sends one ring element and hash value to $P_2$. This results in an additional amortized communication of $\ell$ bits. Thus the offline phase requires two rounds and an amortized communication of $6 \ell$ bits. The cost for online phase follows directly from $\piMult$ (Lemma~\ref{app:piMult})
\end{IEEEproof}

\begin{lemma}[Communication]
	\label{app:piBitExt}
	Protocol $\piBitExt$ (\boxref{fig:piBitExt}) requires $1$ round and a communication of $4 \ell + 1$ bits in the offline phase, while it requires $3$ rounds and a communication of $5 \ell + 2$ bits in the online phase.
\end{lemma}
\begin{IEEEproof}
	During the offline phase, parties execute one instance each of $\pivSh$ and $\pivShB$ resulting in one round and a communication of $\ell + 1$ bits (Lemma~\ref{app:pivSh}).
	Also, the offline phase for multiplication is performed resulting in an additional communication of $3 \ell$ bits.
	
	During the online phase, parties first execute an arithmetic multiplication, resulting in one round and communication of $3 \ell$ bits (Lemma~\ref{app:piMult}). The value $\vr \val$ is reconstructed towards both $P_0$ and $P_3$, resulting in an additional round and an amortized communication of $2 \ell$ bits. This is followed by the last round, where parties execute one instance of $\pivShB$ resulting in a communication of $2$ bits. Thus the online phase requires three rounds and an amortized communication of $5 \ell + 2$ bits.
\end{IEEEproof}

\begin{lemma}[Communication]
	\label{app:piReLU}
	Protocol ReLU ($\piReLU$) requires $3$ rounds and a communication of $8 \ell + 2$ bits in the offline phase, while it requires $4$ rounds and a communication of $8 \ell + 2$ bits in the online phase.
\end{lemma}
\begin{IEEEproof}
	The cost follows directly from Lemmas \ref{app:piBitExt} and \ref{app:BitInj}.
\end{IEEEproof}

\begin{lemma}[Communication]
	\label{app:piSig}
	Protocol Sigmoid ($\piSig$) requires $3$ rounds and a communication of $15 \ell + 7$ bits in the offline phase, while it requires $5$ rounds and a communication of $16 \ell + 7$ bits in the online phase.
\end{lemma}
\begin{IEEEproof}
	The cost follows directly from Lemmas \ref{app:piBitExt}, \ref{app:piMult}, \ref{app:BitA} and \ref{app:BitInj}.
\end{IEEEproof}

\paragraph*{Cost Comparison}
\tabref{ConvMLM} provides a comparison of our work with ABY3\cite{MR18}, in terms of ML protocols and special conversions.
\begin{table}[htb!]
	\resizebox{.48\textwidth}{!}{
	\begin{tabular}{l|l|r|r|r|r}
		\toprule
		\multirow{2}[2]{*}{Conv.} & \multirow{2}[2]{*}{Work} 
		& \multicolumn{2}{c}{Offline} 
		& \multicolumn{2}{|c}{Online}\\
		\cmidrule{3-6}
		& & R. & Comm. & R. & Comm.\\
		\midrule
		\multirow{2}{*}{$\piMultTr$} 
		& ABY3        & $2\ell -2$ & $96\ell - 42d - 84$ & $1$ & $12\ell$\\
		& {\bf This}  & $2$ & $6 \ell$ & $1$ & $3 \ell$\\
		\midrule
		\multirow{2}{*}{$\piBitExt$} 
		& ABY3        & $1$ & $24 \ell \log \ell$ & $\log \ell$ & $18 \ell \log \ell$\\
		& {\bf This}  & $1$ & $4 \ell + 1$ & $3$ & $5 \ell + 2$\\
		\midrule
		\multirow{2}{*}{ReLU} 
		& ABY3        & $3$ & $60\ell$    & $3 + \log\ell$ & $45\ell$\\
		& {\bf This}  & $3$ & $8 \ell + 2$ & $4$ & $8 \ell + 2$\\
		\midrule
		\multirow{2}{*}{Sigmoid} 
		& ABY3        & $3$ & $108\ell+12$    & $4 + \log \ell$ & $81 \ell + 9$\\
		& {\bf This}  & $3$ & $15 \ell + 7$ & $5$ & $16 \ell + 7$\\
		\bottomrule
	\end{tabular}
	}
	\caption{ML conversions of ABY3 and Ours. $d$ denotes the number of features}\label{tab:ConvMLM}
\end{table}
\section{Benchmarking}
\label{app:Bench}
\subsection{Motivating 4PC}
\label{app:Comp_Motivation}

In this section, we justify the reason for operating in the 4PC setting over the 3PC setting through the lens of benchmarking and practical efficiency. We begin with the comparison of our 4PC secret sharing scheme with that of Gordon et al.\cite{GordonR018}, which shows the reason we built a new 4PC. In Section~\ref{app:Comp_TotTime} we compare the efficiency of our framework with ABY3. Since machine learning models take a long time to train, monetary cost becomes an important factor to consider when building PPML protocols, as shown in ~\cite{PinkasRTY19}. 

Monetary cost is computed by calculating the total running time for each of the servers and the cost of hiring them, which is based upon the computing power of the server. Total running time is the total time taken for the evaluation phase, which excludes the input sharing and reconstruction phase. So, if we have a protocol that communicates very less but the total running time for the servers is high due to the computation, the monetary cost will also be high. We compare the total running times of our protocol against the others to show that we have a better balance in terms of computation and communication, making our monetary cost lower than the rest.

\begin{table}[ht]
	\centering
	\resizebox{.4\textwidth}{!}{
		\begin{tabular}{l|r|r|r|r|r}
			\toprule
			Ref.        & $P_0$  & $P_1$    & $P_2$    & $P_3$  & Total\\ \midrule
			Gordon      & $7.84$ & $3.13$   & $7.34$   & $3.21$ & $21.52$  \\
			{\bf This}  & $0.00$ & $6.19$   & $6.19$   & $3.81$ & $16.19$  \\
			\bottomrule
		\end{tabular}
	}
	\caption{\small Total Online Runtime (in seconds) of Gordon et al. and {\bf This} for evaluation on an AES-128 circuit (lower = better) over WAN.}\label{tab:Total_Runtime4PC}
\end{table}

\subsubsection{Comparison with the protocol of Gordon et al.~\cite{GordonR018}}
\label{app:Comp_GordonProt}

In Table~\ref{tab:Total_Runtime4PC}, we compare the total online runtime of our 4PC protocol with that of Gordon et al.~\cite{GordonR018} for evaluating an AES-128~\cite{AESBristol} circuit. As evident from the table, $P_0$ does not have to be active during the online phase, except for the sharing and the reconstruction phase. This means we can shut down the server that corresponds to $P_0$ for the entirety of the online evaluation phase, saving a lot in terms of monetary cost. 
The preprocessing phase for both protocols requires just 1 round of interaction.

\subsubsection{Motivation of 4PC for ML}
\label{app:Comp_TotTime}

Here, we argue that even though we operate in the 4PC setting, meaning we have 4 servers active instead of 3 as in ABY3, our total monetary cost is still lower than that of ABY3. Table~\ref{tab:Total_Runtime} shows that our total runtime for both training and prediction phases in the malicious case is lower for all the algorithms considered. For the training phase, we use a batch size of 128. The number of features for both training and prediction is 784.

\begin{table}[ht]
	\centering
	\resizebox{.48\textwidth}{!}{
		\begin{tabular}{c|c|r|r|r|r}
			\toprule
			Phase & Ref. 
			& \makecell{Linear\\Regression} & \makecell{Logistic\\Regression} & \makecell{NN} & \makecell{CNN} \\ \midrule
			\multirow{2}{*}{\makecell{Training\\($s$)}} 
			& ABY3        & $2.01$  & $8.92$   & $38.41$    & $41.45$\\
			& {\bf This}  & $0.92$  & $3.76$   & $13.07$    & $13.19$\\
			\midrule
			\multirow{2}{*}{\makecell{Prediction\\($s$)}} 
			& ABY3        & $1.45$  & $8.36$   & $21.12$    & $22.48$\\
			& {\bf This}  & $0.44$  & $2.74$   & $6.90$     & $6.93$\\
			\bottomrule
		\end{tabular}
	}
	\caption{\small Total Online Runtime (in seconds) of ABY3 (Malicious) and {\bf This} for Training and Prediction of Linear, Logistic, NN, and CNN models for $d = 784$ (lower = better) over a WAN setting.}\label{tab:Total_Runtime}
\end{table}

\subsection{Comparison with the ML framework of ABY3~\cite{MR18} in the semi-honest setting}
\label{app:Comp_ABY3Semi}
We compare the performance of our protocol with the semi-honest version of ABY3, giving them an advantage in terms of the threat model. We use ABY3S to denote the ABY3 protocol in the semi-honest setting. Our performance is the same for linear regression but as the protocols get more complex, the difference in performance increases in our favour. This is because in linear regression though we have the additional overhead of a hash, the cost is amortized, making our cost the same as ABY3S. For the other three algorithms, however, we outperform ABY3S due to our efficient protocols for bit extraction, bit injection, and Bit2A. Although we are worse than ABY3 in the offline phase due to our preprocessing for multiplication, our online phase is a lot more efficient for both training and prediction phases. 

\begin{table}[htb!]
\centering
\resizebox{.48\textwidth}{!}{
	\begin{tabular}{c|l|r|r}
		\toprule
		Algorithm & Ref. & LAN ($\#$it/sec) & WAN ($\#$it/min) \\
		\midrule
		\multirow{2}{*}{\makecell[l]{Linear\\Regression}} 
		& ABY3S        & $1098.90$ 	 & $195.13$ \\
		& {\bf This}   & $1098.90$ 	 & $195.13$ \\ \midrule
		\multirow{2}{*}{\makecell[l]{Logistic\\Regression}} 
		& ABY3S        & $90.29$ 	 & $35.48$ \\
		& {\bf This}   & $307.41$ 	 & $55.75$ \\ \midrule
		\multirow{2}{*}{\makecell[l]{Neural\\Networks}} 
		& ABY3S        & $1.01$ 	 & $8.13$ \\
		& {\bf This}   & $23.00$ 	 & $13.94$ \\ \midrule
		\multirow{2}{*}{\makecell[l]{CNN~~~~~~}} 
		& ABY3S        & $0.37$ 	 & $7.13$ \\
		& {\bf This}   & $10.46$ 	 & $13.86$ \\
		\bottomrule
	\end{tabular}
}
\caption{\small Comparison of Online Phase of ABY3 (Semi-Honest) and {\bf This} for ML Training (higher = better)}\label{tab:CompABY3Semi}
\end{table}

In Table~\ref{tab:CompABY3Semi}, we compare the online phase of both protocols for ML training, in terms of the number of iterations per second they can process. In Table~\ref{tab:Inf_BenchSemi} and Table~\ref{tab:CompABY3InfTPSemi} we compare the online phase for prediction through two benchmarking units, one being the runtime and the other being the throughput, which is the number of queries we can process per second.

\begin{table}[htb!]
	\centering
	\resizebox{.48\textwidth}{!}{
		\begin{tabular}{c|c|r|r|r|r}
			\toprule
			Network & Ref. 
			& \makecell{Linear\\Regression} & \makecell{Logistic\\Regression} & \makecell{NN} & \makecell{CNN} \\ \midrule
			\multirow{2}{*}{\makecell{LAN\\($ms$)}} 
			& ABY3S        & $0.30$  & $9.14$   & $480.81$  & $1185.70$\\
			& {\bf This}  & $0.30$  & $2.55$   & $17.17$   & $39.63$\\
			\midrule
			\multirow{2}{*}{\makecell{WAN\\($s$)}} 
			& ABY3S        & $0.16$  & $1.54$   & $4.07$    & $4.47$\\
			& {\bf This}  & $0.16$  & $0.93$   & $2.31$    & $2.31$\\
			\bottomrule
		\end{tabular}
	}
	\caption{\small Online Runtime of ABY3 (Semi-honest) and {\bf This} for Secure Prediction of Linear, Logistic, NN, and CNN models for $d = 784$ (lower = better)}\label{tab:Inf_BenchSemi}
	\vspace{-5mm}
\end{table}

\begin{table}[htb!]
	\centering
	\resizebox{.44\textwidth}{!}{
		\begin{tabular}{c|l|r|r}
			\toprule
			Algorithm & Ref. & \makecell{LAN\\(queries/sec)} & \makecell{WAN\\(queries/min)} \\
			\midrule
			\multirow{2}{*}{\makecell[l]{Linear\\Regression}} 
			& ABY3S        & $106666.67$ 	 & $12488.62$ \\
			& {\bf This}   & $106666.67$ 	 & $12488.62$ \\ \midrule
			\multirow{2}{*}{\makecell[l]{Logistic\\Regression}} 
			& ABY3S        & $3512.62$ 	     & $1248.86$ \\
			& {\bf This}   & $12549.02$ 	 & $2081.46$ \\ \midrule
			\multirow{2}{*}{\makecell[l]{Neural\\Networks}} 
			& ABY3S        & $66.41$ 	 & $211.18$ \\
			& {\bf This}   & $153.39$ 	 & $368.13$ \\ \midrule
			\multirow{2}{*}{\makecell[l]{CNN}} 
			& ABY3S        & $21.47$ 	 & $51.54$ \\
			& {\bf This}   & $37.43$ 	 & $89.84$ \\
			\bottomrule
		\end{tabular}
	}
	\caption{\small Comparison of Online TP of ABY3 (Semi-Honest) and {\bf This} for Secure Prediction (higher = better)}\label{tab:CompABY3InfTPSemi}
\end{table}

\section{Security of our Constructions}
\label{app:4PCProof}
In this section, we prove the security of our $\piFourPC$ protocol in the $\{\FSETUP, \FZERO\}$-hybrid model, using the real world-ideal world paradigm. The ideal world functionality that realises $\piFourPC$ is given in \boxref{fig:FFOURPC}.

\begin{systembox}{$\FFOURPC$}{Ideal world 4PC functionality}{fig:FFOURPC}
	\justify $\FFOURPC$ interacts with the parties in $\Partyset$ and the adversary $\Sim$ and is parameterized by a $4$-ary function $f$, represented by a publicly known arithmetic circuit $\ckt$ over $\Z{\ell}$.
	\begin{description}
		\item {\bf Input: } Upon receiving the input $\wx_, \ldots, \wx_{\IS}$ from the respective parties in $\Partyset$, do the following: if $(\INPUT, *)$ message was received from $P_i$ corresponding to $\wx_j$, then ignore. Otherwise record $\wx_j' = \wx_j$ internally. If $\wx_j' \ne \Z{\ell}$, consider $\wx_j' = \abort$.
		\item {\bf Output to adversary: } If there exists $j \in \{1, \ldots, \IS \}$ such that $\wx_j' = \abort$, send $(\OUTPUT, \bot)$ to all the parties. Else, send $(\OUTPUT, (\wy_1,\ldots,\wy_{\OS}))$ to the adversary $\Sim$, where $(\wy_1,\ldots,\wy_{\OS}) = f(\wx_1',\ldots,\wx_{\IS}')$.
		\item {\bf Output to selected honest parties: } Receive $(\SELECT, \{I\})$ from adversary $\Sim$, where $\{I\}$ denotes a subset of the honest parties. If an honest party belongs to $I$, send $(\OUTPUT, \bot)$, else send $(\OUTPUT, (\wy_1,\ldots,\wy_{\OS}))$.
	\end{description}
\end{systembox}

We begin with the case when $P_0$ is corrupted. 
\begin{theorem}
	\label{lemma:PiFourPCCaseI}
	In  $\{\FSETUP, \FZERO\}$-hybrid model,  $\piFourPC$ securely realizes the functionality $\FFOURPC$ against a static, malicious adversary $\Adv$, who corrupts $P_0$.
\end{theorem}
\vspace{-2mm}
\begin{IEEEproof}
	Let $\Adv$ be a real-world adversary corrupting the distributor $P_0$ during the protocol $\piFourPC$. We present an ideal-world adversary (simulator) $\SimFourPC$ for $\Adv$ in \boxref{fig:SimFourPCI} that simulates messages for corrupt $P_0$. The only communication to $P_0$ is during the output-reconstruction stage in the online phase. $\SimFourPC$ can easily simulate these messages, with the knowledge of function output and the masks corresponding to the circuit-output wires.
	\begin{simulatorbox}{$\SimFourPC$}{Simulator for the case of corrupt $P_0$}{fig:SimFourPCI}
		\justify The simulator plays the role of the honest parties $P_1, P_2, P_3$ and simulates  each step of $\piFourPC$ to corrupt $P_0$ as follows and finally outputs $\Adv$'s output. The simulator initializes a boolean variable $\flag = 0$, which indicates whether an honest party aborts during the protocol.
		%
		\justify \algoHead{Offline Phase:} $\SimFourPC$ emulates $\FSETUP$ and gives the keys $(k_{01}, k_{02}, k_{03}, k_{012}, k_{013}, k_{023}$ and $\Key{\Partyset})$ to $P_0$. By emulating $\FSETUP$, it learns the $\pad$-masks for all the wires in $\ckt$. 
		\begin{myitemize}
			\item[--] \justify {\em Sharing Circuit-input Values}: Here the simulator has to do nothing, as the offline phase involves no communication.
			\item[--] {\em Gate Evaluation}:  No simulation is needed for the offline phase of an addition gate. For a multiplication gate $\gate$, the simulator emulates $\FZERO$ and gives $A, B, \Gamma$ to $P_0$. It then  receives hash of $\GammaxyA, \GammaxyB$ and $\GammaxyC$ from $P_0$ on behalf of $P_2, P_3$ and $P_1$ respectively. $\SimFourPC$ computes $\GammaxyV{i}$ for $i \in \EInSet$ and sets $\flag = 1$, if any of the hash values received is inconsistent with the values computed. 
			\item[--] {\em Output Reconstruction}:  Here the simulator has to do nothing, as the offline phase involves no communication.
		\end{myitemize}
		\noindent If $\flag = 1$, $\SimFourPC$ invokes $\FFOURPC$ with input $\bot$ on behalf of $P_0$.
		
		\justify \algoHead{Online Phase:} 
		
		\begin{myitemize}
			\item[--] \justify {\em Sharing Circuit-input Values}: For every input $\val_j$ of $P_0$, $\SimFourPC$ receives $\Mask{\val_j}$ of behalf of $\ESet$. $\SimFourPC$ sets $\flag=1$ if the received values mismatch. Else, it computes the input $\val_j$ using $\Mask{\val_j}$ and the $\pad$-masks obtained in the offline phase.
			\item[--] {\em Gate Evaluation}:  No simulation is needed for the online phase of both addition and multiplication gates. 
			\item[--] {\em Obtaining function outputs}:  If $\flag = 1$, $\SimFourPC$ invokes $\FFOURPC$ with input $\bot$ on behalf of $P_0$. Else it sends inputs $\{\val_j\}$ extracted on behalf of $P_0$ to $\FFOURPC$ and receives the function outputs $\vy_1,\ldots, \vy_{\OS}$.
			\item[--] {\em Output Reconstruction}:  For each output $\vy_j$ for $j \in 1,\ldots, \OS$, $\SimFourPC$ computes $\Mask{\vy_j} = \vy_j + \Pad{\vy_j}$, and sends $\Mask{\vy_j}$ and $\Hash(\Mask{\vy_j})$ to $P_0$ on behalf of $P_1$ and $P_2$ respectively. In parallel, he receives $\Hash(\PadV{\vy_j}{i})$ for $i \in \EInSet$ from $P_0$, on behalf of $\ESet$. $\SimFourPC$ initializes the set $I$ to $\emptyset$. $P_i$ for $i \in \EInSet$, is added to $I$, if hash of $\PadV{\vy_j}{i}$ mismatches with the corresponding hash received from $P_0$. $\SimFourPC$ then sends $I$ to $\FFOURPC$ and terminates.
		\end{myitemize}
	\end{simulatorbox}
	\vspace{-3mm}
	The proof now simply follows from the fact that simulated view and real-world view of the adversary are computationally indistinguishable.
\end{IEEEproof}

We next consider the case, when the adversary corrupts one of the evaluators, say $P_1$. The cases of corrupt $P_2$ and $P_3$ are handled symmetrically.
\begin{theorem}
	\label{lemma:PiFourPCCaseII}
	In the $\{\FSETUP, \FZERO\}$-hybrid model,  $\piFourPC$ securely realizes the functionality $\FFOURPC$ against a static, malicious adversary $\Adv$, who corrupts $P_1$.
\end{theorem}
\vspace{-3mm}
\begin{IEEEproof}
	Let $\Adv$ be a real-world adversary corrupting the evaluator $P_1$ during the protocol $\piFourPC$. We now present the steps of the ideal-world adversary (simulator) $\SimFourPC$ for $\Adv$ for this case in \boxref{fig:SimFourPCII}.
	\begin{simulatorbox}{$\SimFourPC$}{Simulator for the case of corrupt $P_1$}{fig:SimFourPCII}
		\justify The simulator plays the role of the honest parties $P_0, P_2, P_3$ and simulates each step of $\piFourPC$ to corrupt $P_1$ as follows and finally outputs $\Adv$'s output. The simulator initializes a boolean variable $\flag = 0$, which indicates whether an honest party aborts during the protocol.
		%
		
		\justify \algoHead{Offline Phase:} $\SimFourPC$ emulates $\FSETUP$ and gives the keys $(k_{01}, k_{12}, k_{13}, k_{012}, k_{013}, k_{123}$ and $\Key{\Partyset})$ to $P_1$. By emulating $\FSETUP$, it learns the $\pad$-masks for all the wires in $\ckt$. 
		\begin{myitemize}
			\item[--] \justify {\em Sharing Circuit-input Values}: Here the simulator has to do nothing, as the offline phase involves no communication.
			\item[--] {\em Gate Evaluation}:  No simulation is needed for the offline phase of an addition gate. For a multiplication gate $\multgate$, the simulator emulates $\FZERO$ and gives $A$ to $P_1$. $\SimFourPC$ computes $\GammaxyV{i}$ for $i \in \EInSet$ and sends $\GammaxyC$ and $\Hash(\GammaxyC)$ to $P_1$ on behalf of $P_2$ and $P_0$ respectively. It receives $\GammaxyB$ from $P_1$ on behalf of $P_3$ and sets $\flag = 1$, if the received value is inconsistent. 
			\item[--] {\em Output Reconstruction}:  Here the simulator has to do nothing, as the offline phase involves no communication.
		\end{myitemize}
		\noindent If $\flag = 1$, $\SimFourPC$ invokes $\FFOURPC$ with input $\bot$ on behalf of $P_0$.
		
		\justify \algoHead{Online Phase:} 
		
		\begin{myitemize}
			\item[--] \justify {\em Sharing Circuit-input Values}: For every input $\val_j$ of $P_1$, $\SimFourPC$ receives $\Mask{\val_j}$ of behalf of $P_2$ and $P_3$. $\SimFourPC$ sets $\flag=1$ if the received values mismatch. Else, it computes the input $\val_j$ using $\Mask{\val_j}$ and the $\pad$-masks obtained in the offline phase. For every input $\val_k$ of $P_i$ for $i \in \{0,2,3\}$, $\SimFourPC$ sets $\val_k = 0$ and sends $\Mask{\val_k} = 0 + \Pad{\val_k}$ to $P_1$ on behalf of $P_i$. $\SimFourPC$ performs the exchange of hash of $\Mask{\val_k}$ honestly.
			\item[--] {\em Gate Evaluation}:  No simulation is needed for the online phase of addition gates. For a multiplication gate $\multgate$, $\SimFourPC$ computes and sends $\sqrA{\Mask{\wz}'}$ and $\Hash(\sqrA{\Mask{\wz}'})$ to $P_1$ on behalf of $P_2$ and $P_3$ respectively. It receives $\Hash(\sqrB{\Mask{\wz}'})$ and $\sqrC{\Mask{\wz}'}$ from $P_1$ on behalf of $P_2$ and $P_3$ respectively.$\SimFourPC$ sets $\flag=1$, if any of the received values are inconsistent.
			\item[--] {\em Obtaining function outputs}:  If $\flag = 1$, $\SimFourPC$ invokes $\FFOURPC$ with input $\bot$ on behalf of $P_1$. Else it sends inputs $\{\val_j\}$ extracted on behalf of $P_1$ to $\FFOURPC$ and receives the function outputs $\vy_1,\ldots, \vy_{\OS}$.
			\item[--] {\em Output Reconstruction}:  For each output $\vy_j$ for $j \in 1,\ldots, \OS$, $\SimFourPC$ computes $\Pad{\vy_j} = \Mask{\vy_j} - \vy_j$, and sends $\Pad{\vy_j}$ and $\Hash(\Pad{\vy_j})$ to $P_1$ on behalf of $P_2$ and $P_0$ respectively. In parallel, it receives $\PadC{\vy_j}$ and $\Hash(\Mask{\vy_j})$ from $P_1$ on behalf of $P_3$ and $P_0$ respectively. $P_0$ is added to $I$, if hash of $\Mask{\vy_j}$ mismatches with the corresponding hash received from $P_1$. Similarly, $P_3$ is added to $I$, if $\PadC{\vy_j}$ mismatches with the corresponding value received from $P_1$. $\SimFourPC$ then sends $I$ to $\FFOURPC$ and terminates.
		\end{myitemize}
	\end{simulatorbox}
\end{IEEEproof}

\subsection{Security Proof in Detail}
This section covers the security proofs for most of our constructions. The proofs for the rest can be easily derived. The proofs use the real-world/ideal-world paradigm in which $\Adv$ is the adversary in the real-world and $\Sim$ is the simulator for the ideal-world, which acts as the honest parties in the protocol and simulates messages received by $\Adv$. The simulator maintains a $\flag$ which is set to $0$ at the start of the protocol. If an honest party $\abort$, the $\flag$ is set to $1$. Simulator for a particular protocol is represented as $\Sim$ with the protocol name as the subscript.

The simulation for a circuit $\ckt$ proceeds as follows: We start with the input sharing phase, and $\Sim$ sets the input of the honest parties to $0$. The simulator can extract the input of the $\Adv$ from the sharing protocol $\piSh$, details of which are provided in the simulation for $\piSh$. Doing so gives the $\Sim$ all the inputs for the entire circuit, which means it can compute all the intermediate values and the output of the circuit. As we will see later, $\Sim$ will use this information to simulate each component of a circuit $\ckt$.

For each of the constructions, we provide simulation proof for the case of corrupt $P_0$ and $P_1$. The cases of corrupt $P_2$ and $P_3$ follows similar to that of $P_1$.

\paragraph{Sharing Protocol}

The ideal functionality realising protocol $\piSh$ is presented in \boxref{fig:FSh}.
\begin{systembox}{$\FSh$}{Functionality for protocol $\piSh$}{fig:FSh}
	\justify
	$\FSh$ interacts with the parties in $\Partyset$ and the adversary $\Sim$. $\FSh$ receives the input $\val$ from party $P_i$ while it receives $\bot$ from the other parties. If $\val = \bot$, then send $\bot$ to every party, else proceed with the computation.
	
	\begin{description}
		\item {\bf Computation of output: } Randomly select $\PadA{\val}, \PadB{\val}, \PadC{\val}$ from $\Z{\ell}$ and set $\Mask{\val} = \val + \PadA{\val} + \PadB{\val} + \PadC{\val}$. The output shares are set as:	
		
		{\small
			\begin{center}
				\begin{tabu} to 1\textwidth { l  l }
					$\shr{\val}_{P_0} = (\PadA{\val}, \PadB{\val}, \PadC{\val})$ & 
					$\shr{\val}_{P_1} = (\Mask{\val}, \PadB{\val}, \PadC{\val})$\\ \\
					$\shr{\val}_{P_2} = (\Mask{\val}, \PadC{\val}, \PadA{\val})$ &
					$\shr{\val}_{P_3} = (\Mask{\val}, \PadA{\val}, \PadB{\val})$\\
				\end{tabu}
			\end{center}
		}
		
		\item {\bf Output to adversary: }  If $\Sim$ sends $\abort$, then send $(\OUTPUT, \bot)$ to all the parties. Otherwise,  send $(\OUTPUT, \shr{\val}_{\Sim})$ to the adversary $\Sim$, where $\shr{\val}_{\Sim}$ denotes the share of $\val$ corresponding to the corrupt party.
		
		\item {\bf Output to selected honest parties: } Receive $(\SELECT, \{I\})$ from adversary $\Sim$, where $\{I\}$ denotes a subset of the honest parties. If an honest parties $P_i$ belongs to $I$, send $(\OUTPUT, \bot)$, else send $(\OUTPUT, \shr{\val}_{i})$,  where $\shr{\val}_{i}$ denotes the share of $\val$ corresponding to the honest party $P_i$.
	\end{description}
\end{systembox}

The simulator for the case of corrupt $P_0$ appears in \boxref{fig:Sim_ShA}.
\begin{simulatorbox}{$\Sim_{\Sh}$}{Simulator $\Sim_{\Sh}$ for the case of corrupt $P_0$}{fig:Sim_ShA}
	\justify \algoHead{Offline Phase:} $\Sim_{\Sh}$ emulates $\FSETUP$ and gives the keys $(k_{01}, k_{02}, k_{03}, k_{012}, k_{013}, k_{023}$ and $\Key{\Partyset})$ to $\Adv$. By emulating $\FSETUP$, it learns the $\Pad{}$-values corresponding to input $\val$. 
	\begin{myitemize}
		\item[--] If $P_i = P_0$, $\Sim_{\Sh}$ computes $\PadV{\val}{j}$ for $j \in {1,2,3}$ on behalf of each $P_j$ using the shared key.
		\item[--] If $P_i = P_k$ for $k \in \{1,2,3\}$, $\Sim_{\Sh}$ computes $\PadV{\val}{k}$ using the key $
		\Key{\Partyset}$. In addition, the $\Pad{\val}$-shares corresponding to the honest parties are computed honestly by $\Sim_{\Sh}$.
	\end{myitemize}
	
	\justify \algoHead{Online Phase:} 
	\begin{myitemize}
		\item[--] If $P_i = P_0$, $\Sim_{\Sh}$ receives $\Mask{\val}$ of behalf of $\ESet$. $\Sim_{\Sh}$ sets $\flag=1$ if the received values mismatch. Else, it computes the input $\val = \Mask{\val} - \PadA{\val} - \PadB{\val} - \PadC{\val}$.
		\item[--] If $P_i \ne P_0$, $\Sim_{\Sh}$ sets $\val = 0$ by assigning $\Mask{\val} = \PadA{\val} + \PadB{\val} + \PadC{\val}$. 
	\end{myitemize}
	%
	\noindent If $\flag = 0$ and $P_i = P_0$, $\Sim_{\Sh}$ invokes $\FSh$ with input $\val$ on behalf of $P_0$. Else it invokes $\FSh$ with input $\bot$ on behalf of $P_0$.	
\end{simulatorbox}

The simulator for the case of corrupt $P_1$ appears in \boxref{fig:Sim_ShB}.
\begin{simulatorbox}{$\Sim_{\Sh}$}{Simulator $\Sim_{\Sh}$ for the case of corrupt $P_1$}{fig:Sim_ShB}
	\justify \algoHead{Offline Phase:} $\Sim_{\Sh}$ emulates $\FSETUP$ and gives the keys $(k_{01}, k_{12}, k_{13}, k_{012}, k_{013}, k_{123}$ and $\Key{\Partyset})$ to $\Adv$. By emulating $\FSETUP$, it learns the $\Pad{}$-values corresponding to input $\val$. 
	\begin{myitemize}
		\item[--] If $P_i = P_1$, $\Sim_{\Sh}$ computes $\PadV{\val}{1}$ using the key $\Key{\Partyset}$.
		\item[--] If $P_i = P_k$ for $k \in \{0,2,3\}$, the $\Pad{\val}$-shares corresponding to the honest parties are computed honestly by $\Sim_{\Sh}$.
	\end{myitemize}
	
	\justify \algoHead{Online Phase:} 
	\begin{myitemize}
		\item[--] If $P_i = P_1$, $\Sim_{\Sh}$ receives $\Mask{\val}$ of behalf of $P_2, P_3$. $\Sim_{\Sh}$ sets $\flag=1$ if the received values mismatch. Else, it computes the input $\val = \Mask{\val} - \PadA{\val} - \PadB{\val} - \PadC{\val}$.
		\item[--] If $P_i \ne P_1$, $\Sim_{\Sh}$ sets $\val = 0$ by assigning $\Mask{\val} = \PadA{\val} + \PadB{\val} + \PadC{\val}$. $\Sim_{\Sh}$ sends $\Mask{\val}$ to $P_1$ on behalf of the sender
	\end{myitemize}
	%
	\noindent If $\flag = 0$ and $P_i = P_1$, $\Sim_{\Sh}$ invokes $\FSh$ with input $\val$ on behalf of $P_1$. Else it invokes $\FSh$ with input $\bot$ on behalf of $P_1$.	
\end{simulatorbox}

\paragraph{Verifiable Arithmetic/Boolean Sharing}

The ideal functionality realising protocol $\pivSh$ is presented in \boxref{fig:FvSh}.
\begin{systembox}{$\FvSh$}{Functionality for protocol $\pivSh$}{fig:FvSh}
	\justify
	$\FvSh$ interacts with the parties in $\Partyset$ and the adversary $\Sim$. $\FSh$ receives the input $\val$ from parties $P_i, P_j$ while it receives $\bot$ from the other parties. If the received values mismatch, then send $\bot$ to every party, else proceed with the computation.
	
	\begin{description}
		\item {\bf Computation of output: } Randomly select $\PadA{\val}, \PadB{\val}, \PadC{\val}$ from $\Z{\ell}$ and set $\Mask{\val} = \val + \PadA{\val} + \PadB{\val} + \PadC{\val}$. The output shares are set as:	
		
		{\small
			\begin{center}
				\begin{tabu} to 1\textwidth { l  l }
					$\shr{\val}_{P_0} = (\PadA{\val}, \PadB{\val}, \PadC{\val})$ & 
					$\shr{\val}_{P_1} = (\Mask{\val}, \PadB{\val}, \PadC{\val})$\\ \\
					$\shr{\val}_{P_2} = (\Mask{\val}, \PadC{\val}, \PadA{\val})$ &
					$\shr{\val}_{P_3} = (\Mask{\val}, \PadA{\val}, \PadB{\val})$\\
				\end{tabu}
			\end{center}
		}
		
		\item {\bf Output to adversary: }  If $\Sim$ sends $\abort$, then send $(\OUTPUT, \bot)$ to all the parties. Otherwise,  send $(\OUTPUT, \shr{\val}_{\Sim})$ to the adversary $\Sim$, where $\shr{\val}_{\Sim}$ denotes the share of $\val$ corresponding to the corrupt party.
		
		\item {\bf Output to selected honest parties: } Receive $(\SELECT, \{I\})$ from adversary $\Sim$, where $\{I\}$ denotes a subset of the honest parties. If an honest parties $P_i$ belongs to $I$, send $(\OUTPUT, \bot)$, else send $(\OUTPUT, \shr{\val}_{i})$,  where $\shr{\val}_{i}$ denotes the share of $\val$ corresponding to the honest party $P_i$.
	\end{description}
\end{systembox}

The simulator for the case of corrupt $P_0$ appears in \boxref{fig:Sim_vShA}.
\begin{simulatorbox}{$\Sim_{\vSh}$}{Simulator $\Sim_{\vSh}$ for the case of corrupt $P_0$}{fig:Sim_vShA}
	\justify \algoHead{Offline Phase:} The simulation for the offline phase is similar to that of $\piSh(P_i, \val)$ for the case of corrupt $P_0$ (\boxref{fig:Sim_ShA}).
	
	\justify \algoHead{Online Phase:} $\Sim_{\vSh}$ proceeds similar to that of $\piSh(P_i, \val)$ for the case of corrupt $P_0$ (\boxref{fig:Sim_ShA}). In addition, if $P_j = P_0$, it receives $\Hash(\Mask{\val})$ from $\Adv$ on behalf of $P_1, P_2$, and $P_3$.  $\Sim_{\vSh}$ sets $\flag=1$ if the received values mismatch.
	
	\noindent If $\flag = 0$ and $P_j = P_0$, $\Sim_{\vSh}$ invokes $\FvSh$ with input $\val$ on behalf of $P_0$. Else it invokes $\FvSh$ with input $\bot$ on behalf of $P_0$.	
\end{simulatorbox}

The simulator for the case of corrupt $P_1$ appears in \boxref{fig:Sim_vShB}.
\begin{simulatorbox}{$\Sim_{\vSh}$}{Simulator $\Sim_{\vSh}$ for the case of corrupt $P_1$}{fig:Sim_vShB}
	\justify \algoHead{Offline Phase:} The simulation for the offline phase is similar to that of $\piSh(P_i, \val)$ for the case of corrupt $P_1$ (\boxref{fig:Sim_ShB}).
	
	\justify \algoHead{Online Phase:} $\Sim_{\vSh}$ proceeds similar to that of $\piSh(P_i, \val)$ for the case of corrupt $P_1$ (\boxref{fig:Sim_ShB}). In addition, if $P_j = P_1$, it receives $\Hash(\Mask{\val})$ from $\Adv$ on behalf of $P_2$, and $P_3$.  $\Sim_{\vSh}$ sets $\flag=1$ if the received values mismatch.
	
	\noindent If $\flag = 0$ and $P_i = P_1$, $\Sim_{\vSh}$ invokes $\FvSh$ with input $\val$ on behalf of $P_1$. Else it invokes $\FvSh$ with input $\bot$ on behalf of $P_1$.	
\end{simulatorbox}

\paragraph{Reconstruction Protocol}

The ideal functionality realising protocol $\piRec$ is presented in \boxref{fig:FRec}.
\begin{systembox}{$\FRec$}{Functionality for protocol $\piRec$}{fig:FRec}
	\justify
	$\FRec$ interacts with the parties in $\Partyset$ and the adversary $\Sim$. $\FRec$ receives the $\shr{\cdot}$-shares of value $\val$ from party $P_i$ for $i \in \{0,1,2,3\}$. The shares are
	
	{\small
		\begin{center}
			\begin{tabu} to 1\textwidth { l  l }
				$\shr{\val}_{P_0} = (\PadA{\val}', \PadB{\val}', \PadC{\val}')$ & 
				$\shr{\val}_{P_1} = (\Mask{\val}', \PadB{\val}'', \PadC{\val}'')$\\ \\
				$\shr{\val}_{P_2} = (\Mask{\val}'', \PadC{\val}''', \PadA{\val}'')$ &
				$\shr{\val}_{P_3} = (\Mask{\val}''', \PadA{\val}''', \PadB{\val}''')$\\
			\end{tabu}
		\end{center}
	}
	
	$\FRec$ sends $\bot$ to every party if either of the following condition is met: i) $\PadA{\val}' \ne \PadA{\val}'' \ne \PadA{\val}'''$, ii) $\PadB{\val}' \ne \PadB{\val}'' \ne \PadB{\val}'''$, iii) $\PadC{\val}' \ne \PadC{\val}'' \ne \PadC{\val}'''$ or iv) $\Mask{\val}' \ne \Mask{\val}'' \ne \Mask{\val}'''$. Else it proceeds with the computation.
	
	\begin{description}
		\item {\bf Computation of output: } Set $\val = \Mask{\val} - \PadA{\val} - \PadB{\val} - \PadC{\val}$.
		
		\item {\bf Output to adversary: }  If $\Sim$ sends $\abort$, then send $(\OUTPUT, \bot)$ to all the parties. Otherwise,  send $(\OUTPUT, \val)$ to the adversary $\Sim$.
		
		\item {\bf Output to selected honest parties: } Receive $(\SELECT, \{I\})$ from adversary $\Sim$, where $\{I\}$ denotes a subset of the honest parties. If an honest party $P_i$ belongs to $I$, send $(\OUTPUT, \bot)$, else send $(\OUTPUT, \val)$.
	\end{description}
\end{systembox}

The simulator for the case of corrupt $P_0$ appears in \boxref{fig:Sim_RecA}. As mentioned at the beginning of this section, $\Sim$ knows all of the intermediate values and the output of the $\ckt$. $\Sim$ uses this information to simulate the $\piRec$ protocol.
\begin{simulatorbox}{$\Sim_{\Rec}$}{Simulator $\Sim_{\Rec}$ for the case of corrupt $P_0$}{fig:Sim_RecA}
	\justify \algoHead{Online Phase:} 
	\begin{myitemize}
		\item[--] $\Sim_{\Rec}$ computes $\Mask{\val} = \val + \PadA{\val} + \PadB{\val} + \PadC{\val}$. It then sends $\Mask{\val}$ and $\Hash(\Mask{\val})$ to $\Adv$ on behalf of $P_1$ and $P_2$ respectively.
		\item[--] $\Sim_{\Rec}$ receives $\Hash(\PadA{\val}'), \Hash(\PadB{\val}')$ and $\Hash(\PadA{\val}')$ from $\Adv$ on behalf of $P_1, P_2$ and $P_3$ respectively. $\Sim_{\Rec}$ sets $\flag = 1$ if any of the received hash values is inconsistent.
	\end{myitemize}
	%
	\noindent If $\flag = 0$, $\Sim_{\Rec}$ invokes $\FRec$ with input $(\PadA{\val}, \PadB{\val}, \PadC{\val})$ on behalf of $P_0$. Else it invokes $\FRec$ with input $\bot$ on behalf of $P_0$.	
\end{simulatorbox}

The simulator for the case of corrupt $P_1$ appears in \boxref{fig:Sim_RecB}.
\begin{simulatorbox}{$\Sim_{\Rec}$}{Simulator $\Sim_{\Rec}$ for the case of corrupt $P_1$}{fig:Sim_RecB}
	\justify \algoHead{Online Phase:} 
	\begin{myitemize}
		\item[--] $\Sim_{\Rec}$ sends $\PadA{\val}$ and $\Hash(\PadA{\val})$ to $\Adv$ on behalf of $P_2$ and $P_0$ respectively.
		\item[--] $\Sim_{\Rec}$ receives $\PadC{\val}'$ and $\Hash(\Mask{\val}')$ from $\Adv$ on behalf of $P_3$ and $P_0$ respectively. $\Sim_{\Rec}$ sets $\flag = 1$ if any of the received values is inconsistent.
	\end{myitemize}
	%
\noindent If $\flag = 0$, $\Sim_{\Rec}$ invokes $\FRec$ with input $(\Mask{\val}, \PadB{\val}, \PadC{\val})$ on behalf of $P_1$. Else it invokes $\FRec$ with input $\bot$ on behalf of $P_1$.	
\end{simulatorbox}

\paragraph{Multiplication}

The ideal functionality realising protocol $\piMult$ is presented in \boxref{fig:FMult}.
\begin{systembox}{$\FMult$}{Functionality for protocol $\piMult$}{fig:FMult}
	\justify
	$\FMult$ interacts with the parties in $\Partyset$ and the adversary $\Sim$. $\FMult$ receives $\shr{\cdot}$-shares of values $\wx$ and $\wy$ from the parties as input. If $\FMult$ receives $\bot$ from $\Sim$, then send $\bot$ to every party, else proceed with the computation.
	
	\begin{description}
		\item {\bf Computation of output: } Compute $\wx = \Mask{\wx} - \PadA{\wx} - \PadB{\wx} - \PadC{\wx}, \wy = \Mask{\wy} - \PadA{\wy} - \PadB{\wy} - \PadC{\wy}$ and set $\wz = \wx \wy$. Randomly select $\PadA{\wz}, \PadB{\wz}, \PadC{\wz}$ from $\Z{\ell}$ and set $\Mask{\wz} = \wz + \PadA{\wz} + \PadB{\wz} + \PadC{\wz}$. The output shares are set as:	
		
		{\small
			\begin{center}
				\begin{tabu} to 1\textwidth { l  l }
					$\shr{\wz}_{P_0} = (\PadA{\wz}, \PadB{\wz}, \PadC{\wz})$ & 
					$\shr{\wz}_{P_1} = (\Mask{\wz}, \PadB{\wz}, \PadC{\wz})$\\ \\
					$\shr{\wz}_{P_2} = (\Mask{\wz}, \PadC{\wz}, \PadA{\wz})$ &
					$\shr{\wz}_{P_3} = (\Mask{\wz}, \PadA{\wz}, \PadB{\wz})$\\
				\end{tabu}
			\end{center}
		}
		
		\item {\bf Output to adversary: }  If $\Sim$ sends $\abort$, then send $(\OUTPUT, \bot)$ to all the parties. Otherwise,  send $(\OUTPUT, \shr{\wz}_{\Sim})$ to the adversary $\Sim$, where $\shr{\wz}_{\Sim}$ denotes the share of $\wz$ corresponding to the corrupt party.
		
		\item {\bf Output to selected honest parties: } Receive $(\SELECT, \{I\})$ from adversary $\Sim$, where $\{I\}$ denotes a subset of the honest parties. If an honest party $P_i$ belongs to $I$, send $(\OUTPUT, \bot)$, else send $(\OUTPUT, \shr{\wz}_{i})$,  where $\shr{\wz}_{i}$ denotes the share of $\wz$ corresponding to the honest party $P_i$.
	\end{description}
\end{systembox}

The simulator for the case of corrupt $P_0$ appears in \boxref{fig:Sim_MultI}.
\begin{simulatorbox}{$\Sim_{\Mult}$}{Simulator for the case of corrupt $P_0$}{fig:Sim_MultI}
	\justify \algoHead{Offline Phase:}
	\begin{myitemize}
		\item[--] $\Sim_{\Mult}$ samples $\PadV{\wz}{j}$ for $j \in \EInSet$ honestly using the shared keys obtained.
		\item[--] $\Sim_{\Mult}$ emulates $\FZERO$ functionality and gives the values $A, B$, and $\Gamma$ to $\Adv$.
		\item[--] $\Sim_{\Mult}$ receives $\Hash(\GammaxyA), \Hash(\GammaxyB)$ and $\Hash(\GammaxyC)$ from $\Adv$ on behalf of $P_2, P_3$ and $P_1$ respectively.  $\Sim_{\Mult}$ sets $\flag = 1$ if any of the received values is inconsistent.
	\end{myitemize}
	
	\justify \algoHead{Online Phase:} There is nothing to simulate as $P_0$ has no role during the online phase.
	
	\noindent If $\flag = 0$, $\Sim_{\Mult}$ invokes $\FMult$ with input $(\shr{\wx}_{P_0}, \shr{\wy}_{P_0})$ on behalf of $P_0$. Else it invokes $\FMult$ with input $\bot$ on behalf of $P_0$.
\end{simulatorbox}

The simulator for the case of corrupt $P_1$ appears in \boxref{fig:Sim_MultII}.
\begin{simulatorbox}{$\Sim_{\Mult}$}{Simulator for the case of corrupt $P_1$}{fig:Sim_MultII}
	\justify \algoHead{Offline Phase:}
	\begin{myitemize}
		\item[--] $\Sim_{\Mult}$ samples $\PadV{\wz}{j}$ for $j \in \{2,3\}$ honestly using the shared keys obtained.
		\item[--] $\Sim_{\Mult}$ emulates $\FZERO$ functionality and gives the value $A$ to $\Adv$.
		\item[--] $\Sim_{\Mult}$ sends $\GammaxyC$ and $\Hash(\GammaxyC)$ to $\Adv$ on behalf of $P_2$ and $P_0$ respectively. It receives $\GammaxyB$ from $\Adv$ on behalf of $P_3$ and sets $\flag = 1$ if the received value is inconsistent.
	\end{myitemize}
	
	\justify \algoHead{Online Phase:}
	\begin{myitemize}
		\item[--] $\Sim_{\Mult}$ simulates the computation of $\Mask{\wz}'$ shares honestly on behalf of $P_2$ and $P_3$.
		\item[--] $\Sim_{\Mult}$ sends $\Mask{\wz,1}'$ and $\Hash(\Mask{\wz,1}')$ to $\Adv$ on behalf of $P_2$ and $P_3$ respectively.  It receives $\Mask{\wz,3}'$ and $\Hash(\Mask{\wz,2}')$ from $\Adv$ on behalf of $P_3$ and $P_2$ respectively. $\Sim_{\Mult}$ sets $\flag = 1$ if any of the received values is inconsistent.
	\end{myitemize}
	
	\noindent If $\flag = 0$, $\Sim_{\Mult}$ invokes $\FMult$ with input $(\shr{\wx}_{P_1}, \shr{\wy}_{P_1})$ on behalf of $P_1$. Else it invokes $\FMult$ with input $\bot$ on behalf of $P_1$.
\end{simulatorbox}

\paragraph{Dot Product}

The ideal functionality realising protocol $\piDotP$ is presented in \boxref{fig:FDotP}.
\begin{systembox}{$\FDotP$}{Functionality for protocol $\piSh$}{fig:FDotP}
	\justify
	$\FDotP$ interacts with the parties in $\Partyset$ and the adversary $\Sim$. $\FDotP$ receives $\shr{\cdot}$-shares of vectors $\vecX$ and $\vecY$ from the parties as input. Here $\vecX$ and $\vecY$ are $d$-length vectors.	If $\FDotP$ receives $\bot$ from $\Sim$, then send $\bot$ to every party, else proceed with the computation.
	
	\begin{description}
		\item {\bf Computation of output: } Compute $\wx_i = \Mask{\wx_i} - \PadA{\wx_i} - \PadB{\wx_i} - \PadC{\wx_i}, \wy_i = \Mask{\wy_i} - \PadA{\wy_i} - \PadB{\wy_i} - \PadC{\wy_i}$ for $i \in [d]$ and set $\wz = \sum_{i = 1}^{d} \wx_i \wy_i$. Randomly select $\PadA{\wz}, \PadB{\wz}, \PadC{\wz}$ from $\Z{\ell}$ and set $\Mask{\wz} = \wz + \PadA{\wz} + \PadB{\wz} + \PadC{\wz}$. The output shares are set as:	
		
		{\small
			\begin{center}
				\begin{tabu} to 1\textwidth { l  l }
					$\shr{\wz}_{P_0} = (\PadA{\wz}, \PadB{\wz}, \PadC{\wz})$ & 
					$\shr{\wz}_{P_1} = (\Mask{\wz}, \PadB{\wz}, \PadC{\wz})$\\ \\
					$\shr{\wz}_{P_2} = (\Mask{\wz}, \PadC{\wz}, \PadA{\wz})$ &
					$\shr{\wz}_{P_3} = (\Mask{\wz}, \PadA{\wz}, \PadB{\wz})$\\
				\end{tabu}
			\end{center}
		}
		
		\item {\bf Output to adversary: }  If $\Sim$ sends $\abort$, then send $(\OUTPUT, \bot)$ to all the parties. Otherwise,  send $(\OUTPUT, \shr{\wz}_{\Sim})$ to the adversary $\Sim$, where $\shr{\wz}_{\Sim}$ denotes the share of $\wz$ corresponding to the corrupt party.
		
		\item {\bf Output to selected honest parties: } Receive $(\SELECT, \{I\})$ from adversary $\Sim$, where $\{I\}$ denotes a subset of the honest parties. If an honest party $P_i$ belongs to $I$, send $(\OUTPUT, \bot)$, else send $(\OUTPUT, \shr{\wz}_{i})$,  where $\shr{\wz}_{i}$ denotes the share of $\wz$ corresponding to the honest party $P_i$.
	\end{description}
\end{systembox}

The simulator for the case of corrupt $P_0$ appears in \boxref{fig:Sim_DotPI}.
\begin{simulatorbox}{$\Sim_{\DotP}$}{Simulator for the case of corrupt $P_0$}{fig:Sim_DotPI}
	\justify \algoHead{Offline Phase:}
	\begin{myitemize}
		\item[--] $\Sim_{\DotP}$ samples $\PadV{\wz}{j}$ for $j \in \EInSet$ honestly using the shared keys obtained.
		\item[--] $\Sim_{\DotP}$ emulates $\FZERO$ functionality and gives the values $A, B$, and $\Gamma$ to $\Adv$.
		\item[--] $\Sim_{\DotP}$ receives $\Hash(\GammaxyA), \Hash(\GammaxyB)$ and $\Hash(\GammaxyC)$ from $\Adv$ on behalf of $P_2, P_3$ and $P_1$ respectively.  $\Sim_{\DotP}$ sets $\flag = 1$ if any of the received values is inconsistent.
	\end{myitemize}
	
	\justify \algoHead{Online Phase:} There is nothing to simulate as $P_0$ has no role during the online phase.
	
	\noindent If $\flag = 0$, $\Sim_{\DotP}$ invokes $\FDotP$ with input $(\{\shr{\wx_i}_{P_0}, \shr{\wy_i}_{P_0}\}_{i \in [d]})$ on behalf of $P_0$. Else it invokes $\FDotP$ with input $\bot$ on behalf of $P_0$.
\end{simulatorbox}

The simulator for the case of corrupt $P_1$ appears in \boxref{fig:Sim_DotPII}.
\begin{simulatorbox}{$\Sim_{\DotP}$}{Simulator for the case of corrupt $P_1$}{fig:Sim_DotPII}
	\justify \algoHead{Offline Phase:}
	\begin{myitemize}
		\item[--] $\Sim_{\DotP}$ samples $\PadV{\wz}{j}$ for $j \in \{2,3\}$ honestly using the shared keys obtained.
		\item[--] $\Sim_{\DotP}$ emulates $\FZERO$ functionality and gives the value $A$ to $\Adv$.
		\item[--] $\Sim_{\DotP}$ sends $\GammaxyC$ and $\Hash(\GammaxyC)$ to $\Adv$ on behalf of $P_2$ and $P_0$ respectively. It receives $\GammaxyB$ from $\Adv$ on behalf of $P_3$ and sets $\flag = 1$ if the received value is inconsistent.
	\end{myitemize}
	
	\justify \algoHead{Online Phase:}
	\begin{myitemize}
		\item[--] $\Sim_{\DotP}$ simulates the computation of $\Mask{\wz}'$ shares honestly on behalf of $P_2$ and $P_3$.
		\item[--] $\Sim_{\DotP}$ sends $\Mask{\wz,1}'$ and $\Hash(\Mask{\wz,1}')$ to $\Adv$ on behalf of $P_2$ and $P_3$ respectively.  It receives $\Mask{\wz,3}'$ and $\Hash(\Mask{\wz,2}')$ from $\Adv$ on behalf of $P_3$ and $P_2$ respectively. $\Sim_{\DotP}$ sets $\flag = 1$ if any of the received values is inconsistent.
	\end{myitemize}
	
	\noindent If $\flag = 0$, $\Sim_{\DotP}$ invokes $\FDotP$ with input $(\{\shr{\wx_i}_{P_1}, \shr{\wy_i}_{P_1}\}_{i \in [d]})$ on behalf of $P_1$. Else it invokes $\FDotP$ with input $\bot$ on behalf of $P_1$.
\end{simulatorbox}

\paragraph{Bit to Arithmetic Sharing ($\BitA$)}

The ideal functionality realising protocol $\PiBitA$ is presented in \boxref{fig:FBitA}.
\begin{systembox}{$\FBitA$}{Functionality for protocol $\PiBitA$}{fig:FBitA}
	\justify
	$\FBitA$ interacts with the parties in $\Partyset$ and the adversary $\Sim$. $\FBitA$ receives $\shareB{\cdot}$-share of a bit $\bitb$ from the parties as input. If $\FBitA$ receives $\bot$ from $\Sim$, then send $\bot$ to every party, else proceed with the computation.
	
	\begin{description}
		\item {\bf Computation of output: } Compute $\bitb = \Mask{\bitb} \xor \PadA{\bitb} \xor \PadB{\bitb} \xor \PadC{\bitb}$. Let $\bitb'$ denotes the value of bit $\bitb$ over an arithmetic ring $\Z{\ell}$. Randomly select $\PadA{\bitb}', \PadB{\bitb}', \PadC{\bitb}'$ from $\Z{\ell}$ and set $\Mask{\bitb}' = \bitb' + \PadA{\bitb}' + \PadB{\bitb}' + \PadC{\bitb}'$. The output shares are set as:
		
		{\small
			\begin{center}
				\begin{tabu} to 1\textwidth { l  l }
					$\shr{\bitb}'_{P_0} = (\PadA{\bitb}', \PadB{\bitb}', \PadC{\bitb}')$ & 
					$\shr{\bitb}'_{P_1} = (\Mask{\bitb}', \PadB{\bitb}', \PadC{\bitb}')$\\ \\
					$\shr{\bitb}'_{P_2} = (\Mask{\bitb}', \PadC{\bitb}', \PadA{\bitb}')$ &
					$\shr{\bitb}'_{P_3} = (\Mask{\bitb}', \PadA{\bitb}', \PadB{\bitb}')$\\
				\end{tabu}
			\end{center}
		}
		
		\item {\bf Output to adversary: }  If $\Sim$ sends $\abort$, then send $(\OUTPUT, \bot)$ to all the parties. Otherwise,  send $(\OUTPUT, \shr{\bitb'}_{\Sim})$ to the adversary $\Sim$, where $\shr{\bitb'}_{\Sim}$ denotes the share of $\bitb'$ corresponding to the corrupt party.
		
		\item {\bf Output to selected honest parties: } Receive $(\SELECT, \{I\})$ from adversary $\Sim$, where $\{I\}$ denotes a subset of the honest parties. If an honest party $P_i$ belongs to $I$, send $(\OUTPUT, \bot)$, else send $(\OUTPUT, \shr{\bitb'}_{i})$,  where $\shr{\bitb'}_{i}$ denotes the share of $\bitb'$ corresponding to the honest party $P_i$.
	\end{description}
\end{systembox}

The simulator for the case of corrupt $P_0$ appears in \boxref{fig:Sim_BitAI}.
\begin{simulatorbox}{$\Sim_{\BitA}$}{Simulator for the case of corrupt $P_0$}{fig:Sim_BitAI}
	\justify \algoHead{Offline Phase:}
	\begin{myitemize}
		\item[--] Corresponding to the invocation of $\piaSh(P_0, \vu)$ protocol, $\Sim_{\BitA}$ receives $\val_3$ from $\Adv$ on behalf of both $P_1$ and $P_2$.  $\Sim_{\BitA}$ sets $\flag = 1$ if the received values mismatch.
		\item[--] $\Sim_{\BitA}$ performs the check honestly and sets $\flag = 1$ if the verification fails.
	\end{myitemize}
	
	\justify \algoHead{Online Phase:} The steps corresponding to $\pivSh$ and $\piMult$ are simulated similar to $\Sim_{\vSh}$ (\boxref{fig:Sim_vShA}) and $\Sim_{\Mult}$ (\boxref{fig:Sim_MultI}) respectively, for the case of corrupt $P_0$.
	
	\noindent If $\flag = 0$, $\Sim_{\BitA}$ invokes $\FBitA$ with input $(\shareB{\bitb}_{P_0})$ on behalf of $P_0$. Else it invokes $\FBitA$ with input $\bot$ on behalf of $P_0$.
\end{simulatorbox}

The simulator for the case of corrupt $P_1$ appears in \boxref{fig:Sim_BitAII}.
\begin{simulatorbox}{$\Sim_{\BitA}$}{Simulator for the case of corrupt $P_1$}{fig:Sim_BitAII}
	\justify \algoHead{Offline Phase:}
	\begin{myitemize}
		\item[--] Corresponding to the invocation of $\piaSh(P_0, \vu)$ protocol, $\Sim_{\BitA}$ sends $\val_3$ to $\Adv$ on behalf of $P_0$. It then sends $\Hash(\val_3)$ to $\Adv$ on behalf of $P_2$ and receives $\Hash(\val_3')$ back. $\Sim_{\BitA}$ sets $\flag = 1$ if it receives inconsistent hash value.
		\item[--] $\Sim_{\BitA}$ receives $(\wx_1, \wy_1)$ from $\Adv$ on behalf of $P_3$, performs the check honestly and sets $\flag = 1$ if the verification fails.
	\end{myitemize}
	
	\justify \algoHead{Online Phase:} The steps corresponding to $\pivSh$ and $\piMult$ are simulated similar to $\Sim_{\vSh}$ (\boxref{fig:Sim_vShB}) and $\Sim_{\Mult}$ (\boxref{fig:Sim_MultII}) respectively, for the case of corrupt $P_1$.
	
	\noindent If $\flag = 0$, $\Sim_{\BitA}$ invokes $\FBitA$ with input $(\shareB{\bitb}_{P_1})$ on behalf of $P_1$. Else it invokes $\FBitA$ with input $\bot$ on behalf of $P_1$.
\end{simulatorbox}

\paragraph{Bit Injection ($\BitInj$)}

The ideal functionality realising protocol $\PiBitInj$ is presented in \boxref{fig:FBitInj}.
\begin{systembox}{$\FBitInj$}{Functionality for protocol $\PiBitInj$}{fig:FBitInj}
	\justify
	$\FBitInj$ interacts with the parties in $\Partyset$ and the adversary $\Sim$. $\FBitInj$ receives $(\shareB{\bitb}, \shr{\val})$ from the parties as input. If $\FBitInj$ receives $\bot$ from $\Sim$, then send $\bot$ to every party, else proceed with the computation.
	
	\begin{description}
		\item {\bf Computation of output: } Compute $\bitb = \Mask{\bitb} \xor \PadA{\bitb} \xor \PadB{\bitb} \xor \PadC{\bitb}, \val = \Mask{\val} - \PadA{\val} - \PadB{\val} - \PadC{\val}$ and set $(\bitb \val) = \bitb \cdot \val$. Randomly select $\PadA{(\bitb \val)}, \PadB{(\bitb \val)}, \PadC{(\bitb \val)}$ from $\Z{\ell}$ and set $\Mask{(\bitb \val)} = {(\bitb \val)} + \PadA{(\bitb \val)} + \PadB{(\bitb \val)} + \PadC{(\bitb \val)}$. The output shares are set as:
		
				\begin{align*}
					\shr{(\bitb \val)}_{P_0} &= (\PadA{(\bitb \val)}, \PadB{(\bitb \val)}, \PadC{(\bitb \val)}) \\ 
					\shr{(\bitb \val)}_{P_1} &= (\Mask{(\bitb \val)}, \PadB{(\bitb \val)}, \PadC{(\bitb \val)}) \\ 
					\shr{(\bitb \val)}_{P_2} &= (\Mask{(\bitb \val)}, \PadC{(\bitb \val)}, \PadA{(\bitb \val)}) \\
					\shr{(\bitb \val)}_{P_3} &= (\Mask{(\bitb \val)}, \PadA{(\bitb \val)}, \PadB{(\bitb \val)}) 
				\end{align*}
		
		\item {\bf Output to adversary: }  If $\Sim$ sends $\abort$, then send $(\OUTPUT, \bot)$ to all the parties. Otherwise,  send $(\OUTPUT, \shr{(\bitb \val)}_{\Sim})$ to the adversary $\Sim$, where $\shr{(\bitb \val)}_{\Sim}$ denotes the share of $(\bitb \val)$ corresponding to the corrupt party.
		
		\item {\bf Output to selected honest parties: } Receive $(\SELECT, \{I\})$ from adversary $\Sim$, where $\{I\}$ denotes a subset of the honest parties. If an honest party $P_i$ belongs to $I$, send $(\OUTPUT, \bot)$, else send $(\OUTPUT, \shr{(\bitb \val)}_{i})$,  where $\shr{(\bitb \val)}_{i}$ denotes the share of $(\bitb \val)$ corresponding to the honest party $P_i$.
	\end{description}
\end{systembox}

The simulator for the case of corrupt $P_0$ appears in \boxref{fig:Sim_BitInjI}.
\begin{simulatorbox}{$\Sim_{\BitInj}$}{Simulator for the case of corrupt $P_0$}{fig:Sim_BitInjI}
	\justify \algoHead{Offline Phase:}
	\begin{myitemize}
		\item[--] Corresponding to the invocation of $\piaSh(P_0, \vy_j)$ protocol, $\Sim_{\BitInj}$ receives $\wy_{j,3}$ from $\Adv$ on behalf of both $P_1$ and $P_2$.  $\Sim_{\BitInj}$ sets $\flag = 1$ if the received values mismatch.
		\item[--] $\Sim_{\BitInj}$ performs the check honestly and sets $\flag = 1$ if the verification fails.
	\end{myitemize}
	
	\justify \algoHead{Online Phase:} The steps corresponding to $\pivSh$ are simulated similar to $\Sim_{\vSh}$ (\boxref{fig:Sim_vShA}), for the case of corrupt $P_0$.
	
	\noindent If $\flag = 0$, $\Sim_{\BitInj}$ invokes $\FBitInj$ with input $(\shareB{\bitb}_{P_0}, \shr{\val}_{P_0})$ on behalf of $P_0$. Else it invokes $\FBitInj$ with input $\bot$ on behalf of $P_0$.
\end{simulatorbox}

The simulator for the case of corrupt $P_1$ appears in \boxref{fig:Sim_BitInjII}.
\begin{simulatorbox}{$\Sim_{\BitInj}$}{Simulator for the case of corrupt $P_1$}{fig:Sim_BitInjII}
	\justify \algoHead{Offline Phase:}
	\begin{myitemize}
		\item[--] Corresponding to the invocation of $\piaSh(P_0, \vu)$ protocol, $\Sim_{\BitInj}$ sends $\wy_{j,3}$ to $\Adv$ on behalf of $P_0$. It then sends $\Hash(\wy_{j,3})$ to $\Adv$ on behalf of $P_2$ and receives $\Hash(\wy_{j,3}')$ back. $\Sim_{\BitInj}$ sets $\flag = 1$ if it receives inconsistent hash value.
		\item[--] $\Sim_{\BitInj}$ emulates $\FZERO$ functionality and gives $A$ to $\Adv$.
		\item[--] $\Sim_{\BitInj}$ receives $\wz_2$ from $\Adv$ on behalf of $P_3$, performs the check honestly and sets $\flag = 1$ if the verification fails.
	\end{myitemize}
	
	\justify \algoHead{Online Phase:} The steps corresponding to $\pivSh$ are simulated similar to $\Sim_{\vSh}$ (\boxref{fig:Sim_vShB}), for the case of corrupt $P_1$.
	
	\noindent If $\flag = 0$, $\Sim_{\BitInj}$ invokes $\FBitInj$ with input $(\shareB{\bitb}_{P_1}, \shr{\val}_{P_1})$ on behalf of $P_1$. Else it invokes $\FBitInj$ with input $\bot$ on behalf of $P_1$.
\end{simulatorbox}

\paragraph{Multiplication with Truncation}

The ideal functionality realising protocol $\piMultTr$ is presented in \boxref{fig:FMultTr}.
\begin{systembox}{$\FMultTr$}{Functionality for protocol $\piMultTr$}{fig:FMultTr}
	\justify
	$\FMultTr$ interacts with the parties in $\Partyset$ and the adversary $\Sim$. $\FMultTr$ receives $\shr{\cdot}$-shares of values $\wx$ and $\wy$ from the parties as input. If $\FMultTr$ receives $\bot$ from $\Sim$, then send $\bot$ to every party, else proceed with the computation.
	
	\begin{description}
		\item {\bf Computation of output: } Compute $\wx = \Mask{\wx} - \PadA{\wx} - \PadB{\wx} - \PadC{\wx}, \wy = \Mask{\wy} - \PadA{\wy} - \PadB{\wy} - \PadC{\wy}$ and set $\wz^{\vt} = (\wx \wy)^{\vt}$. Randomly select $\PadA{\wz^{\vt}}, \PadB{\wz^{\vt}}, \PadC{\wz^{\vt}}$ from $\Z{\ell}$ and set $\Mask{\wz^{\vt}} = \wz^{\vt} + \PadA{\wz^{\vt}} + \PadB{\wz^{\vt}} + \PadC{\wz^{\vt}}$. The output shares are set as:	
		
		{\small
			\begin{center}
				\begin{tabu} to 1\textwidth { l  l }
					$\shr{\wz^{\vt}}_{P_0} = (\PadA{\wz^{\vt}}, \PadB{\wz^{\vt}}, \PadC{\wz^{\vt}})$ & 
					$\shr{\wz^{\vt}}_{P_1} = (\Mask{\wz^{\vt}}, \PadB{\wz^{\vt}}, \PadC{\wz^{\vt}})$\\ \\
					$\shr{\wz^{\vt}}_{P_2} = (\Mask{\wz^{\vt}}, \PadC{\wz^{\vt}}, \PadA{\wz^{\vt}})$ &
					$\shr{\wz^{\vt}}_{P_3} = (\Mask{\wz^{\vt}}, \PadA{\wz^{\vt}}, \PadB{\wz^{\vt}})$\\
				\end{tabu}
			\end{center}
		}
		
		\item {\bf Output to adversary: }  If $\Sim$ sends $\abort$, then send $(\OUTPUT, \bot)$ to all the parties. Otherwise,  send $(\OUTPUT, \shr{\wz^{\vt}}_{\Sim})$ to the adversary $\Sim$, where $\shr{\wz^{\vt}}_{\Sim}$ denotes the share of $\wz^{\vt}$ corresponding to the corrupt party.
		
		\item {\bf Output to selected honest parties: } Receive $(\SELECT, \{I\})$ from adversary $\Sim$, where $\{I\}$ denotes a subset of the honest parties. If an honest party $P_i$ belongs to $I$, send $(\OUTPUT, \bot)$, else send $(\OUTPUT, \shr{\wz^{\vt}}_{i})$,  where $\shr{\wz^{\vt}}_{i}$ denotes the share of $\wz^{\vt}$ corresponding to the honest party $P_i$.
	\end{description}
\end{systembox}

The simulator for the case of corrupt $P_0$ appears in \boxref{fig:Sim_MultTrI}.
\begin{simulatorbox}{$\Sim_{\MultTr}$}{Simulator for the case of corrupt $P_0$}{fig:Sim_MultTrI}
	\justify \algoHead{Offline Phase:}
	\begin{myitemize}
		\item[--] The steps corresponding to offline phase of $\piMult$ are simulated similar to offline phase of $\Sim_{\Mult}$ (\boxref{fig:Sim_MultI}), for the case of corrupt $P_0$.
		\item[--] Corresponding to the invocation of $\piaSh(P_0, \vrt)$ protocol, $\Sim_{\MultTr}$ receives $\vrt_3$ from $\Adv$ on behalf of both $P_1$ and $P_2$.  $\Sim_{\MultTr}$ sets $\flag = 1$ if the received values mismatch.
		\item[--] $\Sim_{\MultTr}$ performs the check honestly and sets $\flag = 1$ if the verification fails.
	\end{myitemize}
	
	\justify \algoHead{Online Phase:} There is nothing to simulate as $P_0$ has no role during the online phase.
	
	\noindent If $\flag = 0$, $\Sim_{\MultTr}$ invokes $\FMultTr$ with input $(\shr{\wx}_{P_0}, \shr{\wy}_{P_0})$ on behalf of $P_0$. Else it invokes $\FMultTr$ with input $\bot$ on behalf of $P_0$.
\end{simulatorbox}

The simulator for the case of corrupt $P_1$ appears in \boxref{fig:Sim_MultTrII}.
\begin{simulatorbox}{$\Sim_{\MultTr}$}{Simulator for the case of corrupt $P_1$}{fig:Sim_MultTrII}
	\justify \algoHead{Offline Phase:}
	\begin{myitemize}
		\item[--] The steps corresponding to offline phase of $\piMult$ are simulated similar to offline phase of $\Sim_{\Mult}$ (\boxref{fig:Sim_MultII}), for the case of corrupt $P_1$.
		\item[--] Corresponding to the invocation of $\piaSh(P_0, \vrt)$ protocol, $\Sim_{\MultTr}$ sends $\vrt_3$ to $\Adv$ on behalf of $P_0$. It then sends $\Hash(\vrt_3)$ to $\Adv$ on behalf of $P_2$ and receives $\Hash({\vrt_3}')$ back. $\Sim_{\MultTr}$ sets $\flag = 1$ if it receives inconsistent hash value.
		\item[--] $\Sim_{\MultTr}$ receives  $(\vm_1, \Hash(\vc))$ from $\Adv$ on behalf of $P_2$, performs the check honestly and sets $\flag = 1$ if the verification fails.
	\end{myitemize}
	
	\justify \algoHead{Online Phase:}
	\begin{myitemize}
		\item[--] $\Sim_{\MultTr}$ simulates the computation of $\Mask{\wz}'$ shares honestly on behalf of $P_2$ and $P_3$.
		\item[--] $\Sim_{\MultTr}$ sends $\Mask{\wz,1}'$ and $\Hash(\Mask{\wz,1}')$ to $\Adv$ on behalf of $P_2$ and $P_3$ respectively.  It receives $\Mask{\wz,3}'$ and $\Hash(\Mask{\wz,2}')$ from $\Adv$ on behalf of $P_3$ and $P_2$ respectively. $\Sim_{\MultTr}$ sets $\flag = 1$ if any of the received values is inconsistent.
	\end{myitemize}
	
	\noindent If $\flag = 0$, $\Sim_{\MultTr}$ invokes $\FMultTr$ with input $(\shr{\wx}_{P_1}, \shr{\wy}_{P_1})$ on behalf of $P_1$. Else it invokes $\FMultTr$ with input $\bot$ on behalf of $P_1$.
\end{simulatorbox}

\paragraph{Secure Comparison}

The ideal functionality realising protocol $\piBitExt$ is presented in \boxref{fig:FBitExt}.
\begin{systembox}{$\FBitExt$}{Functionality for protocol $\piBitExt$}{fig:FBitExt}
	\justify
	$\FBitExt$ interacts with the parties in $\Partyset$ and the adversary $\Sim$. $\FBitExt$ receives $\shr{\cdot}$-share of value $\val$ from the parties as input. If $\FBitExt$ receives $\bot$ from $\Sim$, then send $\bot$ to every party, else proceed with the computation.
	
	\begin{description}
		\item {\bf Computation of output: } Compute $\val = \Mask{\val} - \PadA{\val} - \PadB{\val} - \PadC{\val}$ and set $\bitb = \MSB(\val)$ where $\MSB$ denotes the most significant bit. Randomly select $\PadA{\bitb}, \PadB{\bitb}, \PadC{\bitb}$ from $\Z{1}$ and set $\Mask{\bitb} = \bitb \xor \PadA{\bitb} \xor \PadB{\bitb} \xor \PadC{\bitb}$. The output shares are set as:
		
		{\small
			\begin{center}
				\begin{tabu} to 1\textwidth { l  l }
					$\shareB{\bitb}_{P_0} = (\PadA{\bitb}, \PadB{\bitb}, \PadC{\bitb})$ & 
					$\shareB{\bitb}_{P_1} = (\Mask{\bitb}, \PadB{\bitb}, \PadC{\bitb})$\\ \\
					$\shareB{\bitb}_{P_2} = (\Mask{\bitb}, \PadC{\bitb}, \PadA{\bitb})$ &
					$\shareB{\bitb}_{P_3} = (\Mask{\bitb}, \PadA{\bitb}, \PadB{\bitb})$\\
				\end{tabu}
			\end{center}
		}
		
		\item {\bf Output to adversary: }  If $\Sim$ sends $\abort$, then send $(\OUTPUT, \bot)$ to all the parties. Otherwise,  send $(\OUTPUT, \shr{\bitb}_{\Sim})$ to the adversary $\Sim$, where $\shr{\bitb}_{\Sim}$ denotes the share of $\bitb$ corresponding to the corrupt party.
		
		\item {\bf Output to selected honest parties: } Receive $(\SELECT, \{I\})$ from adversary $\Sim$, where $\{I\}$ denotes a subset of the honest parties. If an honest party $P_i$ belongs to $I$, send $(\OUTPUT, \bot)$, else send $(\OUTPUT, \shr{\bitb}_{i})$,  where $\shr{\bitb}_{i}$ denotes the share of $\bitb$ corresponding to the honest party $P_i$.
	\end{description}
\end{systembox}

The simulator for the case of corrupt $P_0$ appears in \boxref{fig:Sim_BitExtI}.
\begin{simulatorbox}{$\Sim_{\BitExt}$}{Simulator for the case of corrupt $P_0$}{fig:Sim_BitExtI}
	%
	\justify 
	\algoHead{Offline Phase:}
	\begin{myitemize}
		\item[--] $\Sim_{\BitExt}$ samples a random $\vr$ on behalf of $P_1, P_2$ and set $\wx = \MSB(\vr)$. 
		\item[--] The steps corresponding to $\pivSh$ are simulated similar to $\Sim_{\vSh}$ (\boxref{fig:Sim_vShA}), for the case of corrupt $P_0$.
	\end{myitemize}
	%
	\justify 
	\algoHead{Online Phase:} 
	\begin{myitemize}
		\item[--] The steps corresponding to $\piMult$ and $\piRec$ are simulated similar to $\Sim_{\Mult}$ (\boxref{fig:Sim_MultI}) and $\Sim_{\Rec}$ (\boxref{fig:Sim_RecA}) respectively, for the case of corrupt $P_0$.
		\item[--] The steps corresponding to $\pivSh$ is simulated similar to $\Sim_{\vSh}$ (\boxref{fig:Sim_vShA}), for the case of corrupt $P_0$.
	\end{myitemize}
	
	\noindent If $\flag = 0$, $\Sim_{\BitExt}$ invokes $\FBitExt$ with input $(\PadA{\val}, \PadB{\val}, \PadC{\val})$ on behalf of $P_0$. Else it invokes $\FBitExt$ with input $\bot$ on behalf of $P_0$.
\end{simulatorbox}

The simulator for the case of corrupt $P_1$ appears in \boxref{fig:Sim_BitExtII}.
\begin{simulatorbox}{$\Sim_{\BitExt}$}{Simulator for the case of corrupt $P_0$}{fig:Sim_BitExtII}
	%
	\justify 
	\algoHead{Offline Phase:}
	\begin{myitemize}
		\item[--] $\Sim_{\BitExt}$ samples a random $\vr$ on behalf of $P_2$ and set $\wx = \MSB(\vr)$. 
		\item[--] The steps corresponding to $\pivSh$ are simulated similar to $\Sim_{\vSh}$ (\boxref{fig:Sim_vShB}), for the case of corrupt $P_1$.
	\end{myitemize}
	%
	\justify 
	\algoHead{Online Phase:} 
	\begin{myitemize}
		\item[--] The steps corresponding to $\piMult$ and $\piRec$ are simulated similar to $\Sim_{\Mult}$ (\boxref{fig:Sim_MultII}) and $\Sim_{\Rec}$ (\boxref{fig:Sim_RecB}) respectively, for the case of corrupt $P_1$.
		\item[--] The steps corresponding to $\pivSh$ is simulated similar to $\Sim_{\vSh}$ (\boxref{fig:Sim_vShB}), for the case of corrupt $P_1$.
	\end{myitemize}
	
	\noindent If $\flag = 0$, $\Sim_{\BitExt}$ invokes $\FBitExt$ with input $(\Mask{\val}, \PadB{\val}, \PadC{\val})$ on behalf of $P_1$. Else it invokes $\FBitExt$ with input $\bot$ on behalf of $P_1$.
\end{simulatorbox}

\end{document}